\documentclass{article} 

\usepackage{iclr2027-conference,times}

\usepackage{amsmath,amsfonts,bm}

\def\eqref#1{equation~\ref{#1}}

\def\1{\bm{1}}

\DeclareMathAlphabet{\mathsfit}{\encodingdefault}{\sfdefault}{m}{sl}
\SetMathAlphabet{\mathsfit}{bold}{\encodingdefault}{\sfdefault}{bx}{n}

\usepackage{hyperref}
\usepackage[normalem]{ulem}
\usepackage{dsfont}
\usepackage{url}
\usepackage{graphicx}
\usepackage{xspace}
\usepackage{makecell}
\usepackage{array}
\usepackage{colortbl}
\usepackage{multirow}
\usepackage{subcaption}
\usepackage{tabularx}
\usepackage{color}
\usepackage{pifont}
\usepackage[utf8]{inputenc}
\usepackage[T1]{fontenc}
\usepackage{CJKutf8}
\usepackage{enumitem}
\usepackage{diagbox}
\usepackage{listings}
\usepackage[flushleft]{threeparttable}
\usepackage{booktabs}
\usepackage{wrapfig}
\usepackage{placeins}
\usepackage{algorithmicx,algorithm}
\usepackage{marvosym}

\newcommand{\icon}{\raisebox{-4.1pt}{\includegraphics[width=1.3em]{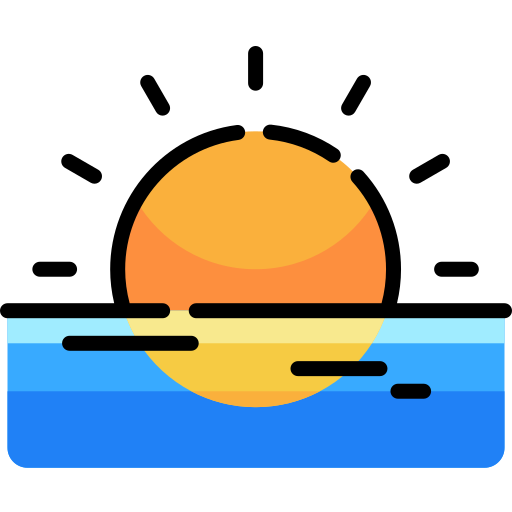}}\xspace}

\definecolor{fbApp}{HTML}{ffe4e3}
\definecolor{mydarkblue}{rgb}{0,0.3,0.9}

\title{\icon Aurora-X: built for eXtreme time series forecasting}

\preprint
\author{Xingjian Wu, Chenjuan Guo\textsuperscript{\Letter}, Xiangfei Qiu, Zhigang Hu, Hanyin Cheng, Peng Chen,\\ \textbf{Yang Shu}, \textbf{Jilin Hu}, \textbf{Bin Yang} \\East China Normal University
\\
\texttt{\{xjwu,xfqiu,zghu,hycheng,pchen\}@stu.ecnu.edu.cn}, \\
\texttt{\{cjguo,yshu,jlhu,byang\}@dase.ecnu.edu.cn}
}
\begin{document}

\maketitle

\begin{abstract}
Time series foundation models (TSFMs) enable cross-domain forecasting, but their development as general-purpose forecasters remains constrained by underexplored training potential and limited architectural versatility. To address these challenges, we introduce Aurora-X, a billion-scale TSFM with a progressive curriculum and a unified architecture. We first use channel-independent pretraining to learn temporal patterns, then introduce cross-variable dependencies, varied context and horizon lengths, and future covariates if available during midtraining. Variable-resolution post-training further enables an adjustable temporal span per token at inference. With fixed model weights, this supports longer histories under a fixed token budget or fewer tokens for the same history, enabling test-time scaling. With a versatile architecture, Aurora-X supports cross-variable modeling, covariate conditioning, and parallel decoding of future patches for probabilistic forecasting. These are supported by a novel pattern-guided mixture-of-experts that expands model capacity through sparse activation and uses shallow patch similarities to constrain deep-layer routing, guiding expert specialization across heterogeneous time series. Furthermore, we propose an implicit quantile network head that predicts arbitrary quantiles to characterize predictive distributions, enhancing probabilistic forecasting flexibility. Comprehensive experiments on GIFT-Eval, TIME, FEV-Bench, TFB, and DAG-Bench demonstrate state-of-the-art forecasting performance against pretrained TSFMs and task-specific supervised models.
\end{abstract}
\begin{figure*}[!htbp]
    \centering
    \includegraphics[width=0.97\linewidth]{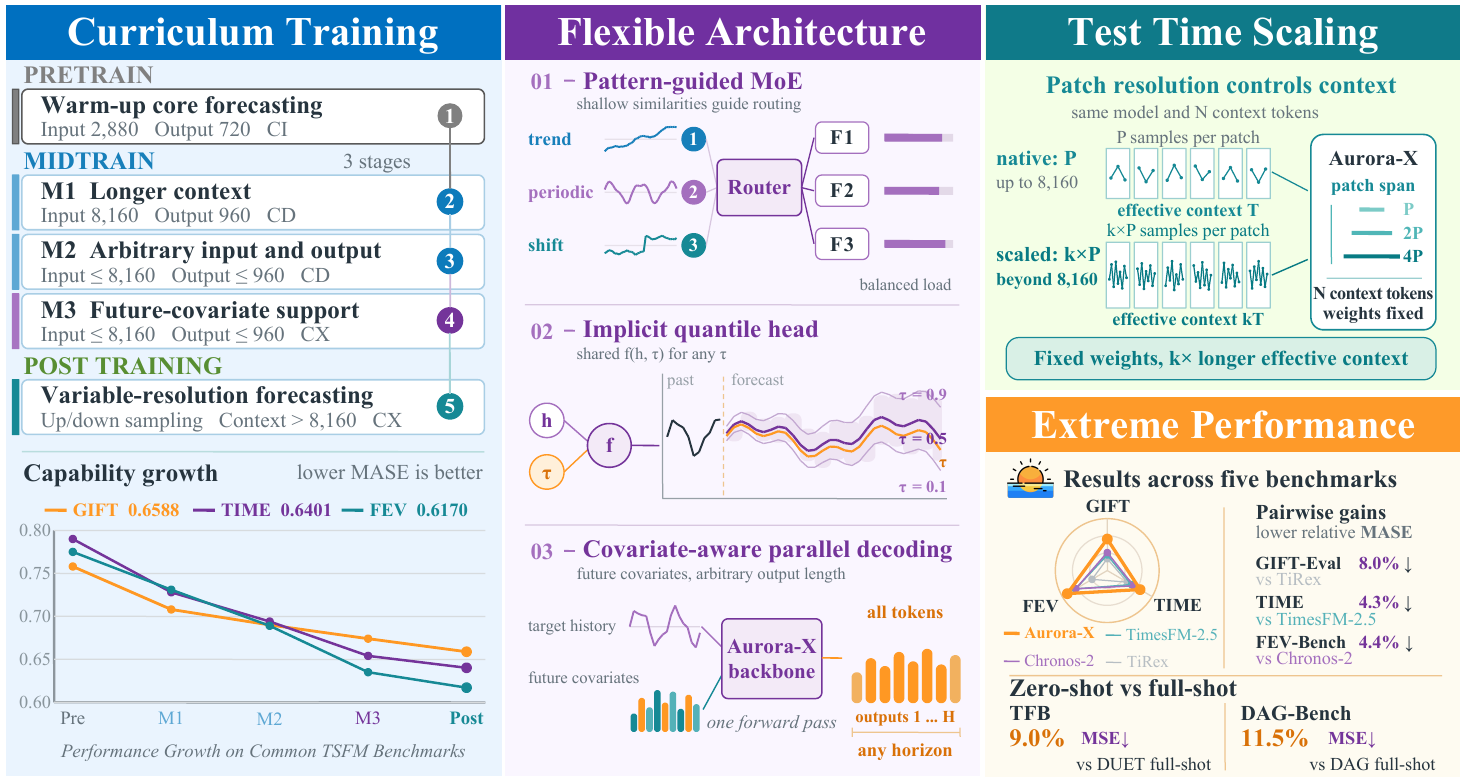}
    \caption{Overview of Aurora-X. CI, CD, and CX denote channel-independent, joint multivariate, and future-covariate training, respectively.}
\label{fig:intro}
\end{figure*}

\clearpage
\section{Introduction}

Time series foundation models (TSFMs)~\citep{timer,google2025timesfm25,ansarichronos,wu2026aurora} learn transferable temporal representations through large-scale pretraining on heterogeneous corpora, enabling zero-shot forecasting across datasets. Beyond zero-shot transfer, practical forecasting requires a model to integrate related variables and future covariates, characterize uncertainty flexibly, and adapt to the available history. However, existing TSFMs still face two challenges in developing and combining these capabilities.

\textbf{[1] Underexplored training potential.} In large language models (LLMs), pretraining builds broad linguistic competence and general knowledge, while midtraining strengthens specialized skills and extends context capacity through targeted continued pretraining~\citep{liu2024deepseek,bai2023qwen}. Post-training further develops deliberative reasoning~\citep{guo2025deepseek} and reasoning-budget control~\citep{aggarwal2025l1}. Although staged training is emerging in TSFMs~\citep{ansari2025chronos2,liu2026timers1}, how to align these stages with data characteristics, task difficulty, and target capabilities remains underexplored, particularly the potential of post-training to expand forecasting capabilities and inference-time flexibility.

\begin{wraptable}{r}{0.5\textwidth}
    \vspace{-1\baselineskip}
    \centering
    \hypersetup{hidelinks}
    \definecolor{tsfmSupported}{HTML}{237A44}
    \definecolor{tsfmPartial}{HTML}{946200}
    \definecolor{tsfmAbsent}{HTML}{B23838}
    \providecommand{\modelref}[3]{\nocite{#1}#2~\hyperlink{cite.#1}{\textcolor{mydarkblue}{(#3)}}}
    \providecommand{\cmark}{\textcolor{tsfmSupported}{\ding{51}}}
    \providecommand{\xmark}{\textcolor{tsfmAbsent}{\ding{55}}}
    \providecommand{\pmark}{\textcolor{tsfmPartial}{\ensuremath{\circ}}}
    \caption{TSFM capabilities: \cmark{} supported; \pmark{} partial; \xmark{} unsupported. Cross-var. includes target--covariate interactions. Future cov.: future covariate values available at forecast time. Prob.: probabilistic forecasting. Cont. quant.: continuous quantile functions or sampling. Par.: parallel decoding within supported horizons (Baguan-TS: inferred). TimesFM-2.5: external regression (XReg) for covariates; blockwise decoding. TTS: test-time scaling.}
    \label{tab:model-features}
    \setlength{\tabcolsep}{1.6pt}
    \renewcommand{\arraystretch}{1.05}
    \scriptsize
    \resizebox{\linewidth}{!}{%
    \begin{tabular}{@{}lcccccc@{}}
        \toprule
        \textbf{Model}
        & \makecell{\textbf{Cross-}\\\textbf{var.}}
        & \makecell{\textbf{Future}\\\textbf{cov.}}
        & \textbf{Prob.}
        & \makecell{\textbf{Cont.}\\\textbf{quant.}}
        & \textbf{Par.}
        & \textbf{TTS} \\
        \midrule
        \modelref{auer2026tirex}{TiRex}{2026} & \xmark & \xmark & \cmark & \pmark & \pmark & \cmark \\
        \modelref{liu2026timers1}{Timer-S1}{2026} & \xmark & \xmark & \cmark & \pmark & \pmark & \xmark \\
        \modelref{liu2025sundial}{Sundial}{2025} & \xmark & \xmark & \cmark & \cmark & \xmark & \xmark \\
        \modelref{google2025timesfm25}{TimesFM-2.5}{2025} & \pmark & \pmark & \cmark & \cmark & \pmark & \xmark \\
        \modelref{timemoe}{Time-MoE}{2024} & \xmark & \xmark & \xmark & \xmark & \xmark & \xmark \\
        \modelref{sun2025xihe}{Xihe}{2025} & \xmark & \xmark & \cmark & \pmark & \cmark & \xmark \\
        \modelref{wu2026aurora}{Aurora}{2026} & \xmark & \xmark & \cmark & \cmark & \cmark & \cmark \\
        \modelref{ansari2025chronos2}{Chronos-2}{2025} & \cmark & \cmark & \cmark & \pmark & \cmark & \xmark \\
        \modelref{liu2026falcon}{Falcon-X}{2026} & \cmark & \xmark & \cmark & \pmark & \cmark & \xmark \\
        \modelref{podest2026tirex2}{TiRex-2}{2026} & \cmark & \cmark & \cmark & \pmark & \pmark & \xmark \\
        \modelref{khwaja2026toto}{Toto-2.0}{2026} & \cmark & \xmark & \cmark & \pmark & \cmark & \xmark \\
        \modelref{yang2026baguan}{Baguan-TS}{2026} & \cmark & \cmark & \cmark & \pmark & \cmark & \xmark \\
        \midrule
        \textbf{Aurora-X (Ours)} & \cmark & \cmark & \cmark & \cmark & \cmark & \cmark \\
        \bottomrule
    \end{tabular}%
    }
    \par
    \vspace{-0.4\baselineskip}
\end{wraptable}

\textbf{[2] Limited architectural versatility.} Channel independence (CI) facilitates pretraining on heterogeneous data, but CI TSFMs such as TiRex~\citep{auer2026tirex} and Timer-S1~\citep{liu2026timers1} do not explicitly model cross-variable dependencies or natively condition on future covariates. Chronos-2~\citep{ansari2025chronos2} and Falcon-X~\citep{liu2026falcon} use predefined quantiles during training and inference, limiting flexibility in characterizing predictive distributions. Sundial~\citep{liu2025sundial} and Time-MoE~\citep{timemoe} extend forecasts autoregressively across output blocks, requiring sequential computation and potentially accumulating errors. Overall, existing TSFM architectures have yet to combine all these capabilities within a single model, and few support test-time scaling (Table~\ref{tab:model-features}).

To address these challenges, we introduce \textbf{Aurora-X}, a billion-scale TSFM with a progressive curriculum and a versatile architecture, as shown in Figure~\ref{fig:intro}. During pretraining, we adopt the CI strategy to learn general temporal patterns, then introduce joint variables, varied context and horizon lengths, and future covariates during midtraining. Through upsampling and downsampling, post-training further enables variable-resolution forecasting with an adjustable temporal span per token. With fixed model weights, this supports longer histories under a fixed token budget or fewer tokens for the same history, enabling test-time scaling.

With a versatile architecture, Aurora-X supports cross-variable and covariate conditioning, long input histories, and parallel decoding. To expand model capacity, we also propose a pattern-guided mixture of experts (MoE) with sparsely activated experts. Shallow patch similarities constrain routing in deeper layers, guiding expert specialization across heterogeneous time series. We also propose an implicit quantile network (IQN) head conditioned on continuous quantile levels. By predicting arbitrary quantiles, it characterizes predictive distributions and enhances probabilistic forecasting flexibility. Our contributions are summarized as:
\begin{itemize}[leftmargin=2em,itemsep=-0.1em]
\item[\ding{182}] We propose \textbf{Aurora-X}, a billion-scale TSFM that supports cross-variable and covariate conditioning, long input histories, variable-resolution inference, and arbitrary quantile prediction within a unified architecture.
\item[\ding{183}] We develop a progressive curriculum to learn temporal patterns during pretraining, extend cross-variable and covariate capabilities during midtraining, and enable variable-resolution inference through post-training.
\item[\ding{184}] We combine pattern-guided MoE routing with covariate-aware parallel decoding and an implicit quantile head. Shallow patch similarities guide deep-layer expert routing, while the head supports flexible probabilistic forecasting through arbitrary quantile prediction.
\item[\ding{185}] Evaluations against TSFMs on GIFT-Eval~\citep{aksu2024gift}, TIME~\citep{qiaotime}, and FEV-Bench~\citep{shchur2025fev}, and supervised models on TFB~\citep{qiu2024tfb} and DAG-Bench~\citep{qiu2025dag}, demonstrate state-of-the-art forecasting performance with favorable computational cost and inference latency.
\end{itemize}

\section{Related Work}
\subsection{Time Series Foundation Models}
To transfer forecasting knowledge across domains, Timer~\citep{timer}, TimesFM~\citep{timesfm}, and Chronos~\citep{ansarichronos} pretrain on heterogeneous corpora. Accommodating forecasting tasks with different numbers and combinations of variables motivates any-variate attention in Moirai~\citep{woo2024moirai}. Chronos-2~\citep{ansari2025chronos2} and TiRex-2~\citep{podest2026tirex2} support multivariate forecasting with known future covariates, while Baguan-TS~\citep{yang2026baguan} integrates covariates into sequence-native in-context learning.

For uncertainty estimation, models offer different ways to access predictive quantiles. Moirai uses mixture distributions, Chronos-2 and Toto-2.0~\citep{khwaja2026toto} regress fixed quantile grids, and Baguan-TS derives quantiles from discretized distributions. Flow matching in Sundial~\citep{liu2025sundial} and Aurora~\citep{wu2026aurora} supports continuous generative distributions, with quantiles estimated through sampling; Aurora also conditions on text and images. Decoding imposes another constraint: TimesFM and Sundial generate successive output blocks autoregressively, while Timer-S1~\citep{liu2026timers1} retains serial hidden-state computation within a single forward pass over its supported horizon. Aurora-X directly predicts marginal quantiles at arbitrary levels over a continuous range while decoding future patches in parallel.

\subsection{Training and Inference Adaptation}
Expert specialization and training progression offer complementary routes to broader capabilities. Sparse experts in Time-MoE~\citep{timemoe} and Timer-S1 scale capacity, while Pathformer~\citep{chen2024pathformer} and TFPS~\citep{sun2024learning} guide specialization through scale-specific pathways and clustering encoded patches, respectively. Aurora-X regularizes deep-layer routing with learned shallow patch similarities, complementing load balancing. Capability development can draw on task diversity within pretraining, as in Moirai's variable context and prediction lengths, or subsequent stages: Timer-S1 uses continued pretraining and context extension, while Chronos-2 undergoes post-training for longer contexts.

Inference-time flexibility comes from computational budgets and input adaptation. Sundial varies sampling and flow-step budgets, while Baguan-TS combines retrieval with ensembles. Aurora and TiRex~\citep{auer2026tirex} adapt temporal resolution for periodic patterns. Aurora-X's curriculum progressively develops temporal modeling, cross-variable dependencies, and covariate conditioning; multi-scale post-training makes temporal resolution controllable at inference, allowing historical coverage to be matched to the context-token budget.

\section{Aurora-X}
\label{sec:aurorax}

\begin{figure*}[!htbp]
    \centering
    \includegraphics[width=1\linewidth]{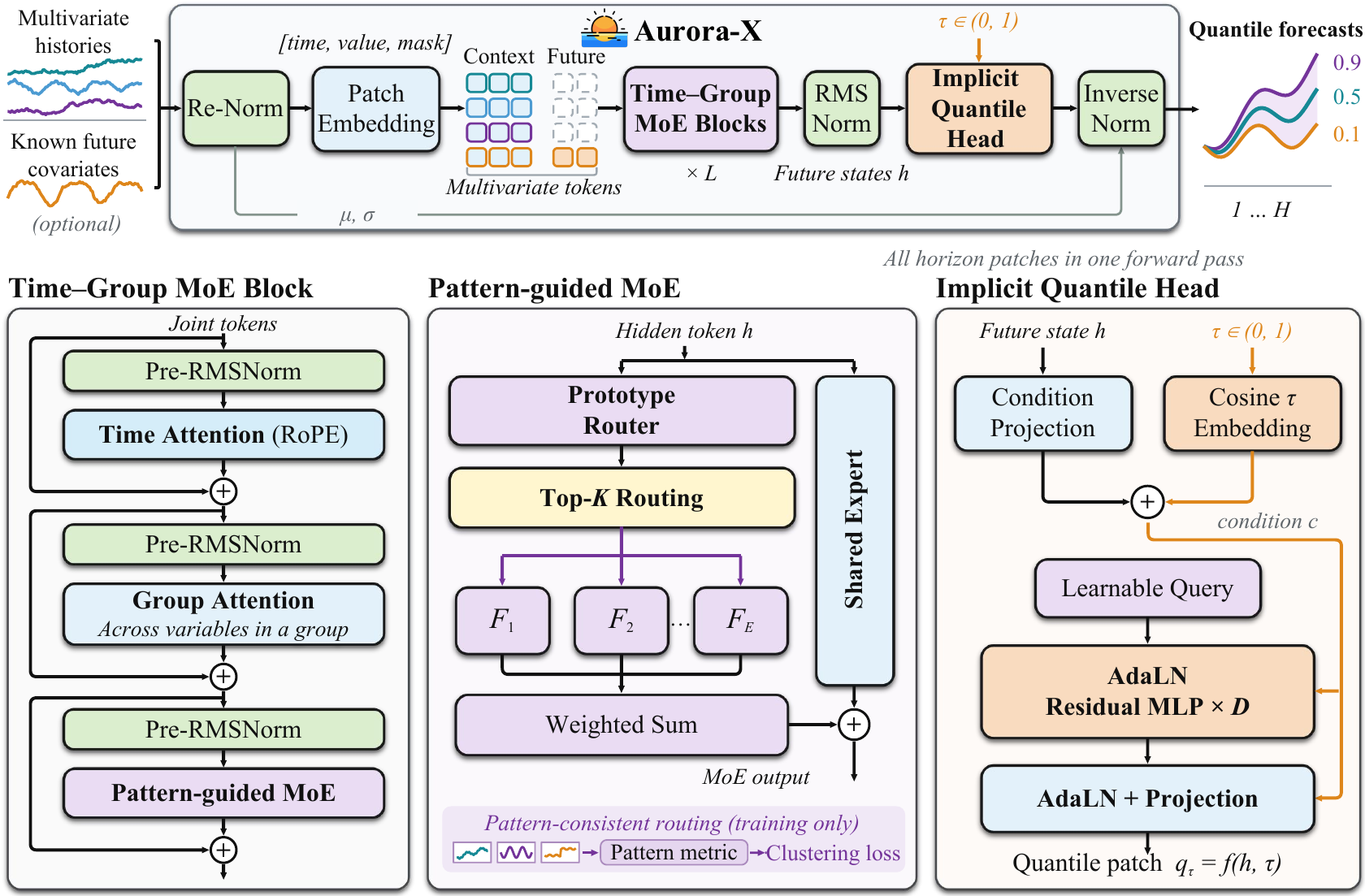}
    \caption{Overview of Aurora-X (1.05B total parameters, $L=12$). Lower panels show a Time--Group block, its MoE sublayer, and the quantile head. Patch Embedding is fed into the blocks, whose normalized future states condition parallel quantile decoding.}
    \label{fig: overview}
\end{figure*}

Aurora-X is a zero-shot forecaster for cross-domain forecasting. Given histories $X\in\mathbb{R}^{C\times T}$ of target and auxiliary variables, optional future covariates if available, and quantile levels $\tau\in(0,1)$, we predict target quantiles over the next $H$ time steps. Figure~\ref{fig: overview} shows the full interface: patch embeddings pass through Time--Group MoE blocks to the quantile head. We pretrain on individual variables, then introduce joint variables and future covariates during midtraining.

\subsection{Unified Backbone}
\label{sec:time-group}

\paragraph{Patch embedding.}
We first apply reversible instance normalization (RevIN)~\citep{kim2021reversible} using each variable's observed-history statistics, followed by $\operatorname{arcsinh}$~\citep{ansari2025chronos2} to compress extreme values. Non-overlapping patches of native length $P$ provide $N=\lceil T/P\rceil$ context tokens for history and $M=\lceil H/P\rceil$ future tokens for prediction positions and available covariates. To support these inputs in parallel forecasting, we use a shared embedding:
\begin{equation}
    z_{v,i}^{(0)}=\operatorname{Embed}\!\left([\mathbf{t}_i;\tilde{\mathbf{x}}_{v,i};\mathbf{m}_{v,i}]\right)\in\mathbb{R}^{d},
    \label{eq:tokenization}
\end{equation}
where $\mathbf t_i,\tilde{\mathbf x}_{v,i},\mathbf m_{v,i}\in\mathbb R^P$ denote relative times, normalized values, and observation masks, and $\operatorname{Embed}:\mathbb R^{3P}\to\mathbb R^d$ is a shared residual MLP. Future targets receive zero values and masks during both training and inference, while known covariates retain observations. We exclude incomplete history patches as attention keys but keep future tokens active.

\paragraph{Time and group attention.}
We encode $Z^{(0)}\in\mathbb{R}^{C\times(N+M)\times d}$ through $L$ Time--Group MoE blocks. Each block models temporal dependencies within variables, mixes variables in group $\mathcal G_g$ at aligned patch positions, and updates tokens through MoE:
\begin{gather}
 U^{(\ell)}_{v,:,:}=Z^{(\ell-1)}_{v,:,:}+\operatorname{Attn}_{T,\ell}^{\mathrm{RoPE}}\!\left(\operatorname{RN}(Z^{(\ell-1)}_{v,:,:})\right),\nonumber\\
 V^{(\ell)}_{\mathcal G_g,i,:}=U^{(\ell)}_{\mathcal G_g,i,:}+\operatorname{Attn}_{G,\ell}\!\left(\operatorname{RN}(U^{(\ell)}_{\mathcal G_g,i,:})\right),\label{eq:time-group}\\
 Z^{(\ell)}=V^{(\ell)}+\operatorname{MoE}_{\ell}\!\left(\operatorname{RN}(V^{(\ell)})\right),\nonumber
\end{gather}
where $\operatorname{RN}$ denotes RMSNorm~\citep{zhang2019root}, and $U^{(\ell)}$ and $V^{(\ell)}$ are the intermediate states. Temporal attention is bidirectional over $(N+M)\times d$ states; group attention operates on $|\mathcal G_g|\times d$ states, both using unscaled scores. Singleton groups retain group attention during CI pretraining; midtraining enables cross-variable mixing. Finally, RMSNorm provides $h_{v,j}=\operatorname{RN}(Z^{(L)})_{v,N+j,:}\in\mathbb R^d$, $j=1,\ldots,M$, to the quantile head.

\subsection{Pattern-Guided Mixture of Experts}
\label{sec:pattern-moe}

To accommodate heterogeneous temporal patterns, we replace each Time--Group block's feed-forward sublayer with a sparsely routed MoE trained jointly with the backbone. Sparse activation expands model capacity, but load balancing alone does not guide specialization as deep temporal and cross-variable mixing can weaken pattern distinctions. We therefore guide specialization with shallow patch similarities, encouraging similar local shapes to share experts.

\paragraph{Prototype router and top-$K$ routing.}
Each layer contains one shared expert and $E=16$ routed experts ($F_1,\ldots,F_E$ in Figure~\ref{fig: overview}), with $K=4$ activated per token. Given the MoE input $h=h^{(\ell)}_{v,i}=\operatorname{RN}(V^{(\ell)})_{v,i,:}$, we compute its affinity to each expert's learnable prototype $c_e\in\mathbb R^d$:
\begin{equation}
    a_e=\frac{(W_K\,h)^{\top}(W_Q\,c_e)}{\sqrt{d_{\mathrm{rout}}}},\qquad
    \mathbf{s}=\operatorname{LN}_{E}(\mathbf{a}),\qquad
    \mathbf{p}=\operatorname{Softmax}(\mathbf{s}),
    \label{eq:prototype-router}
\end{equation}
where $W_K,W_Q\in\mathbb R^{d_{\mathrm{rout}}\times d}$ project states and prototypes, $\mathbf a=[a_e]_{e=1}^{E}$, and $\operatorname{LN}_{E}$ normalizes logits across experts without affine parameters. We select the $K$ highest-scoring experts, denoted by $\mathcal E(h)$, and combine their outputs with the shared expert:
\begin{equation}
    \operatorname{MoE}(h)=F_{\mathrm{shared}}(h)+
    \sum_{e\in\mathcal{E}(h)}g_e(h)F_e(h),\qquad
    g_e(h)=\frac{p_e(h)}{\sum_{e'\in\mathcal{E}(h)}p_{e'}(h)}.
    \label{eq:expert-aggregation}
\end{equation}
Here $F_e,F_{\mathrm{shared}}:\mathbb R^d\to\mathbb R^d$ preserve the token dimension; each layer learns its own experts, prototypes, and projections.

\paragraph{Pattern metric and clustering loss.}
To preserve local pattern information in deeper layers, we learn a pattern metric from normalized context value patches during training. For patch $\tilde{\mathbf{x}}_{b,i}\in\mathbb R^P$ in minibatch variable row $b$, we compute
\begin{equation}
    u_{b,i}=\operatorname{Normalize}\!\left(W_{\mathrm{pat}}\operatorname{sg}(\tilde{\mathbf{x}}_{b,i})\right),\qquad
    S_{b,ij}=\exp\!\left(\frac{u_{b,i}^{\top}u_{b,j}-1}{\sigma_{\mathrm{pat}}^2}\right),
    \label{eq:pattern-similarity}
\end{equation}
where $W_{\mathrm{pat}}\in\mathbb R^{d_{\mathrm{pat}}\times P}$ is learned per layer, $\operatorname{Normalize}$ denotes $\ell_2$ normalization, and $\sigma_{\mathrm{pat}}$ controls the similarity bandwidth. Similar projected shapes yield larger $S_{b,ij}$, providing a shallow reference for routing. The stop-gradient operator $\operatorname{sg}$ detaches patch inputs while keeping the projection learnable; parameter layer indices are omitted.

For context positions $i,j\leq N$ within each variable row, let $D_{b,ie}=\mathbf{1}[e\in\mathcal{E}(h^{(\ell)}_{b,i})]$ indicate expert selection, with vector $\mathbf D_{b,i}$. We connect these discrete assignments to the pattern metric through a straight-through estimator:
\begin{equation}
    \mathbf{A}_{b,i}=\operatorname{sg}(\mathbf{D}_{b,i})+\mathbf{p}_{b,i}-\operatorname{sg}(\mathbf{p}_{b,i}),\qquad
    \Omega_{b,ij}=\frac{1}{K}\sum_{e=1}^{E}A_{b,ie}A_{b,je}.
    \label{eq:routing-overlap}
\end{equation}
Here $\Omega_{b,ij}$ is the fraction of selected experts shared by two patches in the forward pass, while gradients follow softmax probabilities to update the router and upstream representations. We use this overlap to weight pattern similarities and regularize the projection:
\begin{equation}
    \mathcal{L}_{\mathrm{pat}}
    =-\frac{1}{B}\sum_{b=1}^{B}
    \frac{\sum_{i,j=1}^{N}\Omega_{b,ij}S_{b,ij}}
    {\max\!\left(\sum_{i,j=1}^{N}\Omega_{b,ij},\epsilon\right)},\qquad
    \mathcal{R}_{\mathrm{orth}}
    =\frac{\|\bar W_{\mathrm{pat}}\,\bar W_{\mathrm{pat}}^{\top}-\mathbf{I}\|_F^2}{d_{\mathrm{pat}}^2},
    \label{eq:pattern-loss}
\end{equation}
where $B$ counts variable rows across the minibatch, $\bar W_{\mathrm{pat}}$ contains row-normalized weights, $\mathbf I$ is the identity matrix, and $\epsilon>0$ is a numerical floor. The normalized objective includes self-pairs. For fixed similarities, it favors sharing above the current weighted mean similarity and discourages sharing below it. Learning $W_{\mathrm{pat}}$ draws co-assigned patches together, while orthogonality promotes diverse projection directions. This guides specialization across heterogeneous series by linking deep expert assignments to shallow temporal patterns; Appendix~\ref{app:clustering-theory} provides the analysis and proofs.

We complement pattern guidance with load balancing, $\mathcal L_{\mathrm{bal}}=E\sum_e f_e\bar{p}_e$, to spread utilization across experts. Here $f_e=J^{-1}\sum_tD_{t,e}$ and $\bar{p}_e=J^{-1}\sum_t p_{t,e}$, where $J$ counts both context and future tokens in a minibatch.

\subsection{Implicit Quantile Head}
\label{sec:iqn}

Probabilistic forecasting should characterize predictive distributions flexibly and efficiently. Fixed quantile heads~\citep{ansari2025chronos2,liu2026timers1} describe distributions at predefined levels, limiting direct support for customized intervals and tail queries. Sample-based forecasters~\citep{liu2025sundial,wu2026aurora} estimate these quantities from multiple generated trajectories, introducing sampling overhead. We therefore adopt an implicit quantile network (IQN) conditioned on continuous quantile levels to achieve flexible probabilistic forecasting. It directly predicts arbitrary quantiles in parallel from shared future states, enabling characterization of marginal distributions without trajectory sampling.

\paragraph{Quantile conditioning.}
Given a normalized future state $h\in\mathbb R^d$ from the final Time--Group MoE block (Figure~\ref{fig: overview}), the head $q_{\theta}(h,\tau)\in\mathbb R^P$ predicts the conditional $\tau$-quantile at each patch position. We combine a projection of $h$ with a cosine embedding of the requested level $\tau\in(0,1)$:
\begin{equation}
    \phi(\tau)=\left[\cos(\pi\,n\,\tau)\right]_{n=0}^{R-1},\qquad
    c(h,\tau)=W_h\,h+b_h+\operatorname{MLP}_{\tau}\!\left(\phi(\tau)\right)\in\mathbb R^d.
    \label{eq:quantile-condition}
\end{equation}
Here $\phi(\tau)\in\mathbb R^R$ represents the quantile level at multiple frequencies, and $c(h,\tau)$ integrates it with the future state. We then refine a shared learnable query $\mathbf v_0\in\mathbb R^d$ through $D$ residual MLP layers with adaptive layer normalization:
\begin{equation}
    \mathbf{v}_{a+1}=\mathbf{v}_a+\alpha_a(c)\odot\operatorname{MLP}_a\!\left(
    [1+\gamma_a(c)]\odot\operatorname{LN}(\mathbf{v}_a)+\beta_a(c)\right),
    \label{eq:iqn-block}
\end{equation}
where $a=0,\ldots,D-1$, and $\beta_a(c),\gamma_a(c),\alpha_a(c)\in\mathbb R^d$ provide the condition-dependent shift, scale, and residual gate. This modulation incorporates the future state and requested level at every layer. A final adaptive normalization and projection yield $P$ quantile values, with head parameters shared across variables, patches, and levels.

\paragraph{Quantile Optimization.}
During training, we independently sample $Q$ levels from $\mathcal U(0,1)$ for each target patch at every step. We supervise the corresponding predictions with the masked pinball loss:
\begin{equation}
    \ell_{\mathrm{IQN}}(h,\tilde{\mathbf y})=
    \frac{\sum_{q=1}^{Q}\sum_{t=1}^{P}m_t\,
    \rho_{\tau_q}\!\left(\tilde y_t-q_{\theta,t}(h,\tau_q)\right)}
    {Q\max(1,\sum_{t=1}^{P}m_t)},\qquad
    \rho_{\tau}(e)=\max\{\tau e,(\tau-1)e\}.
    \label{eq:iqn-loss}
\end{equation}
Here $\tilde y_t$ is the normalized target and $m_t$ indicates valid target positions. The loss weights underestimation by $\tau$ and overestimation by $1-\tau$, guiding each prediction toward its requested conditional quantile. Averaging over target patches gives $\mathcal L_{\mathrm{IQN}}$; known covariates provide conditioning and are excluded from this loss. Uniform sampling approximates the integrated pinball loss over quantile levels, without an explicit non-crossing constraint.

During inference, one backbone evaluation provides the future states for parallel decoding of all patch--quantile pairs. For each target and requested level, the $M$ predicted patches are concatenated and cropped to $H$ steps, then transformed by $\sinh$ and inverse RevIN to obtain the forecast.

\subsection{Curriculum and Variable-Resolution Post-Training}
\label{sec:training}

\paragraph{Capability-oriented curriculum.}
We progressively introduce richer forecasting inputs through the curriculum in Figure~\ref{fig:intro}. We first pretrain on individual variables using singleton groups. During midtraining, M1 adds multivariate groups and longer contexts and horizons; M2 varies these lengths; M3 introduces future covariates if available. Finally, multi-scale post-training develops variable-resolution forecasting (Figure~\ref{fig:resample}). Each stage continues the preceding checkpoint. To jointly learn quantile prediction and expert specialization, we optimize
\begin{equation}
    \mathcal{L}=\mathcal{L}_{\mathrm{IQN}}+
    \sum_{\ell=1}^{L}\left[
    \lambda_{\mathrm{bal}}\mathcal{L}_{\mathrm{bal}}^{(\ell)}+
    \lambda_{\mathrm{pat}}\left(\mathcal{L}_{\mathrm{pat}}^{(\ell)}+
    \lambda_{\mathrm{orth}}\mathcal{R}_{\mathrm{orth}}^{(\ell)}\right)\right].
    \label{eq:total-loss}
\end{equation}
Here $\lambda_{\mathrm{bal}}$, $\lambda_{\mathrm{pat}}$, and $\lambda_{\mathrm{orth}}$ control load balancing, pattern guidance, and orthogonality, respectively.

\noindent
\begin{minipage}[t]{0.48\textwidth}
\vspace{0pt}
\frenchspacing
\paragraph{Multi-scale post-training.} Pretraining and midtraining use a fixed temporal span per token (fixed resolution), so longer histories require more tokens. We therefore introduce multi-scale post-training to teach the model to forecast across resolutions, allowing longer histories within a fixed token budget or fewer tokens for the same history at inference. On a one-billion-point subset (Appendix~\ref{app:training-settings}), we resample history and future separately after RevIN and $\operatorname{arcsinh}$, aligning scales and block boundaries across related variables. For native patch length $P=48$ and positive scale $k\in\mathcal K$, we resample blocks of integer length $kP$ to $P$ values; each token covers $kP$ original points. $k<1$ upsamples shorter spans and $k>1$ downsamples longer spans. Let $\tilde x_j^{(k)}=\tilde x_{(j-1)kP+1:jkP}$ and $\mathcal R_{a\rightarrow b}$ denote linear resampling. We construct the augmented sequence as:
\begin{equation}
    \mathcal{A}_k(\tilde x)=\operatorname{Concat}_{j}
    \left[\mathcal{R}_{kP\rightarrow P}\!\left(\tilde x_j^{(k)}\right)\right].
    \label{eq:resolution-augmentation}
\end{equation}
\end{minipage}\hfill
\begin{minipage}[t]{0.50\textwidth}
    \vspace{0pt}
    \captionsetup{hypcap=false}
    \centering
    \includegraphics[width=\linewidth]{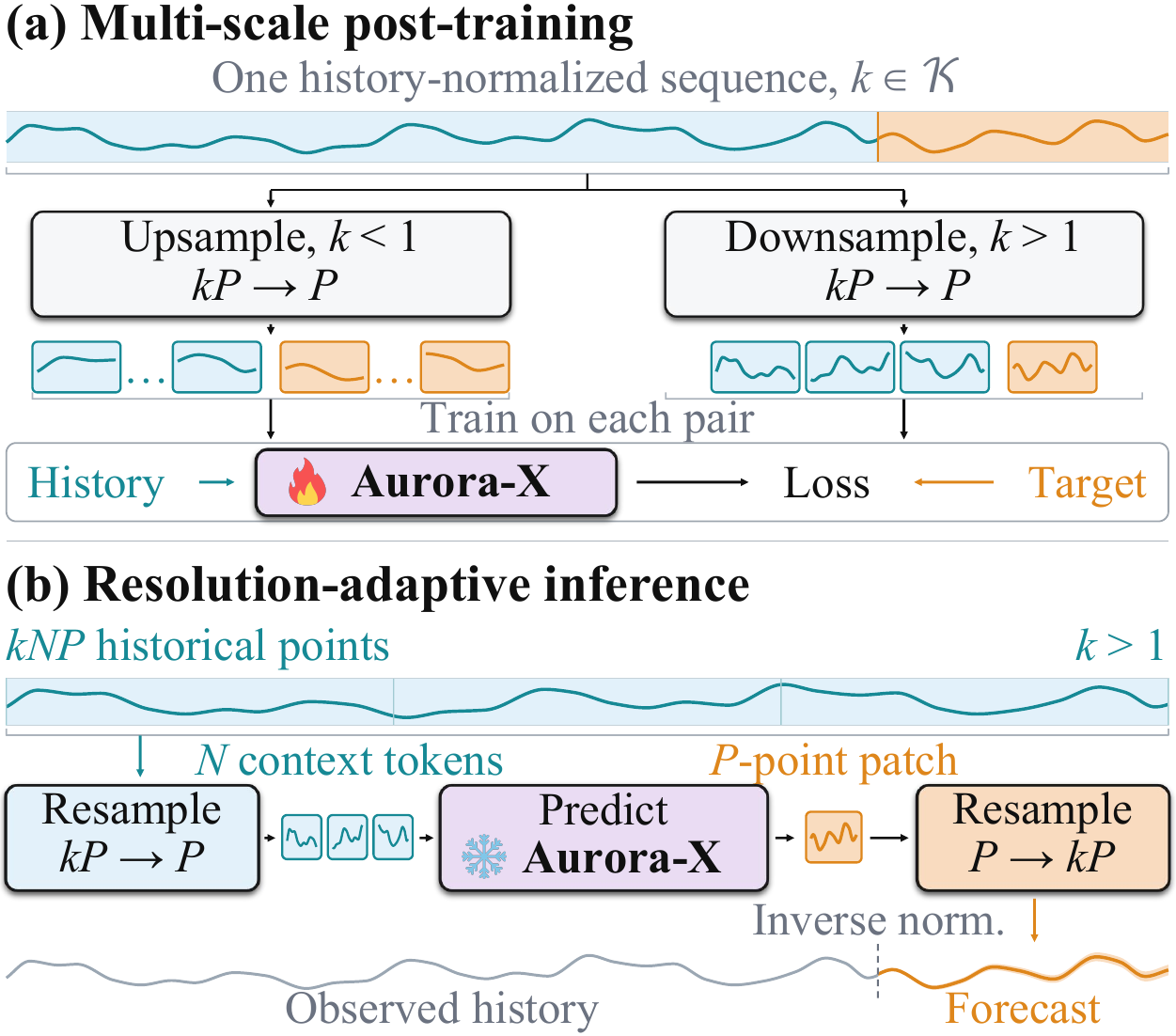}
    \captionof{figure}{(a) Post-training uses resampled history--target pairs. (b) Inference extends historical coverage at a fixed context-token budget and returns forecasts to the requested resolution.}
    \label{fig:resample}
    \vspace{-1mm}
\end{minipage}
\par\medskip
\begin{samepage}
The resampled inputs contain $N_k=\lceil T/(kP)\rceil$ context tokens and $M_k=\lceil H/(kP)\rceil$ future tokens on the native $P$-point grid. Given their normalized future states $h_{v,j}^{(k)}\in\mathbb R^d$ and normalized targets $\tilde y_{v,1:H}$, we apply the same transformation to the targets and optimize
\begin{equation}
    \mathcal L_{\mathrm{IQN}}^{\mathrm{post}}(k)
    =\frac{1}{|\mathcal I_k|}\sum_{(v,j)\in\mathcal I_k}
    \ell_{\mathrm{IQN}}\!\left(h_{v,j}^{(k)},
    \mathcal R_{kP\rightarrow P}\!\left(\tilde y_{v,(j-1)kP+1:jkP}\right)\right),
    \label{eq:post-training-loss}
\end{equation}
\end{samepage}
where $\mathcal I_k$ indexes target patches $(v,j)$, $1\leq j\leq M_k$. We pad incomplete blocks and mask invalid target positions (Equation~(\ref{eq:iqn-loss})). This objective replaces $\mathcal L_{\mathrm{IQN}}$ in Equation~(\ref{eq:total-loss}), while MoE regularization is evaluated on the augmented tokens.

\paragraph{Resolution-adaptive inference.}
During inference, we choose $k$ to control the observations represented by each token (Figure~\ref{fig:resample}(b)). We first map normalized histories and covariates with $\mathcal R_{kP\rightarrow P}$. After quantile decoding, we restore each length-$P$ patch with $\mathcal R_{P\rightarrow kP}$, concatenate the patches, crop to $H$ steps, and apply the inverse transformation. Model weights remain fixed, and time coordinates use the native grid. For $k>1$, a fixed history requires fewer context tokens; at a fixed context-token budget $N$, historical coverage expands from $NP$ to $kNP$ observations. We evaluate both settings in Appendix~\ref{app:token-budget}. Existing models~\citep{wang2025lightgts,wu2025flame,wu2026aurora} also adapt resolution; Appendices~\ref{app:resolution-placement}--\ref{app:lightgts-flex} compare our resampling with projection reparameterization and LightGTS's Flex-resize~\citep{wang2025lightgts}.

\section{Experiments}
\label{sec:experiments}

\begin{wrapfigure}{r}{0.48\textwidth}
    \vspace{-3.5mm}
    \centering
    \includegraphics[width=\linewidth]{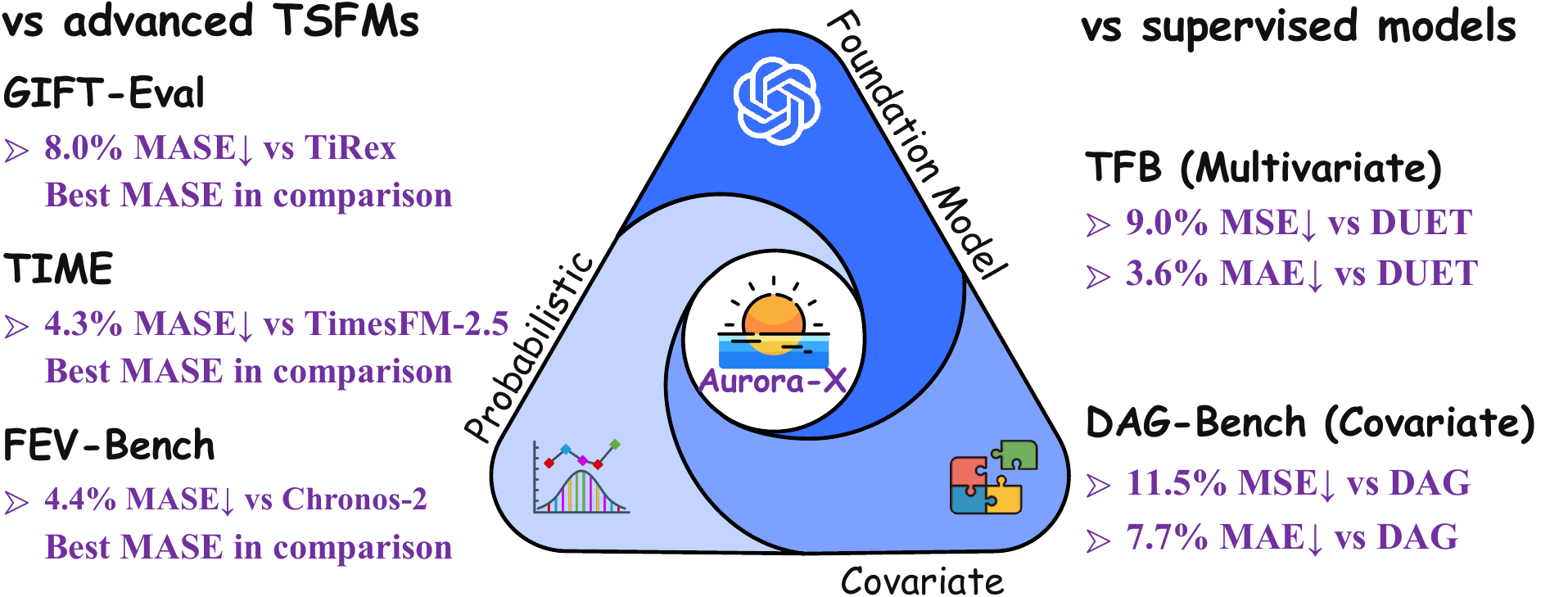}
    \caption{Forecasting performance across five benchmarks.}
    \label{fig:experiment-summary}
    \vspace{-2mm}
\end{wrapfigure}

We evaluate Aurora-X on five benchmarks covering general forecasting, multivariate forecasting, and forecasting with known future covariates. Figure~\ref{fig:experiment-summary} summarizes its overall performance against pretrained TSFMs and task-specific supervised models. We present the GIFT-Eval, TIME, and FEV-Bench results below, followed by analyses of expert routing, forecasting heads, and training-stage gains in Section~\ref{sec:model-analysis}. Comparisons on TFB and DAG-Bench are reported in Appendix~\ref{app:multivariate-covariate}. Appendix~\ref{app:main-results} provides detailed settings, complete results, and expanded probabilistic rankings. Inference efficiency and forecasting visualizations are presented in Appendices~\ref{app:inference-efficiency} and~\ref{app:forecast-visualizations}, respectively.

\subsection{General Forecasting Performance}
\label{sec:general-forecasting}

\begin{figure}[!htbp]
    \centering
    \includegraphics[width=\linewidth,trim=0 6bp 0 0,clip]{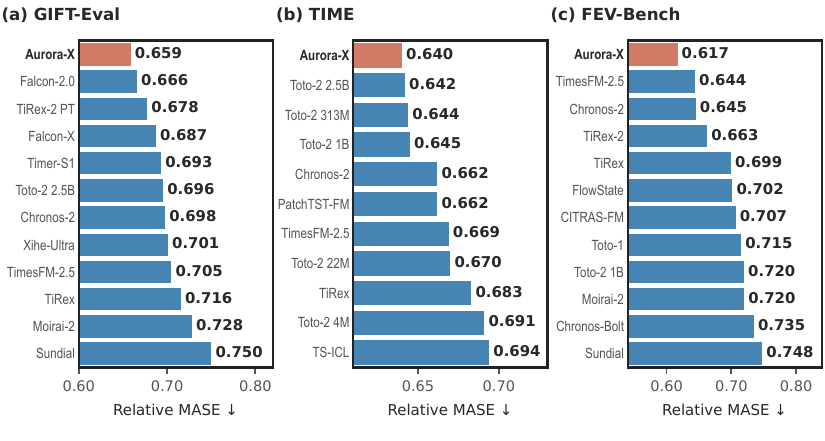}
    \caption{Relative MASE on GIFT-Eval, TIME, and FEV-Bench: geometric means of configuration-wise ratios to Seasonal Naive; lower is better. Aurora-X (1.05B total parameters) is highlighted in color. Model selection is detailed in Appendix~\ref{app:baseline-selection}.}
    \label{fig:mase-comparison}
\end{figure}

\begin{samepage}
As shown in Figure~\ref{fig:mase-comparison}, Aurora-X achieves the lowest relative MASE among the compared TSFMs on GIFT-Eval, TIME, and FEV-Bench. On GIFT-Eval, it reduces relative MASE by 1.1\% compared with Falcon-2.0 and by 8.0\% compared with TiRex. On TIME, the margin over Toto-2.0 (2.5B) is smaller, at 0.3\%, while the reduction relative to TimesFM-2.5 is 4.3\%. On FEV-Bench, Aurora-X improves over TimesFM-2.5 and Chronos-2 by 4.2\% and 4.4\%, respectively. These results show a consistent advantage in point forecasting across the three benchmark collections. Complementary probabilistic results are reported in Appendix~\ref{app:point-probabilistic}.
\par
\end{samepage}

\subsection{Model Analysis}
\label{sec:model-analysis}

The architecture comparisons use the same pretraining corpus, with all MoE variants using 16 routed experts, top-4 selection per token, and the same expert width as Aurora-X. Each comparison replaces the relevant module within the Aurora-X backbone. We examine both expert specialization and forecasting accuracy, then trace how the full curriculum develops the model's capabilities from temporal modeling to covariate conditioning and variable-resolution forecasting.

\par\medskip
\noindent\begin{minipage}{\linewidth}
    \centering
    \includegraphics[width=\linewidth,trim=0 9bp 0 0,clip]{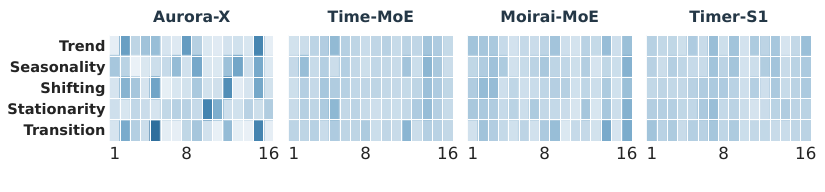}
    \captionof{figure}{Expert-selection shares across five TFB-based scenarios, with 16 routed experts and top-4 selection. Columns index experts; rows sum to one. Darker shading indicates higher normalized selection frequency on a shared color scale.}
    \label{fig:moe-routing}
\end{minipage}
\par\medskip

\paragraph{MoE design.}
Load balancing promotes expert utilization, but specialization requires distinct routing preferences for different temporal patterns. We compare our MoE with implementations based on Time-MoE, Moirai-MoE, and Timer-S1. Following TFB's~\citep{qiu2024tfb} characterization, we select datasets representing five temporal properties and inspect routing at layer 6 of the 12-layer backbone (Figure~\ref{fig:moe-routing}). Distinct rows indicate scenario-dependent expert preferences; similar rows suggest weaker differentiation across these scenarios. This probes whether input-specific routing persists after repeated representation mixing. We then compare forecasting accuracy on GIFT-Eval, TIME, and FEV-Bench (Figure~\ref{fig:moe-performance}). Appendix~\ref{app:architecture-protocol} details dataset selection, routing-frequency aggregation, and the shared forecasting evaluation protocol for these comparisons.

\paragraph{Forecasting head.}
To examine the effect of quantile parameterization, the head comparison includes IQN, fixed quantile regression (Chronos-2), flow matching (Sundial), and a mixture distribution head (Moirai). The backbone architecture, pretraining data, and evaluation horizons are shared. Evaluation uses common quantile levels; generative heads obtain them from their predictive distributions. Figure~\ref{fig:head-performance} reports point accuracy from median forecasts using relative MASE against Seasonal Naive, as in Figures~\ref{fig:mase-comparison} and~\ref{fig:moe-performance}.
\par
\noindent\begin{minipage}[t]{0.48\linewidth}
    \vspace{0pt}
    \centering
    \includegraphics[width=\linewidth,trim=0 18bp 0 9bp,clip]{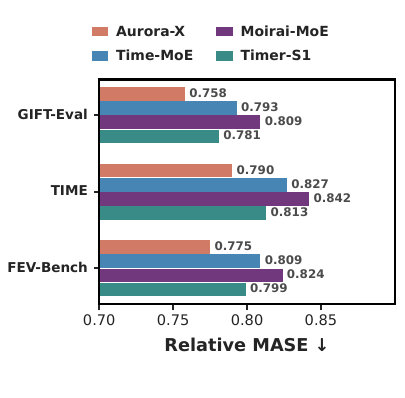}
    \captionof{figure}{Relative MASE of four MoE implementations after pretraining. Lower is better.}
    \label{fig:moe-performance}
\end{minipage}\hfill
\begin{minipage}[t]{0.48\linewidth}
    \vspace{0pt}
    \centering
    \includegraphics[width=\linewidth,trim=0 18bp 0 9bp,clip]{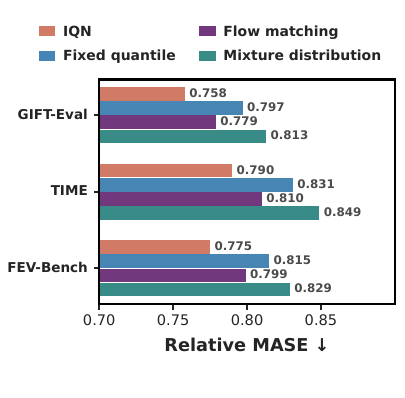}
    \captionof{figure}{Relative MASE of four forecasting heads after pretraining. Lower is better.}
    \label{fig:head-performance}
\end{minipage}
\par

\noindent\begin{minipage}[t]{0.49\textwidth}
\vspace{2mm}
\textbf{Training-stage gains.}
Figure~\ref{fig:training-stage-main} follows Aurora-X through pretraining, three midtraining stages, and post-training. Relative MASE decreases at every transition, while the largest stage-wise reductions occur from Pre to M1 on GIFT-Eval and TIME, and from M2 to M3 on FEV-Bench. These trajectories show that improvements accumulate across stages, with their relative contributions varying by benchmark. Overall, relative MASE decreases by 13.1\%, 19.0\%, and 20.4\% from Pre to Post on GIFT-Eval, TIME, and FEV-Bench, respectively. Post-training yields a further reduction of 2.1--2.8\% relative to M3. Together, these results support a progressive curriculum that develops forecasting capabilities beyond the pretrained checkpoint. Appendix~\ref{app:training-stages} reports the complete stage-wise scores.
\end{minipage}\hfill
\begin{minipage}[t]{0.48\textwidth}
\vspace{6pt}
\centering
\includegraphics[width=\linewidth,trim=0 8bp 0 0,clip]{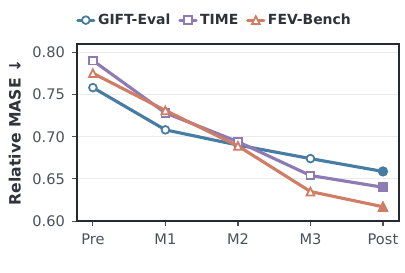}
\captionof{figure}{Relative MASE across training stages. Pre and Post denote pretraining and post-training; M1--M3 denote midtraining stages.}
\label{fig:training-stage-main}
\end{minipage}

\section{Conclusion}
We introduce Aurora-X, a billion-scale time series foundation model combining pattern-guided MoE routing, an implicit quantile head, and a capability-oriented curriculum. Variable-resolution post-training enables context-token budget control and longer historical coverage. Strong performance across five forecasting benchmarks, together with favorable computational cost and inference latency, supports Aurora-X as an \textit{out-of-the-box} tool for decision intelligence.

\clearpage

\bibliography{reference}

@inproceedings{wu2026aurora,
  title     = {Aurora: Towards Universal Generative Multimodal Time Series Forecasting},
  author    = {Wu, Xingjian and Jin, Jianxin and Qiu, Wanghui and Chen, Peng and Shu, Yang and Yang, Bin and Guo, Chenjuan},
  booktitle = {ICLR},
  year      = {2026}
}

@article{aggarwal2025l1,
  title={L1: Controlling how long a reasoning model thinks with reinforcement learning},
  author={Aggarwal, Pranjal and Welleck, Sean},
  journal={arXiv preprint arXiv:2503.04697},
  year={2025}
}

@article{wang2024timexer,
  title={Timexer: Empowering transformers for time series forecasting with exogenous variables},
  author={Wang, Yuxuan and Wu, Haixu and Dong, Jiaxiang and Qin, Guo and Zhang, Haoran and Liu, Yong and Qiu, Yunzhong and Wang, Jianmin and Long, Mingsheng},
  journal={Advances in Neural Information Processing Systems},
  volume={37},
  pages={469--498},
  year={2024}
}

@inproceedings{wu2025srsnet,
  title     = {Enhancing Time Series Forecasting through Selective Representation Spaces: A Patch Perspective},
  author    = {Wu, Xingjian and Qiu, Xiangfei and Cheng, Hanyin and Li, Zhengyu and Hu, Jilin and Guo, Chenjuan and Yang, Bin},
  booktitle = {NeurIPS},
  year      = {2025}
}

@inproceedings{wu2025k2vae,
  title     = {{K${}^2$VAE}: A Koopman-Kalman Enhanced Variational AutoEncoder for Probabilistic Time Series Forecasting},
  author    = {Wu, Xingjian and Qiu, Xiangfei and Gao, Hongfan and Hu, Jilin and Yang, Bin and Guo, Chenjuan},
  booktitle = {ICML},
  year      = {2025}
}

@article{bai2023qwen,
  title={Qwen technical report},
  author={Bai, Jinze and Bai, Shuai and Chu, Yunfei and Cui, Zeyu and Dang, Kai and Deng, Xiaodong and Fan, Yang and Ge, Wenbin and Han, Yu and Huang, Fei and others},
  journal={arXiv preprint arXiv:2309.16609},
  year={2023}
}

@article{guo2025deepseek,
  title={Deepseek-r1: Incentivizing reasoning capability in llms via reinforcement learning},
  author={Guo, Daya and Yang, Dejian and Zhang, Haowei and Song, Junxiao and Wang, Peiyi and Zhu, Qihao and Xu, Runxin and Zhang, Ruoyu and Ma, Shirong and Bi, Xiao and others},
  journal={arXiv preprint arXiv:2501.12948},
  year={2025}
}

@article{liu2024deepseek,
  title={Deepseek-v3 technical report},
  author={Liu, Aixin and Feng, Bei and Xue, Bing and Wang, Bingxuan and Wu, Bochao and Lu, Chengda and Zhao, Chenggang and Deng, Chengqi and Zhang, Chenyu and Ruan, Chong and others},
  journal={arXiv preprint arXiv:2412.19437},
  year={2024}
}

@article{yang2026baguan,
  title={Baguan-TS: A Sequence-Native In-Context Learning Model for Time Series Forecasting with Covariates},
  author={Yang, Linxiao and Jiang, Xue and Xu, Gezheng and Zhou, Tian and Yang, Min and Zhu, ZhaoYang and Geng, Linyuan and Zeng, Zhipeng and Chen, Qiming and Gu, Xinyue and others},
  journal={arXiv preprint arXiv:2603.17439},
  year={2026}
}

@article{sun2025xihe,
  title={Xihe: Scalable Zero-Shot Time Series Learner Via Hierarchical Interleaved Block Attention},
  author={Sun, Yinbo and Fang, Yuchen and Zhu, Zhibo and Li, Jia and Liu, Yu and Deng, Qiwen and Zhou, Jun and Yu, Hang and Lu, Xingyu and Ma, Lintao},
  journal={arXiv preprint arXiv:2510.21795},
  year={2025}
}

@article{podest2026tirex2,
  title={TiRex-2: Generalizing TiRex to Multivariate Data and Streaming},
  author={Podest, Patrick and Pichler, Marco and B{\"u}rger, Elias and Z{\'o}lyomi, Levente and Voggenberger, Bernhard and Berghammer, Wilhelm and Klotz, Daniel and B{\"o}ck, Sebastian and Klambauer, G{\"u}nter and Hochreiter, Sepp},
  journal={arXiv preprint arXiv:2607.01204},
  year={2026}
}

@article{auer2026tirex,
  title={Tirex: Zero-shot forecasting across long and short horizons with enhanced in-context learning},
  author={Auer, Andreas and Podest, Patrick and Klotz, Daniel and B{\"o}ck, Sebastian and Klambauer, G{\"u}nter and Hochreiter, Sepp},
  journal={Advances in Neural Information Processing Systems},
  volume={38},
  pages={57529--57580},
  year={2026}
}

@inproceedings{timesfm,
  title={A decoder-only foundation model for time-series forecasting},
  author={Das, Abhimanyu and Kong, Weihao and Sen, Rajat and Zhou, Yichen},
  booktitle={Forty-first International Conference on Machine Learning},
  year={2024}
}

@misc{google2025timesfm25,
  title={{TimesFM 2.5: 200M PyTorch checkpoint}},
  author={{Google Research}},
  year={2025},
  howpublished={Hugging Face},
  note={Model card and implementation}
}

@inproceedings{woo2024moirai,
  title={Unified Training of Universal Time Series Forecasting Transformers},
  author={Woo, Gerald and Liu, Chenghao and Kumar, Akshat and Xiong, Caiming and Savarese, Silvio and Sahoo, Doyen},
  booktitle={Forty-first International Conference on Machine Learning},
  year={2024}
}

@article{liu2025sundial,
  title={Sundial: A Family of Highly Capable Time Series Foundation Models},
  author={Liu, Yong and Qin, Guo and Shi, Zhiyuan and Chen, Zhi and Yang, Caiyin and Huang, Xiangdong and Wang, Jianmin and Long, Mingsheng},
  journal={arXiv preprint arXiv:2502.00816},
  year={2025}
}

@article{ansarichronos,
  title={Chronos: Learning the Language of Time Series},
  author={Ansari, Abdul Fatir and Stella, Lorenzo and Turkmen, Ali Caner and Zhang, Xiyuan and Mercado, Pedro and Shen, Huibin and Shchur, Oleksandr and Rangapuram, Syama Sundar and Arango, Sebastian Pineda and Kapoor, Shubham and others},
  journal={Transactions on Machine Learning Research},
  year={2024}
}

@article{timemoe,
  title={Time-MoE: Billion-Scale Time Series Foundation Models with Mixture of Experts},
  author={Shi, Xiaoming and Wang, Shiyu and Nie, Yuqi and Li, Dianqi and Ye, Zhou and Wen, Qingsong and Jin, Ming},
  journal={arXiv e-prints},
  pages={arXiv--2409},
  year={2024}
}

@inproceedings{timer,
  title={Timer: Generative Pre-trained Transformers Are Large Time Series Models},
  author={Liu, Yong and Zhang, Haoran and Li, Chenyu and Huang, Xiangdong and Wang, Jianmin and Long, Mingsheng},
  booktitle={International Conference on Machine Learning},
  pages={32369--32399},
  year={2024},
  organization={PMLR}
}

@inproceedings{cai2024msgnet,
  title={Msgnet: Learning multi-scale inter-series correlations for multivariate time series forecasting},
  author={Cai, Wanlin and Liang, Yuxuan and Liu, Xianggen and Feng, Jianshuai and Wu, Yuankai},
  booktitle={Proceedings of the AAAI Conference on Artificial Intelligence},
  volume={38},
  number={10},
  pages={11141--11149},
  year={2024}
}

@inproceedings{liu2023itransformer,
  author       = {Yong Liu and
                  Tengge Hu and
                  Haoran Zhang and
                  Haixu Wu and
                  Shiyu Wang and
                  Lintao Ma and
                  Mingsheng Long},
  title        = {iTransformer: Inverted Transformers Are Effective for Time Series
                  Forecasting},
  booktitle    = {ICLR},
  year         = {2024},
}

@inproceedings{qiu2024tfb,
title   = {{TFB}: Towards Comprehensive and Fair Benchmarking of Time Series Forecasting Methods},
author  = {Xiangfei Qiu and Jilin Hu and Lekui Zhou and Xingjian Wu and Junyang Du and Buang Zhang and Chenjuan Guo and Aoying Zhou and Christian S. Jensen and Zhenli Sheng and Bin Yang},
booktitle = {Proc. {VLDB} Endow.},
pages   = {2363--2377},
year    = {2024}
}

@inproceedings{chen2024pathformer,
  author       = {Peng Chen and
                  Yingying Zhang and
                  Yunyao Cheng and
                  Yang Shu and
                  Yihang Wang and
                  Qingsong Wen and
                  Bin Yang and
                  Chenjuan Guo},
  title        = {Pathformer: Multi-scale Transformers with Adaptive Pathways for Time
                  Series Forecasting},
  booktitle    = {ICLR},
  year         = {2024},
}

@inproceedings{wu2022timesnet,
  author       = {Haixu Wu and
                  Tengge Hu and
                  Yong Liu and
                  Hang Zhou and
                  Jianmin Wang and
                  Mingsheng Long},
  title        = {TimesNet: Temporal 2D-Variation Modeling for General Time Series Analysis},
  booktitle    = {ICLR},
  year         = {2023},
}

@article{das2023long,
  author       = {Abhimanyu Das and
                  Weihao Kong and
                  Andrew Leach and
                  Shaan Mathur and
                  Rajat Sen and
                  Rose Yu},
  title        = {Long-term Forecasting with TiDE: Time-series Dense Encoder},
  journal      = {Trans. Mach. Learn. Res.},
  volume       = {2023},
  year         = {2023},
}

@inproceedings{nie2022time,
  author       = {Yuqi Nie and
                  Nam H. Nguyen and
                  Phanwadee Sinthong and
                  Jayant Kalagnanam},
  title        = {A Time Series is Worth 64 Words: Long-term Forecasting with Transformers},
  booktitle    = {ICLR},
  year         = {2023},

}

@inproceedings{qiu2025duet,
title   = {{DUET}: Dual Clustering Enhanced Multivariate Time Series Forecasting},
author  = {Xiangfei Qiu and Xingjian Wu and Yan Lin and Chenjuan Guo and Jilin Hu and Bin Yang},
booktitle = {SIGKDD},
pages     = {1185-1196},
year    = {2025}
}

@inproceedings{zhang2022crossformer,
  title={Crossformer: Transformer utilizing cross-dimension dependency for multivariate time series forecasting},
  author={Zhang, Yunhao and Yan, Junchi},
  booktitle={{ICLR}},
  year={2022}
}

@inproceedings{zeng2023transformers,
  title={Are transformers effective for time series forecasting?},
  author={Zeng, Ailing and Chen, Muxi and Zhang, Lei and Xu, Qiang},
  booktitle={{AAAI}},
  volume={37},
  number={9},
  pages={11121--11128},
  year={2023}
}

@article{chen2024similarity,
  title={From Similarity to Superiority: Channel Clustering for Time Series Forecasting},
  author={Chen, Jialin and Lenssen, Jan Eric and Feng, Aosong and Hu, Weihua and Fey, Matthias and Tassiulas, Leandros and Leskovec, Jure and Ying, Rex},
  journal={arXiv preprint arXiv:2404.01340},
  year={2024}
}

@inproceedings{kim2021reversible,
  title={Reversible instance normalization for accurate time-series forecasting against distribution shift},
  author={Kim, Taesung and Kim, Jinhee and Tae, Yunwon and Park, Cheonbok and Choi, Jang-Ho and Choo, Jaegul},
  booktitle={ICLR},
  year={2021}
}

@inproceedings{qiu2025tab,
title   = {{TAB}: Unified Benchmarking of Time Series Anomaly Detection Methods},
author  = {Xiangfei Qiu and Zhe Li and Wanghui Qiu and Shiyan Hu and Lekui Zhou and Xingjian Wu and Zhengyu Li and Chenjuan Guo and Aoying Zhou and Zhenli Sheng and Jilin Hu and Christian S. Jensen and Bin Yang},
booktitle = {Proc. {VLDB} Endow.},
year    = {2025},
pages   = {2775--2789}
}

@inproceedings{wu2024catch,
  title={{CATCH}: Channel-Aware multivariate Time Series Anomaly Detection via Frequency Patching},
  author={Wu, Xingjian and Qiu, Xiangfei and Li, Zhengyu and Wang, Yihang and Hu, Jilin and Guo, Chenjuan and Xiong, Hui and Yang, Bin},
  booktitle={ICLR},
  year={2025}
}

@inproceedings{hu2025adaptive,
  title={Adaptive Multi-Scale Decomposition Framework for Time Series Forecasting},
  author={Hu, Yifan and Liu, Peiyuan and Zhu, Peng and Cheng, Dawei and Dai, Tao},
  booktitle={AAAI},
  year={2025}
}

@inproceedings{wang2025lightgts,
  title={LightGTS: A Lightweight General Time Series Forecasting Model},
  author={Wang, Yihang and Qiu, Yuying and Shu, Yang and Rao, Zhongwen and Pan, Lujia and Yang, Bin and Chenjuan, Guo},
  booktitle={ICML},
  year={2025}
}

@inproceedings{wang2025rose,
  title={Towards a General Time Series Forecasting Model with Unified Representation and Adaptive Transfer},
  author={Wang, Yihang and Qiu, Yuying and Zhao, kai and Shu, Yang and Rao, Zhongwen and Pan, Lujia and Yang, Bin and Chenjuan, Guo},
  booktitle={ICML},
  year={2025}
}

@inproceedings{qiu2025dag,
  title={{DAG}: A Dual Correlation Network for Time Series Forecasting with Exogenous Variables},
  author={Qiu, Xiangfei and Zhu, Yuhan and Li, Zhengyu and Wu, Xingjian and Yang, Bin and Hu, Jilin},
  booktitle={ICML},
  year={2026}
}

@article{shchur2025fev,
  title={fev-bench: A realistic benchmark for time series forecasting},
  author={Shchur, Oleksandr and Ansari, Abdul Fatir and Turkmen, Caner and Stella, Lorenzo and Erickson, Nick and Guerron, Pablo and Bohlke-Schneider, Michael and Wang, Yuyang},
  journal={arXiv preprint arXiv:2509.26468},
  year={2025}
}

@article{zhang2019root,
  title={Root mean square layer normalization},
  author={Zhang, Biao and Sennrich, Rico},
  journal={Advances in neural information processing systems},
  volume={32},
  year={2019}
}

@article{qiaotime,
  title={It's TIME: Towards the Next Generation of Time Series Forecasting Benchmarks},
  author={Qiao, Zhongzheng and Pan, Sheng and Wang, Anni and Zhukova, Viktoriya and Liu, Yong and Jiang, Xudong and Wen, Qingsong and Long, Mingsheng and Jin, Ming and Liu, Chenghao},
  journal={arXiv preprint arXiv:2602.12147},
  year={2026}
}

@article{ansari2025chronos2,
  title={Chronos-2: From univariate to universal forecasting},
  author={Ansari, Abdul Fatir and Shchur, Oleksandr and K{\"u}ken, Jaris and Auer, Andreas and Han, Boran and Mercado, Pedro and Rangapuram, Syama Sundar and Shen, Huibin and Stella, Lorenzo and Zhang, Xiyuan and others},
  journal={arXiv preprint arXiv:2510.15821},
  year={2025}
}

@article{liu2026timers1,
  title={Timer-s1: A billion-scale time series foundation model with serial scaling},
  author={Liu, Yong and Su, Xingjian and Wang, Shiyu and Zhang, Haoran and Liu, Haixuan and Wang, Yuxuan and Ye, Zhou and Xiang, Yang and Wang, Jianmin and Long, Mingsheng},
  journal={arXiv preprint arXiv:2603.04791},
  year={2026}
}

@article{khwaja2026toto,
  title={Toto 2.0: Time Series Forecasting Enters the Scaling Era},
  author={Khwaja, Emaad and Lettieri, Chris and Woo, Gerald and Belouadah, Eden and Cenac, Marc and Jarry, Guillaume and Paquin, Enguerrand and Zhao, Xunyi and Zhukov, Viktoriya and Abou-Amal, Othmane and others},
  journal={arXiv preprint arXiv:2605.20119},
  year={2026}
}

@article{liu2026falcon,
  title={Falcon-X: A Time Series Foundation Model for Heterogeneous Multivariate Modeling},
  author={Liu, Yiding and Hu, Yifan and Xia, Hongjie and Liu, Peiyuan and Chen, Hongzhou and Dai, Xilin and Dong, Zewei and Yang, Jiang-Ming},
  journal={arXiv preprint arXiv:2605.27286},
  year={2026}
}

@article{aksu2024gift,
  title={Gift-eval: A benchmark for general time series forecasting model evaluation},
  author={Aksu, Taha and Woo, Gerald and Liu, Juncheng and Liu, Xu and Liu, Chenghao and Savarese, Silvio and Xiong, Caiming and Sahoo, Doyen},
  journal={arXiv preprint arXiv:2410.10393},
  year={2024}
}

@article{sun2024learning,
  title={Learning pattern-specific experts for time series forecasting under patch-level distribution shift},
  author={Sun, Yanru and Xie, Zongxia and Eldele, Emadeldeen and Chen, Dongyue and Hu, Qinghua and Wu, Min},
  journal={arXiv preprint arXiv:2410.09836},
  year={2024}
}

@inproceedings{qiu2025DBLoss,
title   = {{DBLoss}: Decomposition-based Loss Function for Time Series Forecasting},
author  = {Xiangfei Qiu and Xingjian Wu and Hanyin Cheng and Xvyuan Liu and Chenjuan Guo and Jilin Hu and Bin Yang},
booktitle = {NeurIPS},
year    = {2025}
}

@article{wu2025flame,
  title={FLAME: Flow Enhanced Legendre Memory Models for General Time Series Forecasting},
  author={Wu, Xingjian and Cheng, Hanyin and Qiu, Xiangfei and Li, Zhengyu and Hu, Jilin and Guo, Chenjuan and Yang, Bin},
  journal={arXiv preprint arXiv:2512.14253},
  year={2025}
}

@inproceedings{liu2026apn,
  title={Rethinking Irregular Time Series Forecasting: A Simple yet Effective Baseline},
  author={Liu, Xvyuan and Qiu, Xiangfei and Wu, Xingjian and Li, Zhengyu and Guo, Chenjuan and Hu, Jilin and Yang, Bin},
  booktitle={AAAI},
  year={2026}
}

@inproceedings{ma2024followpose,
  title={Follow your pose: Pose-guided text-to-video generation using pose-free videos},
  author={Ma, Yue and He, Yingqing and Cun, Xiaodong and Wang, Xintao and Chen, Siran and Li, Xiu and Chen, Qifeng},
  booktitle={AAAI},
  volume={38},
  number={5},
  pages={4117--4125},
  year={2024}
}

@inproceedings{ma2024followyouremoji,
  title={Follow-your-emoji: Fine-controllable and expressive freestyle portrait animation},
  author={Ma, Yue and Liu, Hongyu and Wang, Hongfa and Pan, Heng and He, Yingqing and Yuan, Junkun and Zeng, Ailing and Cai, Chengfei and Shum, Heung-Yeung and Liu, Wei and others},
  booktitle={SIGGRAPH Asia},
  pages={1--12},
  year={2024}
}

@article{ma2025controllable,
  title={Controllable Video Generation: A Survey},
  author={Ma, Yue and Feng, Kunyu and Hu, Zhongyuan and Wang, Xinyu and Wang, Yucheng and Zheng, Mingzhe and He, Xuanhua and Zhu, Chenyang and Liu, Hongyu and He, Yingqing and others},
  journal={arXiv preprint arXiv:2507.16869},
  year={2025}
}

@inproceedings{yu2025merlin,
  title={Merlin: Multi-View Representation Learning for Robust Multivariate Time Series Forecasting with Unfixed Missing Rates},
  author={Yu, Chengqing and Wang, Fei and Yang, Chuanguang and Shao, Zezhi and Sun, Tao and Qian, Tangwen and Wei, Wei and An, Zhulin and Xu, Yongjun},
  booktitle={SIGKDD},
  pages={3633--3644},
  year={2025}
}

@article{cheng2026metagnsdformer,
  title={MetaGNSDformer: Meta-learning enhanced gated non-stationary informer with frequency-aware attention for point-interval remaining useful life prediction of lithium-ion batteries},
  author={Cheng, Fang and Liu, Hui and Lv, Xinwei},
  journal={Advanced Engineering Informatics},
  volume={69},
  pages={103798},
  year={2026},
}

@ARTICLE{11002729,
  author={Yu, Chengqing and Wang, Fei and Shao, Zezhi and Qian, Tangwen and Zhang, Zhao and Wei, Wei and An, Zhulin and Wang, Qi and Xu, Yongjun},
  journal={IEEE Transactions on Knowledge and Data Engineering}, 
  title={GinAR+: A Robust End-to-End Framework for Multivariate Time Series Forecasting With Missing Values}, 
  year={2025},
  volume={37},
  number={8},
  pages={4635-4648},
 }

@inproceedings{liu2026astgi,
  title={{ASTGI}: Adaptive Spatio-Temporal Graph Interactions for Irregular Multivariate Time Series Forecasting},
  author={Liu, Xvyuan and Qiu, Xiangfei and Cheng, Hanyin and Wu, Xingjian and Guo, Chenjuan and Yang, Bin and Hu, Jilin},
  booktitle={ICLR},
  year={2026}
}

@inproceedings{qiu2026bridging,
  title={Bridging Time and Frequency: A Joint Modeling Framework for Irregular Multivariate Time Series Forecasting},
  author={Qiu, Xiangfei and Yan, Kangjia and Liu, Xvyuan and Wu, Xingjian and Hu, Jilin},
  booktitle={ICML},
  year={2026}
}

@inproceedings{li2026gcgnet,
  title={{GCGNet}: Graph-consistent generative network for time series forecasting with exogenous variables},
  author={Li, Zhengyu and Qiu, Xiangfei and Zhu, Yuhan and Wu, Xingjian and Hu, Jilin and Guo, Chenjuan and Yang, Bin},
  booktitle={ICLR},
  year={2026}
}

@article{wu2026timeart,
  title={{TimeART}: Towards agentic time series reasoning via tool-augmentation},
  author={Wu, Xingjian and Lu, Junkai and Li, Zhengyu and Qiu, Xiangfei and Hu, Jilin and Guo, Chenjuan and Jensen, Christian S and Yang, Bin},
  journal={arXiv preprint arXiv:2601.13653},
  year={2026}
}

@article{ma2026livelight,
  title={LiveLight: Real-time Streaming Video Relighting with Interactive Control},
  author={Ma, Yue and Wang, Jiangming and Wang, Yucheng and Wang, Xilai and Li, Zhiyuan and Wang, Xinyu and Liu, Hongyu and Liang, Ruofan and Zhang, Songchun and Xue, Yuxuan and others},
  journal={arXiv preprint arXiv:2608.01771},
  year={2026}
}

@inproceedings{cheng2026star,
  title={{STAR}: Boosting Time Series Foundation Models for Anomaly Detection Through State-Aware Adapter},
  author={Hanyin Cheng and Ruitong Zhang and Yuning Lu and Peng Chen and Meng Wang and Yang Shu and Bin Yang and Chenjuan Guo},
  booktitle={NeurIPS},
  year={2026}
}

@inproceedings{cheng2026ccd,
  title={{CCD}: Capturing Cross-Correlations with Deformable Convolutional Networks for Multivariate Time Series Forecasting},
  author={Hanyin Cheng and Xingjian Wu and Xiangfei Qiu and Yang Shu and Bin Yang and Chenjuan Guo},
  booktitle={KDD},
  year={2026}
}

@inproceedings{cheng2026kite,
  title={{KITE}: Knowledge-Guided Probabilistic Modeling for Time Series Forecasting with Exogenous Variables},
  author={Hanyin Cheng and Jingrong Zhou and Yang Shu and Chenjuan Guo},
  booktitle={ICML},
  year={2026}
}

@inproceedings{cheng2026cora,
  title={{CoRA}: Boosting Time Series Foundation Models for Multivariate Forecasting through Correlation-aware Adapter},
  author={Hanyin Cheng and Xingjian Wu and Yang Shu and Zhongwen Rao and Lujia Pan and Bin Yang and Chenjuan Guo},
  booktitle={ICLR},
  year={2026}
}
\bibliographystyle{iclr2027-conference}

\clearpage
\appendix
\section{Experimental Settings and Additional Results}
\label{app:main-results}

\subsection{Benchmarks and Evaluation Protocol}
\label{app:result-protocol}
\label{sec:experimental-settings}

We evaluate Aurora-X on five benchmarks covering general forecasting across domains, multivariate prediction, and forecasting with future covariates. Together, they assess transfer across temporal patterns and the ability to exploit different forms of predictive information.

\textbf{GIFT-Eval}~\citep{aksu2024gift} evaluates general forecasting across diverse domains, sampling frequencies, and prediction lengths. We use its 97 configurations, whose short-, medium-, and long-term tasks assess forecasting under heterogeneous temporal scales.

\textbf{TIME}~\citep{qiaotime} is a task-centric benchmark for zero-shot forecasting on newly collected data. It contains 98 tasks from 50 datasets across eight domains, with task configurations designed around application requirements and variate predictability. Its feature-based analyses further characterize performance across temporal patterns.

\textbf{FEV-Bench}~\citep{shchur2025fev} comprises 100 forecasting tasks from 96 datasets across seven domains. It includes univariate and multivariate targets, static attributes, past-only dynamic covariates, and known future covariates, providing varied input structures for assessing general forecasting models.

\textbf{TFB}~\citep{qiu2024tfb} provides a unified evaluation framework for univariate and multivariate forecasting across diverse temporal characteristics. We use eight multivariate datasets: ETTh1, ETTh2, ETTm1, ETTm2, Weather, Solar, Electricity, and Traffic, with prediction lengths of 96, 192, 336, and 720.

\textbf{DAG-Bench}~\citep{qiu2025dag} refers to the twelve real-world covariate-forecasting datasets evaluated in DAG. They cover electricity prices, energy generation, reservoir water levels, and wind power. Forecast horizons are 24 and 360 for ten datasets, and 10 and 30 for Colbun and Rapel. These tasks evaluate how models use future exogenous information to improve target forecasts.

\paragraph{Evaluation protocol.}
We follow the benchmark-defined forecasting tasks, data splits, prediction horizons, and metric definitions. Each benchmark is evaluated under its corresponding input setting, including the availability of future covariates.

\paragraph{Reported aggregation and input configurations.}
Relative MASE is the geometric mean of configuration-wise ratios to Seasonal Naive. FEV-Bench uses the recorded task-level ratios directly, without leaderboard clipping, failure imputation, or training-overlap substitutions. TFB and DAG-Bench report arithmetic horizon averages of MSE and MAE on normalized series; aggregate comparisons weight datasets equally. Means are computed from the reported three-decimal horizon-specific scores, and percentage improvements are calculated before display rounding.

On TFB, Aurora-X uses 2,880 history points and 48 points per patch, except on Solar, where the patch span is 144. The eight baselines taken from the SEER table use 96 history points; other baselines retain their published configurations. GIFT-Eval and TIME use dataset-specific inference spans selected with benchmark feedback. On DAG-Bench, models receive future covariates through the benchmark's input adapters.

\subsection{Baselines and Result Sources}
\label{app:baseline-selection}

\paragraph{Time series foundation models.}
We compare Aurora-X with pretrained forecasters spanning different architectures and probabilistic prediction strategies. Timer-S1~\citep{liu2026timers1} combines sparse MoE blocks with serial-token prediction. Chronos-2~\citep{ansari2025chronos2} uses group attention to share information among related series, target variables, and covariates. TiRex~\citep{auer2026tirex} employs an xLSTM backbone for recurrent state tracking, while TiRex-2~\citep{podest2026tirex2} extends this approach to multivariate inputs, future covariates, and streaming inference. Falcon-X~\citep{liu2026falcon} models heterogeneous variables through a shared latent prototype space. The Xihe family~\citep{sun2025xihe} captures local and global temporal dependencies with hierarchical interleaved block attention; our comparison includes Xihe-Ultra. Sundial~\citep{liu2025sundial} learns predictive distributions through flow matching, and Toto-2.0~\citep{khwaja2026toto} provides forecasting models at multiple parameter scales. We also include TimesFM-2.5, Moirai-2.0, Falcon-2.0, FlowState, CITRAS-FM, Chronos-Bolt, Toto-1.0, PatchTST-FM-Extended, and TS-ICL. Evaluated versions, sizes, and checkpoint settings are identified in the results.

\paragraph{Supervised multivariate forecasting.}
The TFB comparison covers fourteen supervised baselines. DUET~\citep{qiu2025duet} combines temporal and channel clustering, while AMD~\citep{hu2025adaptive} models temporal and channel dependencies through adaptive multiscale decomposition. SRSNet~\citep{wu2025srsnet} adaptively selects and reassembles informative temporal patches. PatchTST~\citep{nie2022time} uses channel-independent patch tokens; iTransformer~\citep{liu2023itransformer} attends over variate tokens, and Crossformer~\citep{zhang2022crossformer} separates temporal and cross-dimension attention. MSGNet~\citep{cai2024msgnet} learns inter-series correlations at multiple temporal scales using adaptive graph convolution. TimesNet~\citep{wu2022timesnet} represents periodic variation in two-dimensional tensors, and DLinear~\citep{zeng2023transformers} provides a linear forecasting baseline. TimeKAN, TimePro, xPatch, Amplifier, and Fredformer complete the comparison.

\paragraph{Forecasting with covariates.}
DAG-Bench includes ten supervised baselines. DAG~\citep{qiu2025dag} models temporal and channel relationships between endogenous and exogenous variables through dual correlation networks. TimeXer~\citep{wang2024timexer} combines patch-wise self-attention with variate-wise cross-attention to incorporate exogenous information. TiDE~\citep{das2023long} uses an MLP encoder--decoder to model temporal dependencies and covariates. We additionally compare with GCGNet, TFT, CrossLinear, DUET, Amplifier, TimeKAN, and PatchTST under the benchmark's covariate-input setting.

\paragraph{Model selection and result sources.}
The main MASE figure compares individual TSFMs and excludes benchmark-specific fine-tuning, ensembles, and tabular foundation-model adaptations. Its GIFT-Eval panel requires complete recorded coverage and no disclosed test overlap; expanded rankings include additional checkpoints, with fine-tuned variants marked explicitly. TIME includes the reported Toto-2.0 sizes and PatchTST-FM-Extended; PatchTST-FM-R1 and OmniScient are excluded. For duplicated TFB baselines, we use the SEER table for DUET, Fredformer, iTransformer, PatchTST, SRSNet, xPatch, Amplifier, and DLinear; the other six methods use the CCD table. CCD and SEER are excluded. DAG-Bench retains all ten published baselines. Each table ranks its displayed models separately; red boldface and blue underlining mark the best and second-best distinct scores, including all ties at three-decimal precision.

\subsection{Model Configuration}
\label{app:model-configuration}

Aurora-X has approximately one billion total parameters. Its backbone comprises 12 Time--Group blocks with a hidden width of 768 and a native patch length of 48. Each MoE layer combines one shared expert with 16 routed experts, selecting four routed experts per token. Group attention, future time encoding, and the arcsinh transformation are enabled. Table~\ref{tab:model-configuration} summarizes the model configuration.

\begin{table}[!ht]
    \centering
    \caption{Aurora-X model configuration. Parameter count denotes total model size; expert counts are per MoE layer.}
    \label{tab:model-configuration}
    \begingroup
    \small
    \renewcommand{\arraystretch}{1.05}
    \begin{tabularx}{\linewidth}{@{}l X r@{}}
        \toprule
        \textbf{Component} & \textbf{Setting} & \textbf{Value} \\
        \midrule
        Model scale & Total parameters & 1B \\
        \midrule
        \multirow{10}{*}{Backbone}
        & Time--Group blocks & 12 \\
        & Hidden width & 768 \\
        & Attention heads & 12 \\
        & Feed-forward hidden width & 3,072 \\
        & Native patch length & 48 \\
        & Backbone and expert activation & ReLU \\
        & Dropout rate & 0.1 \\
        & Time-encoding scale & 8,160 \\
        & RoPE base & 10,000 \\
        & Initialization factor & 0.05 \\
        \midrule
        \multirow{7}{*}{Mixture of experts}
        & Routed experts & 16 \\
        & Shared experts & 1 \\
        & Active routed experts per token & 4 \\
        & Expert hidden width & 3,072 \\
        & Router dimension & 768 \\
        & Shallow-pattern dimension & 32 \\
        & Load balancing & Switch-style \\
        \midrule
        \multirow{5}{*}{Implicit quantile head}
        & Residual blocks & 4 \\
        & Cosine features & 128 \\
        & Quantile samples per replica & 20 \\
        & Target/condition replication factor & 5 \\
        & Head activation & SiLU \\
        \midrule
        \multirow{5}{*}{MoE regularization}
        & Pattern-loss coefficient $\lambda_{\mathrm{pat}}$ & 0.01 \\
        & Pattern-similarity bandwidth $\sigma_{\mathrm{pat}}$ & 1.0 \\
        & Relative orthogonality coefficient $\lambda_{\mathrm{orth}}$ & 0.1 \\
        & Load-balancing coefficient $\lambda_{\mathrm{bal}}$ & 0.001 \\
        & Expert-bias update rate & 0.0 \\
        \bottomrule
    \end{tabularx}
    \endgroup
\end{table}

For IQN training, each target patch and its conditioning state are replicated five times, with 20 independently sampled quantile levels per replica. This gives $Q=100$ quantile samples per original target patch in Equation~(\ref{eq:iqn-loss}); inference can query arbitrary continuous quantile levels. The regularization coefficients follow the nested weighting in Equation~(\ref{eq:total-loss}), with the orthogonality penalty multiplied by both $\lambda_{\mathrm{pat}}$ and $\lambda_{\mathrm{orth}}$. The router uses a Switch-style auxiliary load-balancing loss; the separate expert-bias update rate is zero.

The serialized configuration records \texttt{float32} as its dtype and Transformers version 4.50.1.

\subsection{Training Data and Five-Stage Curriculum}
\label{app:training-settings}

\paragraph{Training corpora.}
Pretraining and midtraining use the open-source TimeBench corpus adopted by Sundial~\citep{liu2025sundial} and Timer-S1~\citep{liu2026timers1}. It brings together heterogeneous time series from domains including meteorology, healthcare, finance, and the Internet of Things. For post-training, we construct a subset of one billion original time points from the pretraining corpora used by LightGTS~\citep{wang2025lightgts} and ROSE~\citep{wang2025rose}. This size refers to the source observations before multi-scale augmentation.

\paragraph{Training resources and schedule.}
We train Aurora-X using 64 NVIDIA H800 GPUs and AdamW with an initial learning rate of $5\times10^{-5}$, cosine annealing, and FP32 precision. We use a per-GPU batch size of 128 and accumulate gradients over three steps, yielding an effective batch size of 24,576. The curriculum consists of pretraining (Pre), three midtraining stages (M1--M3), and post-training (Post), with each stage continuing from the preceding checkpoint. Table~\ref{tab:training-schedule} summarizes the 160K optimization steps across all five stages.

\begin{table}[!htbp]
    \centering
    \caption{Optimization steps across the five training stages of Aurora-X. K denotes 1,000 steps.}
    \label{tab:training-schedule}
    \begin{tabularx}{\linewidth}{@{}l X r@{}}
        \toprule
        \textbf{Stage} & \textbf{Training focus} & \textbf{Steps} \\
        \midrule
        Pre & Channel-independent temporal modeling & 50K \\
        M1 & Cross-variable modeling with extended contexts and horizons & 25K \\
        M2 & Adaptation to varied context lengths and forecast horizons & 25K \\
        M3 & Known future covariate conditioning & 50K \\
        Post & Variable-resolution forecasting & 10K \\
        \midrule
        \textbf{Total} & & \textbf{160K} \\
        \bottomrule
    \end{tabularx}
\end{table}

\textbf{Pre: temporal pattern modeling.} We begin with channel-independent forecasting on TimeBench. Each variable is modeled through its own historical observations, allowing the shared backbone to learn temporal patterns across heterogeneous sources before introducing cross-variable interactions.

\textbf{M1: cross-variable dependency modeling.} We introduce jointly modeled variables and extend the context and forecast horizon. Group attention can then combine each target's temporal history with information from related variables, developing multivariate forecasting capabilities.

\textbf{M2: context and horizon adaptation.} We vary historical context lengths and forecast horizons during training. This exposes the same model to different amounts of conditioning information and prediction demands, supporting flexible forecasting configurations.

\textbf{M3: future covariate conditioning.} We introduce known future covariates alongside target histories. Available covariate values condition the forecast, while prediction losses are evaluated only on target variables. This stage teaches the model to use auxiliary information extending into the forecast horizon.

\textbf{Post: variable-resolution forecasting.} We augment the one-billion-point subset by enumerating temporal spans and resampling each block to the native patch length. Histories, covariates, and targets undergo aligned transformations, with supervision evaluated on the resampled targets. This stage develops the ability to adjust temporal resolution at inference, controlling historical coverage and context-token usage with fixed model weights.

Pre and M1--M3 use the quantile objective and MoE regularizers in Equation~(\ref{eq:total-loss}). Post uses the resampled-target objective in Equation~(\ref{eq:post-training-loss}) with the same regularizers. The construction of paired resolution augmentations and their alignment with inference are detailed in Appendix~\ref{app:resolution-equivalence}.

\subsection{Architecture Comparison Protocol}
\label{app:architecture-protocol}

\paragraph{Shared pretraining setting.}
The architecture studies compare module implementations within Aurora-X, using our pretraining corpus. The protocol matches training exposure, backbone dimensions, and evaluation configurations. All four MoE variants use 16 routed experts with top-4 selection per token and expert width matched to Aurora-X; shared-expert configurations and module-specific objectives are recorded for each variant. The head comparison retains the Aurora-X MoE and changes only the forecasting head and its associated objective. These comparisons use the pretraining checkpoint, while the training-stage analysis follows the full curriculum.

\paragraph{Routing across distinct inputs.}
We follow the characteristic-based selection in TFB~\citep{qiu2024tfb}, Section~5.2.3, which identifies a representative dataset with the highest score for each characteristic. Table~\ref{tab:routing-scenarios} lists the five temporal scenarios used here. These descriptors identify prominent properties, rather than mutually exclusive dataset classes. TFB's transition score describes symbolic transition regularity; shifting captures changes in the value distribution.

\begin{table}[!htbp]
    \centering
    \caption{Routing scenarios following TFB's characteristic analysis. Each dataset is TFB's highest-scoring representative for the corresponding characteristic.}
    \label{tab:routing-scenarios}
    \begin{tabular}{@{}ll@{}}
        \toprule
        \textbf{Characteristic} & \textbf{Dataset} \\
        \midrule
        Trend & FRED-MD \\
        Seasonality & Electricity \\
        Shifting & NYSE \\
        Stationarity & Solar \\
        Transition & PEMS08 \\
        \bottomrule
    \end{tabular}
\end{table}

To isolate temporal routing behavior, we sample channel-wise context windows at the native patch resolution from held-out data. We fix window length and sample equal numbers of valid context patches per dataset, selecting channels and windows uniformly. All variants receive identical probe windows, and routing is evaluated at layer 6 of the 12-layer backbone; future-token routing is excluded. This intermediate-layer probe examines whether distinct input distributions retain different expert preferences after multiple representation transformations. Neither window selection nor dataset assignment depends on a learned router or its shallow-pattern clusters. Let $c_{s,e}$ count selections of expert $e$ in probe window $s$. For a dataset group $\mathcal{G}$, the heatmap reports
\[
P_{\mathcal{G},e}=\frac{1}{|\mathcal{G}|}\sum_{s\in\mathcal{G}}
\frac{c_{s,e}}{\sum_{e'}c_{s,e'}}.
\]
Each sequence receives equal weight; rows sum to one, and darker shading indicates higher normalized expert-selection frequency. Rows describe input-dependent expert preferences; their group-balanced average describes overall utilization. Expert indices are local to each variant, and shared experts are excluded.

\paragraph{Forecasting evaluation.}
All variants use common benchmark configurations and relative MASE aggregation. Head comparisons use a common evaluation quantile grid containing the median. IQN and the fixed quantile head predict these levels directly; distributional heads estimate them from their predictive distributions, with the extraction procedure and inference budget recorded per run.

\begin{figure}[p]
    \centering
    \includegraphics[width=\linewidth,height=0.88\textheight,keepaspectratio]{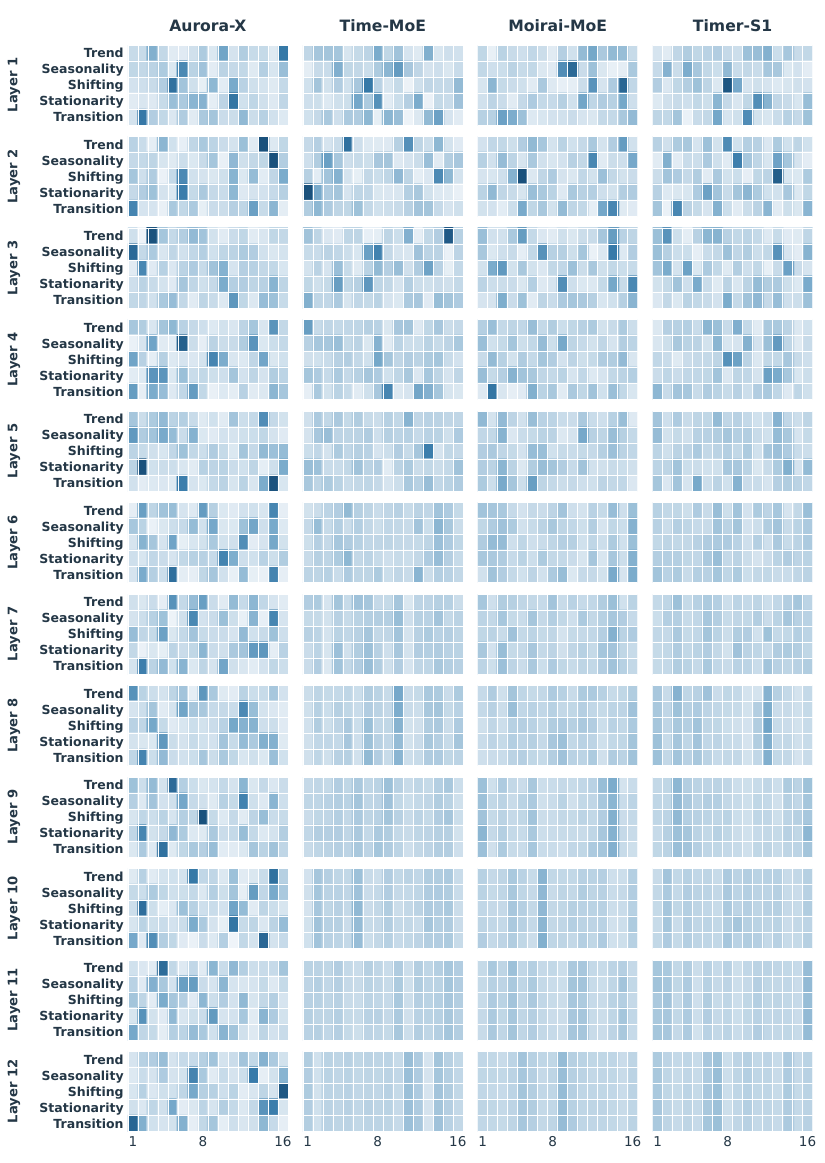}
    \caption{Routing profiles across all 12 layers. Each layer contains four model panels, with five scenario rows and 16 expert columns per panel. Rows sum to one; darker shading indicates higher selection frequency on a shared color scale. Layer 6 matches Figure~\ref{fig:moe-routing}.}
    \label{fig:moe-routing-all-layers}
\end{figure}

\paragraph{Routing across depth.}
Figure~\ref{fig:moe-routing-all-layers} presents routing profiles across all 12 layers, with layer 6 reproducing the profiles in Figure~\ref{fig:moe-routing}. The layer-wise panels compare hotspot locations, scenario overlap, and background utilization across depth. Within these panels, Aurora-X has distinct scenario profiles at every layer. The other three implementations show clear scenario differences in layers 1--3; these differences weaken from layer 4 onward. Their deeper profiles become similar across scenarios within each layer, while the preferred experts can still differ between layers. All panels use a common color scale. Expert indices are local to each layer and variant, so matching column positions across layers do not imply the same learned expert function.

\clearpage
\clearpage

\FloatBarrier

\begin{figure}[!htbp]
    \centering
    \includegraphics[width=\linewidth]{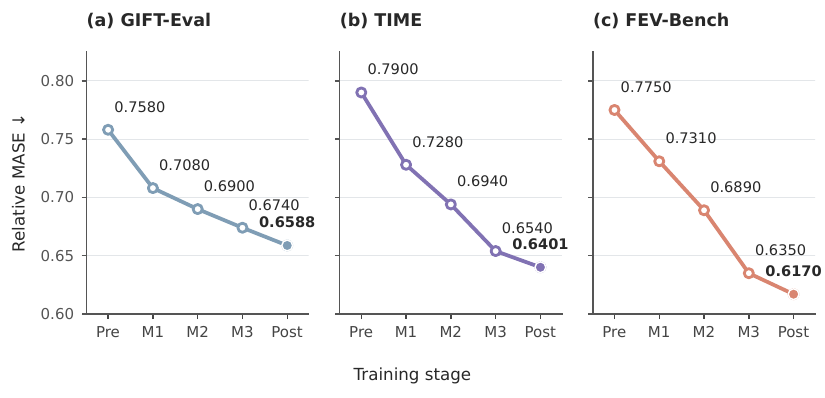}
    \caption{Relative MASE across successive Aurora-X training stages; lower is better. Pre and Post denote pretraining and post-training, while M1--M3 denote successive midtraining stages. All panels use the same vertical scale.}
    \label{fig:training-stages}
\end{figure}

\begin{samepage}
\subsection{Performance across Training Stages}
\label{app:training-stages}

Figure~\ref{fig:training-stages} tracks forecasting accuracy as Aurora-X progresses through the curriculum. Relative MASE decreases at every stage on all three benchmarks. From pretraining to post-training, the reductions are 13.1\% on GIFT-Eval, 19.0\% on TIME, and 20.4\% on FEV-Bench. The largest absolute decrease occurs from Pre to M1 on GIFT-Eval and TIME, and from M2 to M3 on FEV-Bench. Post-training further reduces relative MASE by 2.3\%, 2.1\%, and 2.8\% compared with M3, respectively, showing continued gains after midtraining.
\par
\end{samepage}

\FloatBarrier

\subsection{Expanded Probabilistic Forecasting Rankings}
\label{app:point-probabilistic}

\begin{figure}[!htbp]
    \centering
    \includegraphics[width=\linewidth]{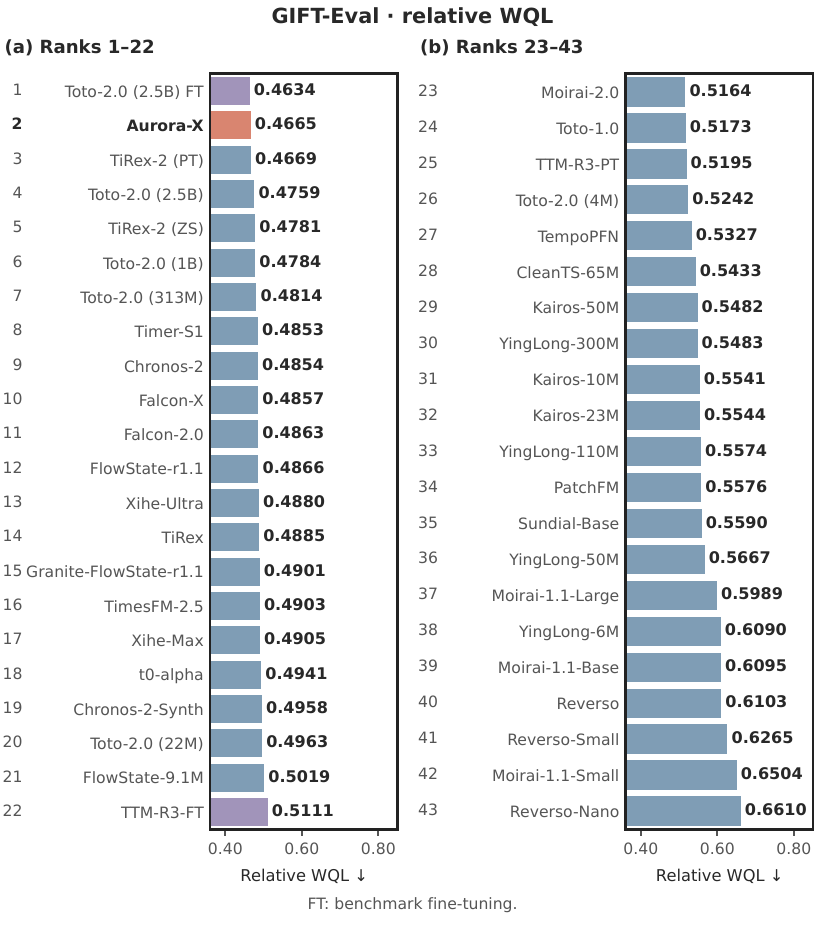}
    \caption{Expanded WQL ranking on GIFT-Eval. The two panels show a single ranking of 43 models and checkpoints. Lavender bars and FT labels identify fine-tuned variants. Scores use four decimal places; lower is better.}
    \label{fig:gift-full-results}
\end{figure}

\begin{figure}[!htbp]
    \centering
    \includegraphics[width=\linewidth]{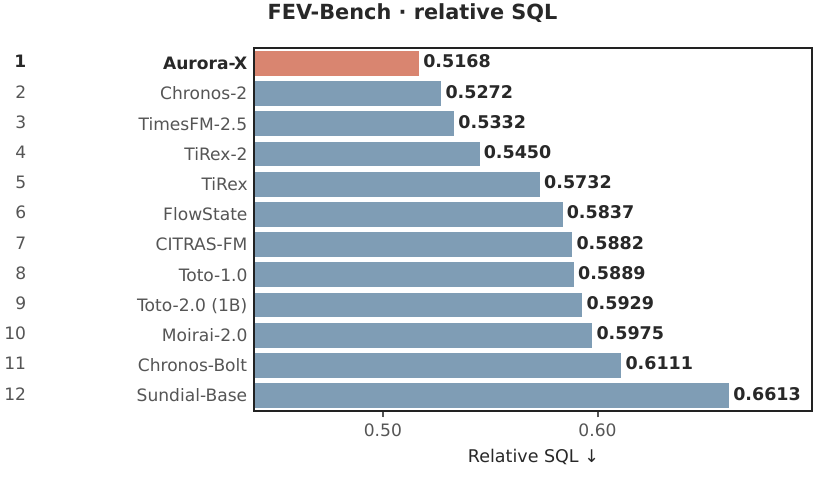}
    \caption{Relative SQL of twelve TSFMs on FEV-Bench. Scores use four decimal places; lower is better.}
    \label{fig:fev-full-results}
\end{figure}

Figures~\ref{fig:gift-full-results} and \ref{fig:fev-full-results} complement the main MASE comparison with expanded probabilistic rankings. GIFT-Eval uses weighted quantile loss (WQL), while FEV-Bench uses scaled quantile loss (SQL), normalized by in-sample seasonal differences. We report geometric means of model-to-Seasonal-Naive metric ratios; lower values are better. Models must have complete recorded MASE coverage and an available probabilistic score.

The GIFT-Eval ranking includes 43 models and checkpoints, with the two fine-tuned variants explicitly distinguished. Aurora-X obtains relative WQL of 0.4665, compared with 0.4669 for TiRex-2 (PT) and 0.4634 for the fine-tuned Toto-2.0 (2.5B) checkpoint. On FEV-Bench, Aurora-X achieves relative SQL of 0.5168, the lowest among the twelve displayed models. Different model sizes and training settings are listed separately.

\clearpage

\subsection{Multivariate and Covariate Forecasting}
\label{app:multivariate-covariate}
\label{sec:multivariate-forecasting}
\label{sec:covariate-forecasting}

Tables~\ref{tab:tfb-main} and \ref{tab:dag-main} report multivariate forecasting on TFB and forecasting with known future covariates on DAG-Bench, using task-specific supervised models as baselines. On TFB, Aurora-X reduces average MSE by 9.0\% and MAE by 3.6\% relative to DUET. It obtains the lowest horizon-averaged MSE on seven of the eight datasets and the lowest MAE on five. On DAG-Bench, Aurora-X reduces average MSE by 11.5\% and MAE by 7.7\% relative to DAG, ranking first in MSE on nine of the twelve datasets and in MAE on six. Its performance is particularly strong on the four SDWPF subsets, where it achieves the lowest values for both metrics. These rankings use the complete baseline sets in Appendices~\ref{app:tfb-full} and \ref{app:dag-full}. The results demonstrate that the shared forecasting model remains competitive with supervised predictors in both settings.

\begin{table}[!htbp]
\centering
\caption{Aurora-X is evaluated against recent advanced supervised baselines on TFB. MSE and MAE are averaged over prediction horizons of $96$, $192$, $336$, and $720$ steps. Lower values are better. Red boldface and blue underlining denote the best and second-best distinct values within this table, including all ties at the displayed precision. Differences in context settings are detailed in Appendix~\ref{app:result-protocol}, and full results are provided in Appendix~\ref{app:tfb-full}.}
\label{tab:tfb-main}
\begingroup
\small
\setlength{\tabcolsep}{2.4pt}
\renewcommand{\arraystretch}{1.08}
\sbox0{%
\begin{tabular}{@{}lcccccccccccccccc@{}}
\toprule
\textbf{Dataset} & \multicolumn{2}{c}{\textbf{Aurora-X}} & \multicolumn{2}{c}{\textbf{DUET}} & \multicolumn{2}{c}{\textbf{TimeKAN}} & \multicolumn{2}{c}{\textbf{AMD}} & \multicolumn{2}{c}{\textbf{TimePro}} & \multicolumn{2}{c}{\textbf{SRSNet}} & \multicolumn{2}{c}{\textbf{xPatch}} & \multicolumn{2}{c}{\textbf{Amplifier}} \\
 & MSE & MAE & MSE & MAE & MSE & MAE & MSE & MAE & MSE & MAE & MSE & MAE & MSE & MAE & MSE & MAE \\
\midrule
ETTh1 & \textcolor{red}{\textbf{0.390}} & \textcolor{red}{\textbf{0.411}} & 0.443 & 0.436 & \textcolor{blue}{\underline{0.431}} & \textcolor{blue}{\underline{0.429}} & 0.439 & 0.430 & 0.438 & 0.438 & 0.442 & 0.434 & 0.444 & \textcolor{blue}{\underline{0.429}} & 0.449 & 0.437 \\
ETTh2 & \textcolor{red}{\textbf{0.353}} & \textcolor{red}{\textbf{0.380}} & 0.372 & 0.398 & 0.391 & 0.412 & 0.371 & 0.398 & 0.377 & 0.404 & 0.376 & 0.402 & \textcolor{blue}{\underline{0.369}} & \textcolor{blue}{\underline{0.392}} & 0.389 & 0.411 \\
ETTm1 & \textcolor{red}{\textbf{0.335}} & \textcolor{red}{\textbf{0.364}} & 0.390 & 0.393 & 0.384 & 0.401 & 0.389 & 0.396 & 0.391 & 0.400 & 0.385 & 0.395 & 0.387 & \textcolor{blue}{\underline{0.383}} & \textcolor{blue}{\underline{0.383}} & 0.396 \\
ETTm2 & \textcolor{red}{\textbf{0.268}} & \textcolor{red}{\textbf{0.313}} & 0.280 & 0.324 & 0.282 & 0.330 & \textcolor{blue}{\underline{0.278}} & 0.323 & 0.281 & 0.326 & 0.284 & 0.330 & 0.280 & \textcolor{blue}{\underline{0.319}} & 0.280 & 0.326 \\
Weather & \textcolor{red}{\textbf{0.211}} & \textcolor{red}{\textbf{0.245}} & 0.251 & 0.273 & \textcolor{blue}{\underline{0.245}} & 0.273 & 0.254 & 0.280 & 0.252 & 0.276 & 0.250 & 0.277 & 0.247 & \textcolor{blue}{\underline{0.266}} & 0.247 & 0.275 \\
Solar & \textcolor{red}{\textbf{0.222}} & \textcolor{blue}{\underline{0.242}} & 0.237 & \textcolor{red}{\textbf{0.233}} & 0.255 & 0.282 & 0.253 & 0.281 & \textcolor{blue}{\underline{0.233}} & 0.267 & 0.250 & 0.281 & 0.277 & 0.268 & \textcolor{red}{\textbf{0.222}} & 0.256 \\
Electricity & \textcolor{blue}{\underline{0.170}} & 0.266 & 0.172 & \textcolor{red}{\textbf{0.259}} & 0.197 & 0.286 & 0.187 & 0.281 & \textcolor{red}{\textbf{0.169}} & \textcolor{blue}{\underline{0.263}} & 0.189 & 0.276 & 0.184 & 0.266 & 0.174 & 0.268 \\
Traffic & \textcolor{red}{\textbf{0.413}} & \textcolor{blue}{\underline{0.270}} & 0.451 & \textcolor{red}{\textbf{0.269}} & 0.455 & 0.318 & 0.500 & 0.324 & \textcolor{blue}{\underline{0.447}} & 0.302 & 0.495 & 0.307 & 0.500 & 0.283 & 0.485 & 0.315 \\
\bottomrule
\end{tabular}%
}
\ifdim\wd0>\linewidth\resizebox{\linewidth}{!}{\usebox0}\else\usebox0\fi
\endgroup
\end{table}

\begin{table}[!htbp]
\centering
\caption{Aurora-X is compared with supervised covariate-aware baselines on DAG-Bench. MSE and MAE are averaged over prediction horizons of $24$ and $360$ steps, or $10$ and $30$ steps for Colbun and Rapel. Lower values are better. Red boldface and blue underlining denote the best and second-best distinct values within this table, including all ties at the displayed precision. Full results for each horizon are reported in Appendix~\ref{app:dag-full}.}
\label{tab:dag-main}
\begingroup
\small
\setlength{\tabcolsep}{2.4pt}
\renewcommand{\arraystretch}{1.08}
\sbox0{%
\begin{tabular}{@{}lcccccccccccccccc@{}}
\toprule
\textbf{Dataset} & \multicolumn{2}{c}{\textbf{Aurora-X}} & \multicolumn{2}{c}{\textbf{DAG}} & \multicolumn{2}{c}{\textbf{GCGNet}} & \multicolumn{2}{c}{\textbf{TimeXer}} & \multicolumn{2}{c}{\textbf{TFT}} & \multicolumn{2}{c}{\textbf{DUET}} & \multicolumn{2}{c}{\textbf{Amplifier}} & \multicolumn{2}{c}{\textbf{TimeKAN}} \\
 & MSE & MAE & MSE & MAE & MSE & MAE & MSE & MAE & MSE & MAE & MSE & MAE & MSE & MAE & MSE & MAE \\
\midrule
NP & \textcolor{red}{\textbf{0.359}} & \textcolor{blue}{\underline{0.346}} & \textcolor{blue}{\underline{0.362}} & \textcolor{red}{\textbf{0.344}} & 0.370 & 0.348 & 0.418 & 0.371 & 0.379 & 0.375 & 0.411 & 0.408 & 0.420 & 0.419 & 0.406 & 0.420 \\
PJM & 0.112 & 0.204 & \textcolor{red}{\textbf{0.094}} & \textcolor{red}{\textbf{0.181}} & \textcolor{blue}{\underline{0.095}} & \textcolor{blue}{\underline{0.187}} & 0.108 & 0.199 & 0.114 & 0.207 & 0.102 & 0.197 & 0.137 & 0.247 & 0.139 & 0.263 \\
BE & \textcolor{red}{\textbf{0.413}} & \textcolor{blue}{\underline{0.283}} & \textcolor{blue}{\underline{0.423}} & \textcolor{red}{\textbf{0.280}} & 0.431 & 0.294 & 0.452 & 0.290 & 0.454 & 0.291 & 0.515 & 0.354 & 0.559 & 0.413 & 0.548 & 0.407 \\
FR & \textcolor{red}{\textbf{0.411}} & \textcolor{blue}{\underline{0.225}} & \textcolor{blue}{\underline{0.414}} & \textcolor{red}{\textbf{0.220}} & 0.415 & 0.234 & 0.428 & 0.241 & 0.504 & 0.257 & 0.496 & 0.327 & 0.554 & 0.408 & 0.548 & 0.374 \\
DE & 0.427 & 0.409 & \textcolor{red}{\textbf{0.370}} & \textcolor{red}{\textbf{0.370}} & \textcolor{blue}{\underline{0.402}} & \textcolor{blue}{\underline{0.389}} & 0.475 & 0.418 & 0.490 & 0.446 & 0.483 & 0.430 & 0.473 & 0.441 & 0.473 & 0.446 \\
Energy & 0.197 & 0.343 & \textcolor{red}{\textbf{0.124}} & \textcolor{red}{\textbf{0.268}} & 0.132 & \textcolor{blue}{\underline{0.278}} & 0.163 & 0.315 & \textcolor{blue}{\underline{0.130}} & 0.283 & 0.203 & 0.368 & 0.233 & 0.389 & 0.219 & 0.381 \\
Sdwpfm1 & \textcolor{red}{\textbf{0.330}} & \textcolor{red}{\textbf{0.388}} & \textcolor{blue}{\underline{0.423}} & 0.461 & 0.424 & \textcolor{blue}{\underline{0.457}} & 0.702 & 0.609 & 0.482 & 0.475 & 0.599 & 0.571 & 0.437 & 0.490 & 0.447 & 0.535 \\
Sdwpfm2 & \textcolor{red}{\textbf{0.340}} & \textcolor{red}{\textbf{0.399}} & 0.478 & \textcolor{blue}{\underline{0.485}} & \textcolor{blue}{\underline{0.475}} & 0.486 & 0.803 & 0.653 & 0.476 & 0.489 & 0.515 & 0.490 & 0.491 & 0.513 & 0.497 & 0.564 \\
Sdwpfh1 & \textcolor{red}{\textbf{0.312}} & \textcolor{red}{\textbf{0.366}} & \textcolor{blue}{\underline{0.449}} & \textcolor{blue}{\underline{0.486}} & 0.450 & 0.500 & 0.746 & 0.644 & 0.479 & 0.492 & 0.539 & 0.516 & 0.537 & 0.598 & 0.577 & 0.638 \\
Sdwpfh2 & \textcolor{red}{\textbf{0.312}} & \textcolor{red}{\textbf{0.383}} & 0.523 & 0.530 & \textcolor{blue}{\underline{0.520}} & 0.536 & 0.891 & 0.719 & 0.566 & \textcolor{blue}{\underline{0.521}} & 0.647 & 0.566 & 0.521 & 0.581 & 0.647 & 0.672 \\
Colbun & \textcolor{red}{\textbf{0.093}} & \textcolor{red}{\textbf{0.145}} & \textcolor{blue}{\underline{0.098}} & \textcolor{blue}{\underline{0.155}} & 0.107 & 0.176 & 0.145 & 0.236 & 0.238 & 0.298 & 0.198 & 0.266 & 0.173 & 0.246 & 0.128 & 0.175 \\
Rapel & \textcolor{red}{\textbf{0.226}} & \textcolor{red}{\textbf{0.279}} & \textcolor{blue}{\underline{0.230}} & \textcolor{blue}{\underline{0.306}} & 0.306 & 0.307 & 0.344 & 0.362 & 0.305 & 0.334 & 0.270 & 0.326 & 0.257 & 0.322 & 0.250 & 0.311 \\
\bottomrule
\end{tabular}%
}
\ifdim\wd0>\linewidth\resizebox{\linewidth}{!}{\usebox0}\else\usebox0\fi
\endgroup
\end{table}

\clearpage

\subsection{Complete Multivariate Forecasting Results}
\label{app:tfb-full}

Tables~\ref{tab:tfb-full-1} and \ref{tab:tfb-full-2} report all TFB results at horizons 96, 192, 336, and 720. Table~\ref{tab:tfb-main} summarizes seven recent advanced supervised baselines; the full tables include all fourteen, divided into two groups with Aurora-X first in each. Each table ranks its displayed models separately, including all ties at the reported precision. Both follow the source selection described above.

\begin{table}[!htbp]
\centering
\caption{Full TFB results, model group 1. Lower is better; red bold and blue underlining indicate the best and second-best distinct values within this table, including all ties at the displayed precision.}
\label{tab:tfb-full-1}
\begingroup
\small
\setlength{\tabcolsep}{2.4pt}
\renewcommand{\arraystretch}{1.08}
\sbox0{%
\begin{tabular}{@{}lccccccccccccccccc@{}}
\toprule
\textbf{Dataset} & $H$ & \multicolumn{2}{c}{\textbf{Aurora-X}} & \multicolumn{2}{c}{\textbf{DUET}} & \multicolumn{2}{c}{\textbf{TimeKAN}} & \multicolumn{2}{c}{\textbf{AMD}} & \multicolumn{2}{c}{\textbf{TimePro}} & \multicolumn{2}{c}{\textbf{SRSNet}} & \multicolumn{2}{c}{\textbf{xPatch}} & \multicolumn{2}{c}{\textbf{Amplifier}} \\
 & & MSE & MAE & MSE & MAE & MSE & MAE & MSE & MAE & MSE & MAE & MSE & MAE & MSE & MAE & MSE & MAE \\
\midrule
\multirow{4}{*}{ETTh1} & 96 & \textcolor{red}{\textbf{0.362}} & \textcolor{red}{\textbf{0.390}} & 0.377 & 0.393 & \textcolor{blue}{\underline{0.374}} & \textcolor{blue}{\underline{0.391}} & 0.375 & 0.392 & 0.375 & 0.398 & 0.383 & 0.395 & 0.378 & \textcolor{red}{\textbf{0.390}} & 0.376 & 0.393 \\
 & 192 & \textcolor{red}{\textbf{0.392}} & \textcolor{red}{\textbf{0.409}} & 0.429 & 0.425 & \textcolor{blue}{\underline{0.421}} & 0.421 & 0.430 & 0.422 & 0.427 & 0.429 & 0.433 & 0.422 & 0.433 & \textcolor{blue}{\underline{0.420}} & 0.442 & 0.430 \\
 & 336 & \textcolor{red}{\textbf{0.404}} & \textcolor{red}{\textbf{0.416}} & 0.471 & 0.446 & \textcolor{blue}{\underline{0.464}} & \textcolor{blue}{\underline{0.440}} & 0.471 & 0.443 & 0.472 & 0.450 & 0.476 & 0.446 & 0.484 & 0.445 & 0.478 & 0.446 \\
 & 720 & \textcolor{red}{\textbf{0.401}} & \textcolor{red}{\textbf{0.427}} & 0.496 & 0.480 & \textcolor{blue}{\underline{0.466}} & \textcolor{blue}{\underline{0.462}} & 0.478 & 0.464 & 0.476 & 0.474 & 0.474 & 0.471 & 0.480 & \textcolor{blue}{\underline{0.462}} & 0.501 & 0.479 \\
\midrule
\multirow{4}{*}{ETTh2} & 96 & \textcolor{blue}{\underline{0.291}} & \textcolor{blue}{\underline{0.334}} & 0.296 & 0.345 & 0.293 & 0.343 & \textcolor{red}{\textbf{0.287}} & 0.338 & 0.293 & 0.345 & 0.296 & 0.345 & \textcolor{red}{\textbf{0.287}} & \textcolor{red}{\textbf{0.332}} & 0.298 & 0.347 \\
 & 192 & \textcolor{red}{\textbf{0.349}} & \textcolor{red}{\textbf{0.373}} & 0.368 & 0.389 & 0.375 & 0.396 & 0.367 & 0.388 & 0.367 & 0.394 & 0.369 & 0.392 & \textcolor{blue}{\underline{0.360}} & \textcolor{blue}{\underline{0.382}} & 0.378 & 0.401 \\
 & 336 & \textcolor{red}{\textbf{0.375}} & \textcolor{red}{\textbf{0.394}} & 0.411 & 0.422 & 0.429 & 0.441 & \textcolor{blue}{\underline{0.410}} & 0.424 & 0.419 & 0.431 & 0.413 & 0.425 & 0.417 & \textcolor{blue}{\underline{0.421}} & 0.428 & 0.437 \\
 & 720 & \textcolor{red}{\textbf{0.396}} & \textcolor{red}{\textbf{0.420}} & \textcolor{blue}{\underline{0.412}} & 0.434 & 0.466 & 0.468 & 0.421 & 0.440 & 0.427 & 0.445 & 0.425 & 0.444 & \textcolor{blue}{\underline{0.412}} & \textcolor{blue}{\underline{0.433}} & 0.452 & 0.460 \\
\midrule
\multirow{4}{*}{ETTm1} & 96 & \textcolor{red}{\textbf{0.285}} & \textcolor{red}{\textbf{0.332}} & 0.324 & 0.354 & 0.327 & 0.365 & 0.327 & 0.361 & 0.326 & 0.361 & 0.319 & 0.358 & \textcolor{blue}{\underline{0.316}} & \textcolor{blue}{\underline{0.343}} & 0.318 & 0.356 \\
 & 192 & \textcolor{red}{\textbf{0.321}} & \textcolor{red}{\textbf{0.354}} & 0.369 & 0.379 & 0.363 & 0.387 & 0.366 & 0.383 & 0.367 & 0.383 & \textcolor{blue}{\underline{0.359}} & 0.381 & 0.369 & \textcolor{blue}{\underline{0.369}} & 0.362 & 0.381 \\
 & 336 & \textcolor{red}{\textbf{0.347}} & \textcolor{red}{\textbf{0.371}} & 0.404 & 0.402 & \textcolor{blue}{\underline{0.389}} & 0.407 & 0.398 & 0.404 & 0.402 & 0.409 & 0.391 & 0.404 & 0.401 & \textcolor{blue}{\underline{0.392}} & 0.393 & 0.404 \\
 & 720 & \textcolor{red}{\textbf{0.387}} & \textcolor{red}{\textbf{0.398}} & 0.463 & 0.437 & \textcolor{blue}{\underline{0.457}} & 0.445 & 0.464 & 0.437 & 0.469 & 0.446 & 0.470 & 0.436 & 0.461 & \textcolor{blue}{\underline{0.429}} & 0.460 & 0.442 \\
\midrule
\multirow{4}{*}{ETTm2} & 96 & \textcolor{red}{\textbf{0.174}} & \textcolor{red}{\textbf{0.248}} & \textcolor{red}{\textbf{0.174}} & 0.255 & 0.178 & 0.262 & \textcolor{blue}{\underline{0.176}} & 0.259 & 0.178 & 0.260 & 0.181 & 0.267 & \textcolor{red}{\textbf{0.174}} & \textcolor{blue}{\underline{0.252}} & 0.178 & 0.261 \\
 & 192 & \textcolor{red}{\textbf{0.235}} & \textcolor{red}{\textbf{0.291}} & 0.243 & 0.302 & 0.244 & 0.308 & 0.242 & 0.302 & 0.242 & 0.303 & 0.243 & 0.306 & \textcolor{blue}{\underline{0.240}} & \textcolor{blue}{\underline{0.297}} & 0.243 & 0.303 \\
 & 336 & \textcolor{red}{\textbf{0.290}} & \textcolor{red}{\textbf{0.329}} & 0.304 & 0.341 & 0.305 & 0.346 & \textcolor{blue}{\underline{0.298}} & 0.337 & 0.303 & 0.342 & 0.306 & 0.346 & 0.302 & \textcolor{blue}{\underline{0.335}} & 0.305 & 0.344 \\
 & 720 & \textcolor{red}{\textbf{0.371}} & \textcolor{red}{\textbf{0.384}} & 0.399 & 0.397 & 0.402 & 0.404 & 0.396 & 0.394 & 0.400 & 0.399 & 0.407 & 0.399 & 0.403 & \textcolor{blue}{\underline{0.393}} & \textcolor{blue}{\underline{0.393}} & 0.397 \\
\midrule
\multirow{4}{*}{Weather} & 96 & \textcolor{red}{\textbf{0.148}} & \textcolor{red}{\textbf{0.189}} & \textcolor{blue}{\underline{0.163}} & \textcolor{blue}{\underline{0.202}} & 0.164 & 0.210 & 0.174 & 0.221 & 0.166 & 0.207 & 0.167 & 0.214 & 0.166 & \textcolor{blue}{\underline{0.202}} & 0.165 & 0.210 \\
 & 192 & \textcolor{red}{\textbf{0.185}} & \textcolor{red}{\textbf{0.226}} & 0.218 & 0.252 & \textcolor{blue}{\underline{0.209}} & 0.250 & 0.219 & 0.259 & 0.216 & 0.254 & 0.215 & 0.255 & 0.210 & \textcolor{blue}{\underline{0.242}} & 0.212 & 0.253 \\
 & 336 & \textcolor{red}{\textbf{0.226}} & \textcolor{red}{\textbf{0.260}} & 0.274 & 0.294 & \textcolor{blue}{\underline{0.264}} & 0.290 & 0.273 & 0.296 & 0.273 & 0.296 & 0.270 & 0.294 & 0.267 & \textcolor{blue}{\underline{0.284}} & 0.267 & 0.293 \\
 & 720 & \textcolor{red}{\textbf{0.286}} & \textcolor{red}{\textbf{0.304}} & 0.349 & 0.343 & \textcolor{blue}{\underline{0.343}} & 0.342 & 0.349 & 0.345 & 0.351 & 0.346 & 0.346 & 0.344 & 0.344 & \textcolor{blue}{\underline{0.335}} & 0.344 & 0.342 \\
\midrule
\multirow{4}{*}{Solar} & 96 & 0.201 & \textcolor{blue}{\underline{0.225}} & 0.200 & \textcolor{red}{\textbf{0.207}} & 0.228 & 0.254 & 0.209 & 0.241 & \textcolor{blue}{\underline{0.196}} & 0.237 & 0.216 & 0.258 & 0.234 & 0.245 & \textcolor{red}{\textbf{0.186}} & 0.232 \\
 & 192 & \textcolor{red}{\textbf{0.223}} & \textcolor{blue}{\underline{0.241}} & \textcolor{blue}{\underline{0.228}} & \textcolor{red}{\textbf{0.233}} & 0.241 & 0.284 & 0.235 & 0.260 & 0.231 & 0.263 & 0.247 & 0.280 & 0.265 & 0.262 & 0.231 & 0.264 \\
 & 336 & \textcolor{blue}{\underline{0.238}} & \textcolor{blue}{\underline{0.252}} & 0.262 & \textcolor{red}{\textbf{0.244}} & 0.273 & 0.292 & 0.294 & 0.316 & 0.250 & 0.281 & 0.268 & 0.294 & 0.301 & 0.280 & \textcolor{red}{\textbf{0.234}} & 0.263 \\
 & 720 & \textcolor{red}{\textbf{0.225}} & \textcolor{red}{\textbf{0.249}} & 0.258 & \textcolor{red}{\textbf{0.249}} & 0.279 & 0.297 & 0.275 & 0.308 & 0.253 & 0.285 & 0.268 & 0.290 & 0.308 & 0.284 & \textcolor{blue}{\underline{0.238}} & \textcolor{blue}{\underline{0.265}} \\
\midrule
\multirow{4}{*}{Electricity} & 96 & 0.146 & 0.244 & \textcolor{blue}{\underline{0.145}} & \textcolor{red}{\textbf{0.233}} & 0.174 & 0.266 & 0.147 & 0.251 & \textcolor{red}{\textbf{0.139}} & \textcolor{blue}{\underline{0.234}} & 0.161 & 0.252 & 0.160 & 0.244 & 0.149 & 0.245 \\
 & 192 & \textcolor{blue}{\underline{0.159}} & 0.255 & 0.163 & \textcolor{red}{\textbf{0.248}} & 0.182 & 0.273 & 0.176 & 0.262 & \textcolor{red}{\textbf{0.156}} & \textcolor{blue}{\underline{0.249}} & 0.172 & 0.261 & 0.169 & 0.253 & 0.165 & 0.260 \\
 & 336 & \textcolor{red}{\textbf{0.172}} & \textcolor{blue}{\underline{0.268}} & \textcolor{blue}{\underline{0.175}} & \textcolor{red}{\textbf{0.262}} & 0.197 & 0.286 & 0.193 & 0.281 & \textcolor{red}{\textbf{0.172}} & 0.269 & 0.190 & 0.279 & 0.185 & \textcolor{blue}{\underline{0.268}} & 0.176 & 0.271 \\
 & 720 & \textcolor{red}{\textbf{0.204}} & \textcolor{blue}{\underline{0.295}} & \textcolor{red}{\textbf{0.204}} & \textcolor{red}{\textbf{0.291}} & 0.236 & 0.320 & 0.232 & 0.329 & \textcolor{blue}{\underline{0.209}} & 0.299 & 0.231 & 0.313 & 0.221 & 0.300 & \textcolor{red}{\textbf{0.204}} & 0.296 \\
\midrule
\multirow{4}{*}{Traffic} & 96 & \textcolor{red}{\textbf{0.389}} & \textcolor{blue}{\underline{0.260}} & \textcolor{blue}{\underline{0.407}} & \textcolor{red}{\textbf{0.252}} & 0.423 & 0.286 & 0.443 & 0.298 & 0.426 & 0.292 & 0.471 & 0.295 & 0.475 & 0.279 & 0.450 & 0.295 \\
 & 192 & \textcolor{red}{\textbf{0.403}} & \textcolor{blue}{\underline{0.265}} & \textcolor{blue}{\underline{0.431}} & \textcolor{red}{\textbf{0.262}} & 0.442 & 0.295 & 0.496 & 0.323 & 0.439 & 0.298 & 0.480 & 0.300 & 0.486 & 0.277 & 0.489 & 0.311 \\
 & 336 & \textcolor{red}{\textbf{0.414}} & \textcolor{red}{\textbf{0.269}} & 0.456 & \textcolor{red}{\textbf{0.269}} & 0.473 & 0.335 & 0.520 & 0.330 & \textcolor{blue}{\underline{0.449}} & 0.307 & 0.496 & 0.306 & 0.500 & \textcolor{blue}{\underline{0.280}} & 0.484 & 0.321 \\
 & 720 & \textcolor{red}{\textbf{0.445}} & \textcolor{red}{\textbf{0.285}} & 0.509 & \textcolor{blue}{\underline{0.292}} & 0.481 & 0.357 & 0.540 & 0.344 & \textcolor{blue}{\underline{0.475}} & 0.309 & 0.531 & 0.328 & 0.537 & 0.295 & 0.517 & 0.333 \\
\bottomrule
\end{tabular}%
}
\ifdim\wd0>\linewidth\resizebox{\linewidth}{!}{\usebox0}\else\usebox0\fi
\endgroup
\end{table}

\begin{table}[!htbp]
\centering
\caption{Full TFB results, model group 2. Lower is better; red bold and blue underlining indicate the best and second-best distinct values within this table, including all ties at the displayed precision.}
\label{tab:tfb-full-2}
\begingroup
\small
\setlength{\tabcolsep}{2.4pt}
\renewcommand{\arraystretch}{1.08}
\sbox0{%
\begin{tabular}{@{}lccccccccccccccccc@{}}
\toprule
\textbf{Dataset} & $H$ & \multicolumn{2}{c}{\textbf{Aurora-X}} & \multicolumn{2}{c}{\textbf{Fredformer}} & \multicolumn{2}{c}{\textbf{MSGNet}} & \multicolumn{2}{c}{\textbf{iTransformer}} & \multicolumn{2}{c}{\textbf{PatchTST}} & \multicolumn{2}{c}{\textbf{Crossformer}} & \multicolumn{2}{c}{\textbf{TimesNet}} & \multicolumn{2}{c}{\textbf{DLinear}} \\
 & & MSE & MAE & MSE & MAE & MSE & MAE & MSE & MAE & MSE & MAE & MSE & MAE & MSE & MAE & MSE & MAE \\
\midrule
\multirow{4}{*}{ETTh1} & 96 & \textcolor{red}{\textbf{0.362}} & \textcolor{red}{\textbf{0.390}} & \textcolor{blue}{\underline{0.378}} & \textcolor{blue}{\underline{0.395}} & 0.390 & 0.411 & 0.386 & 0.405 & 0.414 & 0.419 & 0.423 & 0.448 & 0.384 & 0.402 & 0.397 & 0.412 \\
 & 192 & \textcolor{red}{\textbf{0.392}} & \textcolor{red}{\textbf{0.409}} & \textcolor{blue}{\underline{0.435}} & \textcolor{blue}{\underline{0.424}} & 0.443 & 0.442 & 0.441 & 0.436 & 0.460 & 0.445 & 0.471 & 0.474 & 0.436 & 0.429 & 0.446 & 0.441 \\
 & 336 & \textcolor{red}{\textbf{0.404}} & \textcolor{red}{\textbf{0.416}} & 0.485 & \textcolor{blue}{\underline{0.447}} & \textcolor{blue}{\underline{0.482}} & 0.469 & 0.487 & 0.458 & 0.501 & 0.466 & 0.570 & 0.546 & 0.491 & 0.469 & 0.489 & 0.467 \\
 & 720 & \textcolor{red}{\textbf{0.401}} & \textcolor{red}{\textbf{0.427}} & \textcolor{blue}{\underline{0.496}} & \textcolor{blue}{\underline{0.472}} & \textcolor{blue}{\underline{0.496}} & 0.488 & 0.503 & 0.491 & 0.500 & 0.488 & 0.653 & 0.621 & 0.521 & 0.500 & 0.513 & 0.510 \\
\midrule
\multirow{4}{*}{ETTh2} & 96 & \textcolor{red}{\textbf{0.291}} & \textcolor{red}{\textbf{0.334}} & \textcolor{red}{\textbf{0.291}} & \textcolor{blue}{\underline{0.342}} & 0.329 & 0.371 & \textcolor{blue}{\underline{0.297}} & 0.349 & 0.302 & 0.348 & 0.745 & 0.584 & 0.340 & 0.374 & 0.340 & 0.394 \\
 & 192 & \textcolor{red}{\textbf{0.349}} & \textcolor{red}{\textbf{0.373}} & \textcolor{blue}{\underline{0.372}} & \textcolor{blue}{\underline{0.390}} & 0.402 & 0.414 & 0.380 & 0.400 & 0.388 & 0.400 & 0.877 & 0.656 & 0.402 & 0.414 & 0.482 & 0.479 \\
 & 336 & \textcolor{red}{\textbf{0.375}} & \textcolor{red}{\textbf{0.394}} & \textcolor{blue}{\underline{0.419}} & \textcolor{blue}{\underline{0.431}} & 0.440 & 0.445 & 0.428 & 0.432 & 0.426 & 0.433 & 1.043 & 0.731 & 0.452 & 0.452 & 0.591 & 0.541 \\
 & 720 & \textcolor{red}{\textbf{0.396}} & \textcolor{red}{\textbf{0.420}} & 0.431 & 0.450 & 0.480 & 0.477 & \textcolor{blue}{\underline{0.427}} & \textcolor{blue}{\underline{0.445}} & 0.431 & 0.446 & 1.104 & 0.763 & 0.462 & 0.468 & 0.839 & 0.661 \\
\midrule
\multirow{4}{*}{ETTm1} & 96 & \textcolor{red}{\textbf{0.285}} & \textcolor{red}{\textbf{0.332}} & 0.326 & \textcolor{blue}{\underline{0.361}} & \textcolor{blue}{\underline{0.319}} & 0.366 & 0.334 & 0.368 & 0.329 & 0.367 & 0.404 & 0.426 & 0.338 & 0.375 & 0.346 & 0.374 \\
 & 192 & \textcolor{red}{\textbf{0.321}} & \textcolor{red}{\textbf{0.354}} & \textcolor{blue}{\underline{0.363}} & \textcolor{blue}{\underline{0.384}} & 0.377 & 0.397 & 0.377 & 0.391 & 0.367 & 0.385 & 0.450 & 0.451 & 0.374 & 0.387 & 0.382 & 0.391 \\
 & 336 & \textcolor{red}{\textbf{0.347}} & \textcolor{red}{\textbf{0.371}} & \textcolor{blue}{\underline{0.395}} & \textcolor{blue}{\underline{0.406}} & 0.417 & 0.422 & 0.426 & 0.420 & 0.399 & 0.410 & 0.532 & 0.515 & 0.410 & 0.411 & 0.410 & 0.415 \\
 & 720 & \textcolor{red}{\textbf{0.387}} & \textcolor{red}{\textbf{0.398}} & 0.456 & 0.441 & 0.487 & 0.463 & 0.491 & 0.459 & \textcolor{blue}{\underline{0.454}} & \textcolor{blue}{\underline{0.439}} & 0.666 & 0.589 & 0.478 & 0.450 & 0.473 & 0.451 \\
\midrule
\multirow{4}{*}{ETTm2} & 96 & \textcolor{red}{\textbf{0.174}} & \textcolor{red}{\textbf{0.248}} & 0.177 & \textcolor{blue}{\underline{0.258}} & 0.182 & 0.266 & 0.180 & 0.264 & \textcolor{blue}{\underline{0.175}} & 0.259 & 0.287 & 0.366 & 0.187 & 0.267 & 0.193 & 0.293 \\
 & 192 & \textcolor{red}{\textbf{0.235}} & \textcolor{red}{\textbf{0.291}} & 0.243 & \textcolor{blue}{\underline{0.301}} & 0.248 & 0.306 & 0.250 & 0.309 & \textcolor{blue}{\underline{0.241}} & 0.302 & 0.414 & 0.492 & 0.249 & 0.309 & 0.284 & 0.361 \\
 & 336 & \textcolor{red}{\textbf{0.290}} & \textcolor{red}{\textbf{0.329}} & \textcolor{blue}{\underline{0.302}} & \textcolor{blue}{\underline{0.340}} & 0.312 & 0.346 & 0.311 & 0.348 & 0.305 & 0.343 & 0.597 & 0.542 & 0.321 & 0.351 & 0.382 & 0.429 \\
 & 720 & \textcolor{red}{\textbf{0.371}} & \textcolor{red}{\textbf{0.384}} & 0.404 & \textcolor{blue}{\underline{0.398}} & 0.414 & 0.404 & 0.412 & 0.407 & \textcolor{blue}{\underline{0.402}} & 0.400 & 1.730 & 1.042 & 0.408 & 0.403 & 0.558 & 0.525 \\
\midrule
\multirow{4}{*}{Weather} & 96 & \textcolor{red}{\textbf{0.148}} & \textcolor{red}{\textbf{0.189}} & 0.163 & \textcolor{blue}{\underline{0.207}} & 0.163 & 0.212 & 0.174 & 0.214 & 0.177 & 0.218 & \textcolor{blue}{\underline{0.158}} & 0.230 & 0.172 & 0.220 & 0.195 & 0.252 \\
 & 192 & \textcolor{red}{\textbf{0.185}} & \textcolor{red}{\textbf{0.226}} & 0.224 & 0.258 & 0.211 & \textcolor{blue}{\underline{0.254}} & 0.221 & \textcolor{blue}{\underline{0.254}} & 0.225 & 0.259 & \textcolor{blue}{\underline{0.206}} & 0.277 & 0.219 & 0.261 & 0.237 & 0.295 \\
 & 336 & \textcolor{red}{\textbf{0.226}} & \textcolor{red}{\textbf{0.260}} & 0.278 & 0.298 & 0.273 & 0.299 & 0.278 & \textcolor{blue}{\underline{0.296}} & 0.278 & 0.297 & \textcolor{blue}{\underline{0.272}} & 0.335 & 0.280 & 0.306 & 0.282 & 0.331 \\
 & 720 & \textcolor{red}{\textbf{0.286}} & \textcolor{red}{\textbf{0.304}} & 0.357 & 0.350 & 0.351 & \textcolor{blue}{\underline{0.348}} & 0.358 & 0.349 & 0.354 & \textcolor{blue}{\underline{0.348}} & 0.398 & 0.418 & 0.365 & 0.359 & \textcolor{blue}{\underline{0.345}} & 0.382 \\
\midrule
\multirow{4}{*}{Solar} & 96 & \textcolor{blue}{\underline{0.201}} & \textcolor{red}{\textbf{0.225}} & \textcolor{red}{\textbf{0.189}} & \textcolor{blue}{\underline{0.236}} & 0.210 & 0.246 & 0.203 & 0.237 & 0.234 & 0.286 & 0.310 & 0.331 & 0.250 & 0.292 & 0.290 & 0.378 \\
 & 192 & \textcolor{red}{\textbf{0.223}} & \textcolor{red}{\textbf{0.241}} & \textcolor{blue}{\underline{0.227}} & \textcolor{blue}{\underline{0.259}} & 0.265 & 0.290 & 0.233 & 0.261 & 0.267 & 0.310 & 0.734 & 0.725 & 0.296 & 0.318 & 0.320 & 0.398 \\
 & 336 & \textcolor{red}{\textbf{0.238}} & \textcolor{red}{\textbf{0.252}} & \textcolor{blue}{\underline{0.243}} & 0.286 & 0.294 & 0.318 & 0.248 & \textcolor{blue}{\underline{0.273}} & 0.290 & 0.315 & 0.750 & 0.735 & 0.319 & 0.330 & 0.353 & 0.415 \\
 & 720 & \textcolor{red}{\textbf{0.225}} & \textcolor{red}{\textbf{0.249}} & 0.250 & 0.285 & 0.285 & 0.315 & \textcolor{blue}{\underline{0.249}} & \textcolor{blue}{\underline{0.275}} & 0.289 & 0.317 & 0.769 & 0.765 & 0.338 & 0.337 & 0.357 & 0.413 \\
\midrule
\multirow{4}{*}{Electricity} & 96 & \textcolor{red}{\textbf{0.146}} & 0.244 & \textcolor{blue}{\underline{0.148}} & \textcolor{blue}{\underline{0.242}} & 0.165 & 0.274 & \textcolor{blue}{\underline{0.148}} & \textcolor{red}{\textbf{0.240}} & 0.195 & 0.285 & 0.219 & 0.314 & 0.168 & 0.272 & 0.210 & 0.302 \\
 & 192 & \textcolor{red}{\textbf{0.159}} & \textcolor{blue}{\underline{0.255}} & 0.165 & 0.257 & 0.185 & 0.292 & \textcolor{blue}{\underline{0.162}} & \textcolor{red}{\textbf{0.253}} & 0.199 & 0.289 & 0.231 & 0.322 & 0.184 & 0.289 & 0.210 & 0.305 \\
 & 336 & \textcolor{red}{\textbf{0.172}} & \textcolor{red}{\textbf{0.268}} & 0.180 & 0.274 & 0.197 & 0.304 & \textcolor{blue}{\underline{0.178}} & \textcolor{blue}{\underline{0.269}} & 0.215 & 0.305 & 0.246 & 0.337 & 0.198 & 0.300 & 0.223 & 0.319 \\
 & 720 & \textcolor{red}{\textbf{0.204}} & \textcolor{red}{\textbf{0.295}} & \textcolor{blue}{\underline{0.218}} & \textcolor{blue}{\underline{0.305}} & 0.231 & 0.332 & 0.225 & 0.317 & 0.256 & 0.337 & 0.280 & 0.363 & 0.220 & 0.320 & 0.258 & 0.350 \\
\midrule
\multirow{4}{*}{Traffic} & 96 & \textcolor{red}{\textbf{0.389}} & \textcolor{red}{\textbf{0.260}} & 0.403 & 0.274 & 0.608 & 0.349 & \textcolor{blue}{\underline{0.395}} & \textcolor{blue}{\underline{0.268}} & 0.544 & 0.359 & 0.522 & 0.290 & 0.593 & 0.321 & 0.650 & 0.396 \\
 & 192 & \textcolor{red}{\textbf{0.403}} & \textcolor{red}{\textbf{0.265}} & 0.429 & 0.289 & 0.634 & 0.371 & \textcolor{blue}{\underline{0.417}} & \textcolor{blue}{\underline{0.276}} & 0.540 & 0.354 & 0.530 & 0.293 & 0.617 & 0.336 & 0.598 & 0.370 \\
 & 336 & \textcolor{red}{\textbf{0.414}} & \textcolor{red}{\textbf{0.269}} & 0.441 & 0.295 & 0.669 & 0.388 & \textcolor{blue}{\underline{0.433}} & \textcolor{blue}{\underline{0.283}} & 0.551 & 0.358 & 0.558 & 0.305 & 0.629 & 0.336 & 0.605 & 0.373 \\
 & 720 & \textcolor{red}{\textbf{0.445}} & \textcolor{red}{\textbf{0.285}} & \textcolor{blue}{\underline{0.463}} & \textcolor{blue}{\underline{0.300}} & 0.729 & 0.420 & 0.467 & 0.302 & 0.586 & 0.375 & 0.589 & 0.328 & 0.640 & 0.350 & 0.645 & 0.394 \\
\bottomrule
\end{tabular}%
}
\ifdim\wd0>\linewidth\resizebox{\linewidth}{!}{\usebox0}\else\usebox0\fi
\endgroup
\end{table}

\clearpage
\subsection{Complete Covariate Forecasting Results}
\label{app:dag-full}

Tables~\ref{tab:dag-full-1} and \ref{tab:dag-full-2} report both forecast horizons for each DAG-Bench dataset. Table~\ref{tab:dag-main} summarizes DAG, GCGNet, TimeXer, TFT, DUET, Amplifier, and TimeKAN. Here, all ten supervised baselines, including TiDE, CrossLinear, and PatchTST, are divided into two groups with Aurora-X as the first model in each. Each table ranks its displayed models separately, including all ties at the reported precision. Aurora-X achieves the lowest horizon-averaged MSE and MAE on all four SDWPF subsets.

\begin{table}[!htbp]
\centering
\caption{Full DAG-Bench results, model group 1. Lower is better; red bold and blue underlining indicate the best and second-best distinct values within this table, including all ties at the displayed precision.}
\label{tab:dag-full-1}
\begingroup
\small
\setlength{\tabcolsep}{2.4pt}
\renewcommand{\arraystretch}{1.08}
\sbox0{%
\begin{tabular}{@{}lccccccccccccc@{}}
\toprule
\textbf{Dataset} & $H$ & \multicolumn{2}{c}{\textbf{Aurora-X}} & \multicolumn{2}{c}{\textbf{DAG}} & \multicolumn{2}{c}{\textbf{GCGNet}} & \multicolumn{2}{c}{\textbf{TimeXer}} & \multicolumn{2}{c}{\textbf{TFT}} & \multicolumn{2}{c}{\textbf{TiDE}} \\
 & & MSE & MAE & MSE & MAE & MSE & MAE & MSE & MAE & MSE & MAE & MSE & MAE \\
\midrule
\multirow{2}{*}{NP} & 24 & 0.231 & 0.258 & \textcolor{red}{\textbf{0.202}} & \textcolor{red}{\textbf{0.237}} & \textcolor{blue}{\underline{0.208}} & \textcolor{red}{\textbf{0.237}} & 0.236 & 0.266 & 0.219 & \textcolor{blue}{\underline{0.249}} & 0.284 & 0.301 \\
 & 360 & \textcolor{red}{\textbf{0.487}} & \textcolor{red}{\textbf{0.434}} & \textcolor{blue}{\underline{0.521}} & \textcolor{blue}{\underline{0.451}} & 0.531 & 0.459 & 0.600 & 0.475 & 0.539 & 0.501 & 0.601 & 0.498 \\
\midrule
\multirow{2}{*}{PJM} & 24 & 0.073 & 0.172 & \textcolor{red}{\textbf{0.057}} & \textcolor{red}{\textbf{0.143}} & \textcolor{blue}{\underline{0.060}} & \textcolor{blue}{\underline{0.150}} & 0.075 & 0.166 & 0.095 & 0.195 & 0.106 & 0.214 \\
 & 360 & 0.150 & 0.236 & \textcolor{blue}{\underline{0.130}} & \textcolor{red}{\textbf{0.218}} & \textcolor{red}{\textbf{0.129}} & 0.223 & 0.140 & 0.231 & 0.133 & \textcolor{blue}{\underline{0.219}} & 0.177 & 0.279 \\
\midrule
\multirow{2}{*}{BE} & 24 & \textcolor{red}{\textbf{0.348}} & \textcolor{blue}{\underline{0.234}} & 0.361 & \textcolor{red}{\textbf{0.229}} & \textcolor{blue}{\underline{0.350}} & 0.248 & 0.392 & 0.253 & 0.426 & 0.272 & 0.426 & 0.285 \\
 & 360 & \textcolor{red}{\textbf{0.477}} & 0.331 & 0.485 & 0.330 & 0.511 & 0.340 & 0.512 & \textcolor{blue}{\underline{0.327}} & \textcolor{blue}{\underline{0.482}} & \textcolor{red}{\textbf{0.310}} & 0.571 & 0.364 \\
\midrule
\multirow{2}{*}{FR} & 24 & 0.358 & \textcolor{blue}{\underline{0.187}} & \textcolor{blue}{\underline{0.355}} & \textcolor{red}{\textbf{0.171}} & \textcolor{red}{\textbf{0.347}} & 0.188 & 0.366 & 0.208 & 0.543 & 0.253 & 0.418 & 0.255 \\
 & 360 & \textcolor{red}{\textbf{0.464}} & \textcolor{blue}{\underline{0.262}} & 0.473 & 0.268 & 0.482 & 0.279 & 0.489 & 0.273 & \textcolor{blue}{\underline{0.465}} & \textcolor{red}{\textbf{0.261}} & 0.551 & 0.308 \\
\midrule
\multirow{2}{*}{DE} & 24 & 0.358 & 0.379 & \textcolor{red}{\textbf{0.277}} & \textcolor{red}{\textbf{0.322}} & \textcolor{blue}{\underline{0.280}} & \textcolor{blue}{\underline{0.331}} & 0.339 & 0.362 & 0.380 & 0.383 & 0.367 & 0.383 \\
 & 360 & \textcolor{blue}{\underline{0.495}} & \textcolor{blue}{\underline{0.439}} & \textcolor{red}{\textbf{0.462}} & \textcolor{red}{\textbf{0.418}} & 0.523 & 0.447 & 0.610 & 0.474 & 0.599 & 0.509 & 0.630 & 0.511 \\
\midrule
\multirow{2}{*}{Energy} & 24 & 0.137 & 0.288 & \textcolor{red}{\textbf{0.079}} & \textcolor{red}{\textbf{0.215}} & \textcolor{blue}{\underline{0.081}} & \textcolor{blue}{\underline{0.221}} & 0.122 & 0.273 & 0.093 & 0.235 & 0.103 & 0.248 \\
 & 360 & 0.257 & 0.398 & \textcolor{blue}{\underline{0.169}} & \textcolor{red}{\textbf{0.320}} & 0.182 & 0.334 & 0.204 & 0.357 & \textcolor{red}{\textbf{0.167}} & \textcolor{blue}{\underline{0.331}} & 0.202 & 0.355 \\
\midrule
\multirow{2}{*}{Sdwpfm1} & 24 & \textcolor{red}{\textbf{0.321}} & \textcolor{red}{\textbf{0.373}} & \textcolor{blue}{\underline{0.351}} & \textcolor{blue}{\underline{0.400}} & 0.376 & 0.415 & 0.558 & 0.533 & 0.366 & 0.421 & 0.474 & 0.488 \\
 & 360 & \textcolor{red}{\textbf{0.339}} & \textcolor{red}{\textbf{0.403}} & 0.495 & 0.522 & \textcolor{blue}{\underline{0.472}} & \textcolor{blue}{\underline{0.499}} & 0.845 & 0.684 & 0.597 & 0.528 & 0.492 & 0.526 \\
\midrule
\multirow{2}{*}{Sdwpfm2} & 24 & \textcolor{red}{\textbf{0.333}} & \textcolor{red}{\textbf{0.387}} & \textcolor{blue}{\underline{0.372}} & \textcolor{blue}{\underline{0.414}} & 0.421 & 0.441 & 0.627 & 0.570 & 0.411 & 0.458 & 0.461 & 0.492 \\
 & 360 & \textcolor{red}{\textbf{0.347}} & \textcolor{red}{\textbf{0.411}} & 0.583 & 0.556 & 0.529 & 0.531 & 0.978 & 0.736 & 0.541 & \textcolor{blue}{\underline{0.519}} & \textcolor{blue}{\underline{0.511}} & 0.540 \\
\midrule
\multirow{2}{*}{Sdwpfh1} & 24 & \textcolor{red}{\textbf{0.314}} & \textcolor{red}{\textbf{0.358}} & 0.408 & \textcolor{blue}{\underline{0.438}} & 0.435 & 0.486 & 0.651 & 0.587 & \textcolor{blue}{\underline{0.401}} & 0.460 & 0.434 & 0.489 \\
 & 360 & \textcolor{red}{\textbf{0.310}} & \textcolor{red}{\textbf{0.373}} & 0.489 & 0.534 & \textcolor{blue}{\underline{0.465}} & \textcolor{blue}{\underline{0.514}} & 0.841 & 0.700 & 0.557 & 0.523 & 0.472 & 0.527 \\
\midrule
\multirow{2}{*}{Sdwpfh2} & 24 & \textcolor{red}{\textbf{0.311}} & \textcolor{red}{\textbf{0.374}} & \textcolor{blue}{\underline{0.438}} & \textcolor{blue}{\underline{0.465}} & 0.473 & 0.506 & 0.820 & 0.677 & 0.474 & 0.493 & 0.579 & 0.553 \\
 & 360 & \textcolor{red}{\textbf{0.312}} & \textcolor{red}{\textbf{0.391}} & 0.608 & 0.595 & \textcolor{blue}{\underline{0.566}} & 0.565 & 0.962 & 0.761 & 0.657 & \textcolor{blue}{\underline{0.549}} & 0.619 & 0.614 \\
\midrule
\multirow{2}{*}{Colbun} & 10 & \textcolor{red}{\textbf{0.059}} & 0.112 & \textcolor{blue}{\underline{0.061}} & \textcolor{red}{\textbf{0.094}} & 0.065 & \textcolor{blue}{\underline{0.108}} & 0.113 & 0.172 & 0.092 & 0.135 & 0.089 & 0.131 \\
 & 30 & \textcolor{red}{\textbf{0.126}} & \textcolor{red}{\textbf{0.178}} & \textcolor{blue}{\underline{0.135}} & \textcolor{blue}{\underline{0.215}} & 0.149 & 0.243 & 0.176 & 0.299 & 0.383 & 0.460 & 0.240 & 0.322 \\
\midrule
\multirow{2}{*}{Rapel} & 10 & \textcolor{blue}{\underline{0.165}} & \textcolor{blue}{\underline{0.213}} & \textcolor{red}{\textbf{0.151}} & \textcolor{red}{\textbf{0.203}} & 0.211 & 0.230 & 0.301 & 0.308 & 0.201 & 0.253 & 0.228 & 0.271 \\
 & 30 & \textcolor{red}{\textbf{0.286}} & \textcolor{red}{\textbf{0.345}} & \textcolor{blue}{\underline{0.309}} & 0.408 & 0.401 & \textcolor{blue}{\underline{0.384}} & 0.387 & 0.416 & 0.409 & 0.414 & 0.411 & 0.432 \\
\bottomrule
\end{tabular}%
}
\ifdim\wd0>\linewidth\resizebox{\linewidth}{!}{\usebox0}\else\usebox0\fi
\endgroup
\end{table}

\begin{table}[!htbp]
\centering
\caption{Full DAG-Bench results, model group 2. Lower is better; red bold and blue underlining indicate the best and second-best distinct values within this table, including all ties at the displayed precision.}
\label{tab:dag-full-2}
\begingroup
\small
\setlength{\tabcolsep}{2.4pt}
\renewcommand{\arraystretch}{1.08}
\sbox0{%
\begin{tabular}{@{}lccccccccccccc@{}}
\toprule
\textbf{Dataset} & $H$ & \multicolumn{2}{c}{\textbf{Aurora-X}} & \multicolumn{2}{c}{\textbf{DUET}} & \multicolumn{2}{c}{\textbf{CrossLinear}} & \multicolumn{2}{c}{\textbf{Amplifier}} & \multicolumn{2}{c}{\textbf{TimeKAN}} & \multicolumn{2}{c}{\textbf{PatchTST}} \\
 & & MSE & MAE & MSE & MAE & MSE & MAE & MSE & MAE & MSE & MAE & MSE & MAE \\
\midrule
\multirow{2}{*}{NP} & 24 & \textcolor{blue}{\underline{0.231}} & \textcolor{red}{\textbf{0.258}} & 0.246 & 0.287 & \textcolor{red}{\textbf{0.210}} & \textcolor{blue}{\underline{0.266}} & 0.252 & 0.303 & 0.273 & 0.310 & 0.249 & 0.294 \\
 & 360 & \textcolor{red}{\textbf{0.487}} & \textcolor{red}{\textbf{0.434}} & 0.576 & 0.528 & 0.531 & 0.508 & 0.587 & 0.534 & 0.538 & 0.529 & \textcolor{blue}{\underline{0.530}} & \textcolor{blue}{\underline{0.498}} \\
\midrule
\multirow{2}{*}{PJM} & 24 & \textcolor{blue}{\underline{0.073}} & \textcolor{blue}{\underline{0.172}} & \textcolor{red}{\textbf{0.072}} & \textcolor{red}{\textbf{0.166}} & 0.088 & 0.191 & 0.096 & 0.208 & 0.115 & 0.244 & 0.116 & 0.239 \\
 & 360 & 0.150 & \textcolor{blue}{\underline{0.236}} & \textcolor{red}{\textbf{0.131}} & \textcolor{red}{\textbf{0.228}} & \textcolor{blue}{\underline{0.135}} & 0.254 & 0.177 & 0.285 & 0.162 & 0.281 & 0.149 & 0.279 \\
\midrule
\multirow{2}{*}{BE} & 24 & \textcolor{red}{\textbf{0.348}} & \textcolor{red}{\textbf{0.234}} & 0.432 & 0.272 & \textcolor{blue}{\underline{0.391}} & \textcolor{blue}{\underline{0.259}} & 0.471 & 0.339 & 0.451 & 0.319 & 0.452 & 0.326 \\
 & 360 & \textcolor{red}{\textbf{0.477}} & \textcolor{red}{\textbf{0.331}} & 0.597 & 0.436 & \textcolor{blue}{\underline{0.568}} & \textcolor{blue}{\underline{0.416}} & 0.646 & 0.487 & 0.645 & 0.495 & 0.702 & 0.538 \\
\midrule
\multirow{2}{*}{FR} & 24 & \textcolor{red}{\textbf{0.358}} & \textcolor{red}{\textbf{0.187}} & \textcolor{blue}{\underline{0.384}} & 0.251 & 0.390 & \textcolor{blue}{\underline{0.226}} & 0.459 & 0.348 & 0.454 & 0.296 & 0.518 & 0.368 \\
 & 360 & \textcolor{red}{\textbf{0.464}} & \textcolor{red}{\textbf{0.262}} & 0.607 & 0.403 & \textcolor{blue}{\underline{0.575}} & \textcolor{blue}{\underline{0.370}} & 0.648 & 0.468 & 0.641 & 0.452 & 0.658 & 0.452 \\
\midrule
\multirow{2}{*}{DE} & 24 & \textcolor{red}{\textbf{0.358}} & \textcolor{blue}{\underline{0.379}} & \textcolor{blue}{\underline{0.376}} & \textcolor{red}{\textbf{0.378}} & 0.387 & 0.396 & 0.394 & 0.407 & 0.399 & 0.412 & 0.413 & 0.412 \\
 & 360 & \textcolor{red}{\textbf{0.495}} & \textcolor{red}{\textbf{0.439}} & 0.589 & 0.482 & 0.583 & 0.507 & 0.551 & \textcolor{blue}{\underline{0.474}} & \textcolor{blue}{\underline{0.547}} & 0.479 & 0.589 & 0.498 \\
\midrule
\multirow{2}{*}{Energy} & 24 & 0.137 & \textcolor{blue}{\underline{0.288}} & \textcolor{red}{\textbf{0.117}} & \textcolor{red}{\textbf{0.283}} & 0.241 & 0.418 & 0.138 & 0.306 & \textcolor{blue}{\underline{0.135}} & 0.298 & 0.239 & 0.390 \\
 & 360 & 0.257 & 0.398 & 0.288 & 0.452 & \textcolor{blue}{\underline{0.237}} & \textcolor{blue}{\underline{0.385}} & 0.328 & 0.472 & 0.302 & 0.464 & \textcolor{red}{\textbf{0.214}} & \textcolor{red}{\textbf{0.363}} \\
\midrule
\multirow{2}{*}{Sdwpfm1} & 24 & \textcolor{red}{\textbf{0.321}} & \textcolor{red}{\textbf{0.373}} & 0.551 & 0.564 & \textcolor{blue}{\underline{0.355}} & 0.473 & 0.364 & \textcolor{blue}{\underline{0.445}} & 0.418 & 0.503 & 0.374 & 0.454 \\
 & 360 & \textcolor{red}{\textbf{0.339}} & \textcolor{red}{\textbf{0.403}} & 0.646 & 0.577 & 0.497 & \textcolor{blue}{\underline{0.532}} & 0.510 & 0.535 & \textcolor{blue}{\underline{0.476}} & 0.566 & 0.497 & 0.550 \\
\midrule
\multirow{2}{*}{Sdwpfm2} & 24 & \textcolor{red}{\textbf{0.333}} & \textcolor{red}{\textbf{0.387}} & 0.445 & \textcolor{blue}{\underline{0.452}} & 0.477 & 0.536 & \textcolor{blue}{\underline{0.394}} & 0.462 & 0.474 & 0.538 & 0.418 & 0.492 \\
 & 360 & \textcolor{red}{\textbf{0.347}} & \textcolor{red}{\textbf{0.411}} & 0.584 & \textcolor{blue}{\underline{0.528}} & 0.589 & 0.611 & 0.587 & 0.563 & \textcolor{blue}{\underline{0.520}} & 0.589 & 0.602 & 0.603 \\
\midrule
\multirow{2}{*}{Sdwpfh1} & 24 & \textcolor{red}{\textbf{0.314}} & \textcolor{red}{\textbf{0.358}} & 0.527 & 0.513 & 0.548 & 0.585 & 0.576 & 0.627 & 0.511 & 0.582 & \textcolor{blue}{\underline{0.422}} & \textcolor{blue}{\underline{0.505}} \\
 & 360 & \textcolor{red}{\textbf{0.310}} & \textcolor{red}{\textbf{0.373}} & 0.551 & \textcolor{blue}{\underline{0.519}} & 0.566 & 0.601 & \textcolor{blue}{\underline{0.497}} & 0.569 & 0.643 & 0.694 & 0.513 & 0.548 \\
\midrule
\multirow{2}{*}{Sdwpfh2} & 24 & \textcolor{red}{\textbf{0.311}} & \textcolor{red}{\textbf{0.374}} & 0.629 & 0.563 & 0.468 & 0.540 & 0.473 & 0.533 & 0.580 & 0.614 & \textcolor{blue}{\underline{0.457}} & \textcolor{blue}{\underline{0.527}} \\
 & 360 & \textcolor{red}{\textbf{0.312}} & \textcolor{red}{\textbf{0.391}} & 0.665 & \textcolor{blue}{\underline{0.569}} & 0.608 & 0.609 & \textcolor{blue}{\underline{0.569}} & 0.628 & 0.713 & 0.729 & 0.641 & 0.632 \\
\midrule
\multirow{2}{*}{Colbun} & 10 & \textcolor{red}{\textbf{0.059}} & 0.112 & 0.089 & 0.134 & 0.071 & \textcolor{blue}{\underline{0.102}} & 0.071 & 0.121 & \textcolor{blue}{\underline{0.061}} & \textcolor{red}{\textbf{0.101}} & 0.062 & 0.122 \\
 & 30 & \textcolor{red}{\textbf{0.126}} & \textcolor{red}{\textbf{0.178}} & 0.307 & 0.397 & \textcolor{blue}{\underline{0.182}} & 0.288 & 0.275 & 0.370 & 0.195 & \textcolor{blue}{\underline{0.249}} & 0.417 & 0.496 \\
\midrule
\multirow{2}{*}{Rapel} & 10 & \textcolor{blue}{\underline{0.165}} & \textcolor{blue}{\underline{0.213}} & 0.174 & 0.219 & \textcolor{red}{\textbf{0.163}} & \textcolor{red}{\textbf{0.209}} & 0.181 & 0.227 & 0.174 & 0.231 & \textcolor{red}{\textbf{0.163}} & 0.218 \\
 & 30 & \textcolor{red}{\textbf{0.286}} & \textcolor{red}{\textbf{0.345}} & 0.365 & 0.432 & 0.340 & 0.417 & 0.333 & 0.416 & \textcolor{blue}{\underline{0.325}} & \textcolor{blue}{\underline{0.390}} & 0.374 & 0.445 \\
\bottomrule
\end{tabular}%
}
\ifdim\wd0>\linewidth\resizebox{\linewidth}{!}{\usebox0}\else\usebox0\fi
\endgroup
\end{table}

\clearpage
\subsection{Token Budget and Effective Context Length}
\label{app:token-budget}

\begin{figure}[!htbp]
    \centering
    \includegraphics[width=\linewidth]{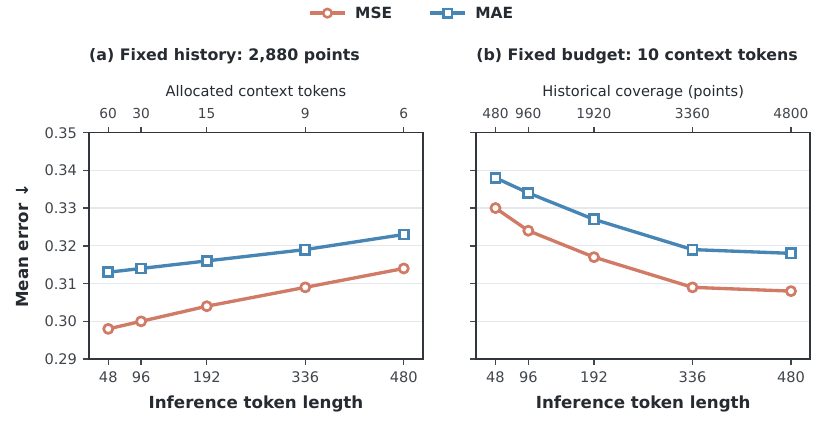}
    \caption{TFB token-budget study. (a) A fixed 2,880-point input with decreasing context-token usage. (b) Ten context tokens with increasing historical coverage. Scores represent averages over eight datasets and four forecast horizons; lower is better.}
    \label{fig:token-budget}
\end{figure}

Resolution-adaptive inference offers two ways to allocate a context budget: compress a given history into fewer tokens, or expose the model to a longer history without increasing the number of context tokens. We consider both settings on the eight TFB datasets, keeping the model checkpoint and forecast horizons fixed. Figure~\ref{fig:token-budget} illustrates the two operating modes.

\paragraph{Evaluation design.}
We vary the temporal span per token, $s$, over $\{48,96,192,336,480\}$. Each token represents $s$ original time points resampled to the native patch length of 48. Forecasts are restored to the original grid before scoring. MSE and MAE are averaged over the four horizons within each dataset, then equally across the eight datasets. All configurations share forecast origins with sufficient observed history, input settings, and quantile queries. Token counts are per variable.

\paragraph{Compressing a fixed history.}
With 2,880 supplied observations, the five spans allocate 60, 30, 15, 9, and 6 context-token slots, respectively. Increasing $s$ maps more original points to each native 48-point patch, lowering temporal resolution while preserving embedding dimensions and model weights. A history of $T$ points therefore uses $N_s=\lceil T/s\rceil$ context slots. Fewer tokens reduce temporal-attention interactions and token-wise projection, routing, and expert computation. This sweep tests whether coarser representations retain predictive information as context-token usage decreases by up to 90\%, using native resolution as the accuracy reference. The same span applies to every dataset, including Solar. At $s=336$, the incomplete leading patch is masked: eight valid patches encode the latest 2,688 points, while normalization uses all 2,880 supplied observations.

\paragraph{Extending history at a fixed context budget.}
With ten context tokens, increasing the span exposes the model to 480, 960, 1,920, 3,360, and 4,800 historical points. This sweep examines whether broader temporal coverage supplies predictive information absent from shorter histories. Together, the two sweeps test the trade-off between temporal detail and historical coverage enabled by variable-resolution post-training. The controlled budget concerns context tokens: for a forecast horizon $H$, the model still allocates $\lceil H/s\rceil$ future tokens. End-to-end compute therefore also depends on the forecast horizon and resampling operations.

\FloatBarrier

\clearpage
\subsection{Inference Efficiency}
\label{app:inference-efficiency}

\paragraph{Experimental setup.}
We evaluate the cost of a complete probabilistic forecast, including the backbone and its native output procedure. The primary comparison uses the same univariate ETT history with context length $L=960$, forecast horizon $H=96$, batch size one, and nine requested quantiles $\{0.1,0.2,\ldots,0.9\}$. Here, $L$ counts original observations, so equal history lengths need not imply equal token counts. All models run in FP32 on one NVIDIA A800 80GB PCIe GPU, with TF32, autocast, and \texttt{torch.compile} disabled. Table~\ref{tab:inference-efficiency} reports six models, including three inference token lengths for Aurora-X.

\begin{table}[!htbp]
\centering
\caption{Inference cost at $L=960$, $H=96$, batch size one, and nine requested quantiles, measured in FP32 on an A800 80GB PCIe GPU. Parentheses give Aurora-X's inference token length. Parameters are total counts; memory is peak allocated GPU memory. Moirai-MoE uses 100 Monte Carlo paths, and TimesFM uses flip averaging.}
\label{tab:inference-efficiency}
\begingroup
\small
\setlength{\tabcolsep}{4pt}
\begin{tabular}{@{}lrrrr@{}}
\toprule
Model & Params (B) & GMACs & Latency (ms) & Memory (GiB) \\
\midrule
Aurora-X (48) & 1.049 & 8.031 & 47.054 & 3.917 \\
Aurora-X (96) & 1.049 & 4.070 & 46.301 & 3.917 \\
Aurora-X (192) & 1.049 & 2.329 & 44.704 & 3.917 \\
\midrule
Toto-2.0-1B & 1.041 & 34.114 & 110.639 & 4.110 \\
Toto-2.0-2.5B & 2.454 & 80.596 & 140.409 & 9.453 \\
Timer-S1 & 8.304 & 32.844 & 107.580 & 31.268 \\
Moirai-MoE-Base & 0.935 & 2883.643 & 829.886 & 3.765 \\
TimesFM-2.5 & 0.231 & 13.963 & 68.689 & 0.881 \\
\bottomrule
\end{tabular}
\endgroup
\end{table}

\paragraph{Native forecasting procedures.}
Aurora-X directly queries the nine IQN levels, while Toto uses its official single-pass parallel forecasting configuration. Timer-S1 retains native generation with RevIN, KV caching, and the executed multi-token prediction layers. Moirai-MoE uses patches of length 16 and 100 Monte Carlo paths to obtain empirical quantiles; at $H=96$, its initial forward pass is followed by five full forward passes over these paths. TimesFM retains its continuous quantile head, flip averaging with two decoding passes, and native output corrections. Only its wrapper-level CPU conversions are removed, with predictions checked against the original API. These settings preserve each model's forecasting procedure; the common quantile request does not equate their predictive distributions or accuracy.

\paragraph{Measurement protocol.}
Matrix multiply--accumulate operations (MACs) are counted from executed linear layers, attention, and routed experts using \texttt{FlopCounterMode} with an independent operator audit. The matrix multiplication within Moirai's distance-based routing is included. Attention counts use complete dense matrix shapes, including masked positions. One GMAC denotes $10^9$ MACs, with two FLOPs per matrix MAC. Normalization, activation, sampling, sorting, and interpolation contribute to latency but are excluded from matrix MACs. Counting uses \texttt{no\_grad}, and its outputs are verified against native inference.

Latency is the median synchronized wall-clock time under \texttt{inference\_mode}, following five warm-up iterations. We use 30 repetitions for Aurora-X's token-length sweep and 20 for each baseline. Inputs are already on the GPU; timing includes normalization, routing, prediction heads, and complete native generation, excluding weight loading and host--device transfers. The measured GPU has no competing compute process. Memory is PyTorch's peak allocated GPU memory, including model parameters and tensors. Parameter counts include all loaded parameters, excluding buffers, rather than only the active MoE parameters.

\clearpage
\begin{figure}[!htbp]
    \centering
    \includegraphics[width=\linewidth]{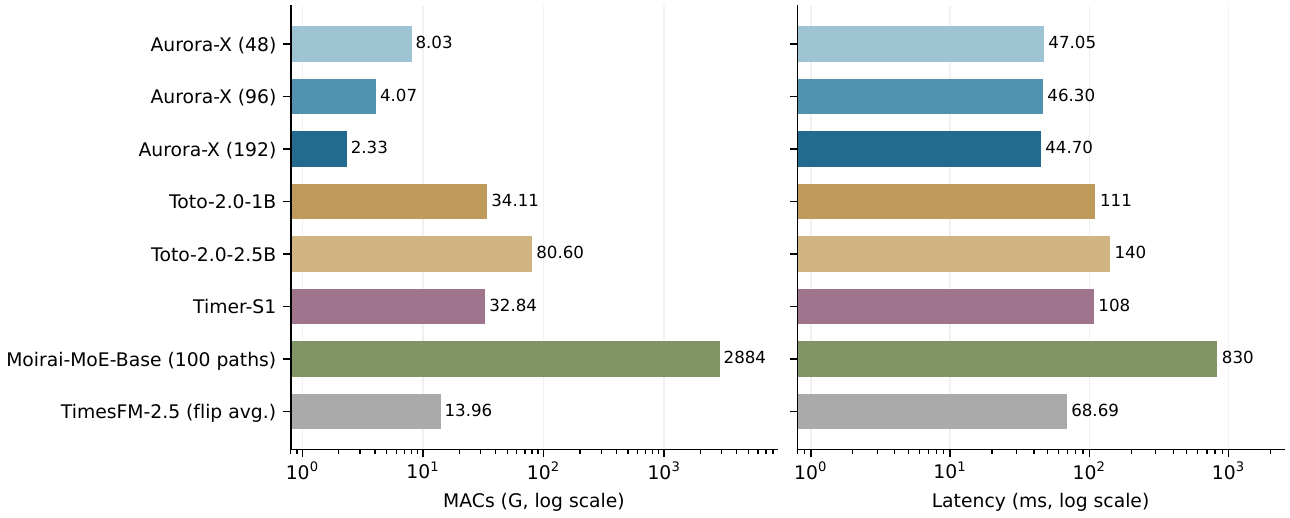}
    \caption{Complete-forecast MACs and median latency for the configurations in Table~\ref{tab:inference-efficiency}. Native output procedures are retained, including Moirai-MoE's 100 sampling paths and TimesFM's flip averaging.}
    \label{fig:inference-efficiency}
\end{figure}

\paragraph{Comparison with pretrained forecasters.}
Figure~\ref{fig:inference-efficiency} compares the resulting computation and latency. With comparable total parameter counts, Aurora-X at its native token length of 48 uses 76.5\% fewer MACs than Toto-2.0-1B and achieves a measured $2.35\times$ speedup. Timer-S1 has a larger total parameter count, whereas TimesFM provides a smaller-model reference. Moirai-MoE's repeated sampling contributes substantially to its complete-forecast cost, so this difference cannot be attributed solely to the backbone architecture.

\begin{figure}[!htbp]
    \centering
    \includegraphics[width=\linewidth]{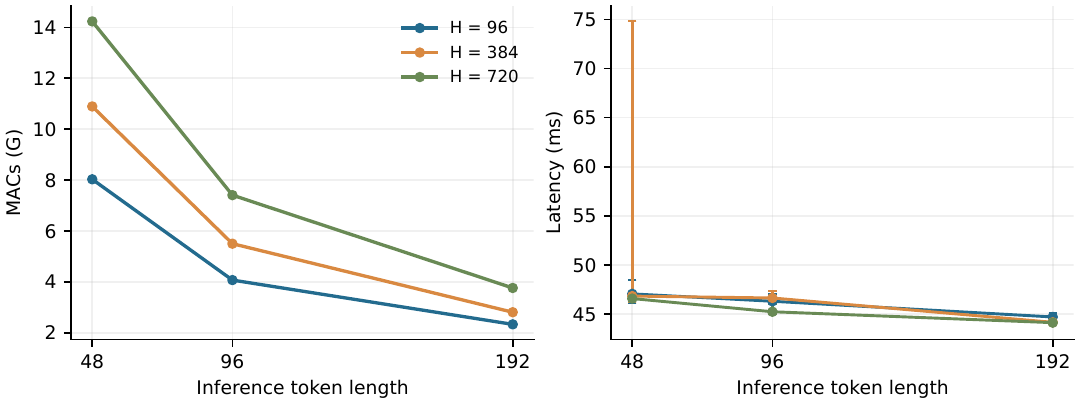}
    \caption{Aurora-X inference cost as the temporal span per token increases, with $L=960$ and three forecast horizons. Each configuration returns nine quantiles. Latency error bars show the empirical 10th--90th percentiles of 30 synchronized runs.}
    \label{fig:inference-token-efficiency}
\end{figure}

\paragraph{Effect of temporal resolution.}
Figure~\ref{fig:inference-token-efficiency} complements the context-budget study in Appendix~\ref{app:token-budget} with measured computation and latency. Longer inference tokens lower temporal resolution by resampling longer blocks to the native patch length of 48. At $L=960$ and $H=96$, lengths 48, 96, and 192 use 20, 10, and 5 context tokens, plus 2, 1, and 1 future tokens, respectively. This shorter backbone sequence reduces attention, routing, and expert computation; output resampling still returns all 96 forecast points. Lengths 96 and 192 reduce MACs by 49.3\% and 71.0\%, respectively, relative to length 48. Native output padding and final truncation remain included in the measurement. The $3.45\times$ reduction in MACs at length 192 yields a much smaller $1.05\times$ latency speedup, demonstrating that matrix computation alone does not determine end-to-end latency. This efficiency sweep does not evaluate the corresponding change in benchmark accuracy.

\FloatBarrier

\clearpage
\subsection{Forecasting with Arbitrary Quantile Queries}
\label{app:forecast-visualizations}

\begin{figure}[!htbp]
    \centering
    \includegraphics[width=\linewidth]{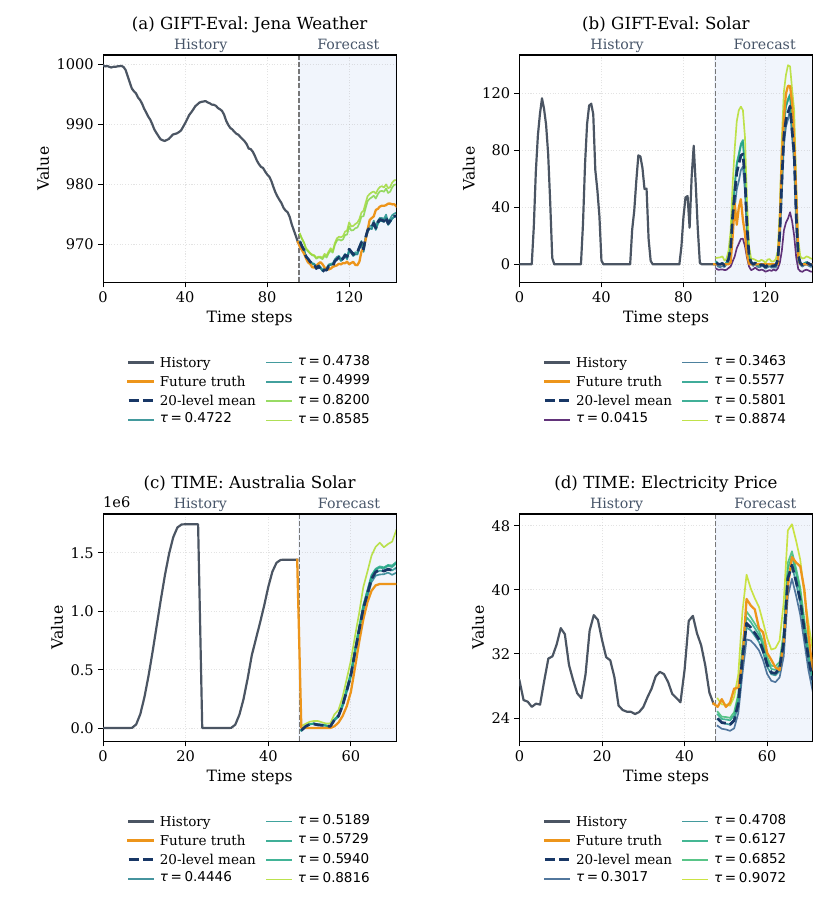}
    \caption{Selected forecasting examples on GIFT-Eval and TIME. Gray and orange denote observed history and future truth. Thin colored curves show five random marginal quantiles; dashed blue shows the independent twenty-level average. The vertical dashed line separates history from forecast. Cases are selected by test error as described in Appendix~\ref{app:forecast-visualizations}.}
    \label{fig:forecast-examples-gift-time}
\end{figure}

\begin{figure}[!htbp]
    \centering
    \includegraphics[width=\linewidth]{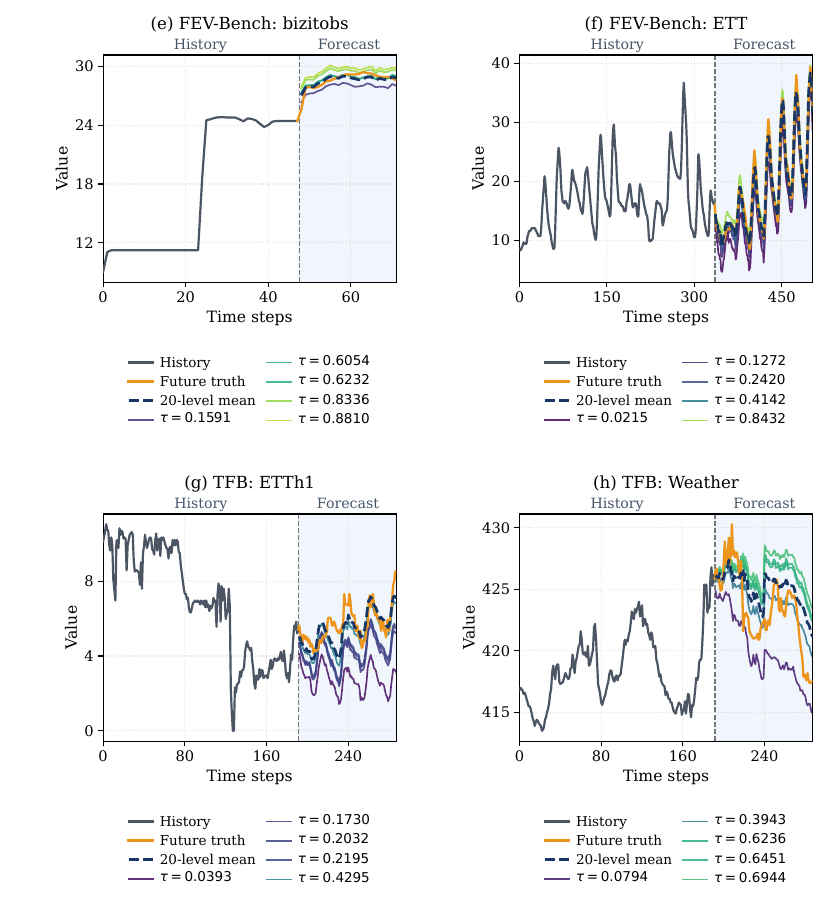}
    \caption{Selected forecasting examples on FEV-Bench and TFB, using the same display and test-error selection protocol as Figure~\ref{fig:forecast-examples-gift-time}. Each panel preserves $2H$ history points, the full $H$-step forecast, and the five originally queried quantile levels. Shading identifies the forecast window.}
    \label{fig:forecast-examples-fev-tfb}
\end{figure}

The implicit quantile head allows users to query forecast levels beyond a predefined output grid. Figures~\ref{fig:forecast-examples-gift-time} and \ref{fig:forecast-examples-fev-tfb} illustrate this interface on eight real forecasting windows from GIFT-Eval, TIME, FEV-Bench, and TFB. Each example combines five directly queried random quantile levels with an independently computed central forecast summary. Appendix~\ref{app:general-forecast-visualizations} complements these examples with point and interval forecasts on all predefined windows.

\paragraph{Case selection.}
We use 24 predefined test windows from twelve dataset configurations. For each window, we score the predetermined display variable by the mean absolute error of its central forecast, divided by the population standard deviation of its cached input history. We retain the lowest-error window per dataset and then the two lowest-error datasets per benchmark; ties use normalized RMSE and the case identifier. This selection uses test targets, yielding illustrative examples with relatively accurate central forecasts rather than an unbiased performance summary. Table~\ref{tab:forecast-examples} lists the resulting inputs.

\begin{table}[!htbp]
\centering
\caption{Input configurations of the eight selected forecasting examples. $L$ is cached input length, $H$ is the forecast horizon, and $C$ is the number of jointly predicted target variables. Each plot displays the listed variable and the last $2H$ history points.}
\label{tab:forecast-examples}
\begingroup\small
\setlength{\tabcolsep}{4pt}
\begin{tabular}{@{}llllrrr@{}}
\toprule
Panel & Benchmark & Dataset & Variable & $L$ & $H$ & $C$ \\
\midrule
(a) & \multirow{2}{*}{GIFT-Eval} & jena\_weather/H & var\_0 & 8112 & 48 & 21 \\
(b) &  & solar/H & var\_0 & 7992 & 48 & 1 \\
\addlinespace[2pt]
(c) & \multirow{2}{*}{TIME} & Australia\_Solar/H & solar\_63726 & 2880 & 24 & 3 \\
(d) &  & epf\_electricity\_price/H & var\_0 & 2880 & 24 & 1 \\
\addlinespace[2pt]
(e) & \multirow{2}{*}{FEV-Bench} & bizitobs\_l2c\_1H & target\_0 & 2280 & 24 & 7 \\
(f) &  & ETT\_1H & OT & 8160 & 168 & 7 \\
\addlinespace[2pt]
(g) & \multirow{2}{*}{TFB} & ETTh1 & OT & 528 & 96 & 7 \\
(h) &  & Weather & OT & 528 & 96 & 21 \\
\bottomrule
\end{tabular}
\endgroup
\end{table}

\paragraph{Quantile queries and central forecasts.}
For each case, five levels are sampled independently from $\mathrm{Uniform}(0.02,0.98)$, rounded to four decimal places, and queried directly through the IQN head. We use a fixed global seed of 20260921 with deterministic case-specific seeds. Separately, for each forecast step $h$, we average predictions at twenty equally spaced levels in $[0.05,0.95]$:
\begin{equation}
    \bar y_h^{(20)}
    =\frac{1}{20}\sum_{j=0}^{19}
    \widehat Q_h\!\left(0.05+\frac{0.90j}{19}\right).
    \label{eq:visualization-mean}
\end{equation}
Here $\widehat Q_h(\tau)$ denotes the predicted marginal quantile at level $\tau$. This finite-grid average is neither the median nor an exact integral over the full predictive distribution. The five random levels do not enter its calculation. Requested levels are sorted for display, while the predicted values and any quantile crossings are preserved.

\paragraph{Inference and interpretation.}
All eight cases use the same Aurora-X checkpoint in FP32, with joint target-variable attention and an inference token length of 48. Model inputs retain the cached histories in Table~\ref{tab:forecast-examples}; incomplete leading patches follow native masking. TFB inputs are standardized using training-split statistics, and predictions are restored to their original units. Plots show only the last $2H$ observed steps before the forecast origin. Solar examples exhibit larger separation between quantile curves near production peaks, whereas the displayed bizitobs window has a narrower spread. These views illustrate how directly queried marginal forecasts vary with the input and forecast step. The curves are not joint trajectory samples, and the selected examples do not establish distributional calibration.

\clearpage

\clearpage
\FloatBarrier

\subsection{Additional Forecasting Visualizations}
\label{app:general-forecast-visualizations}

\begin{figure}[!htbp]
    \centering
    \includegraphics[width=\linewidth]{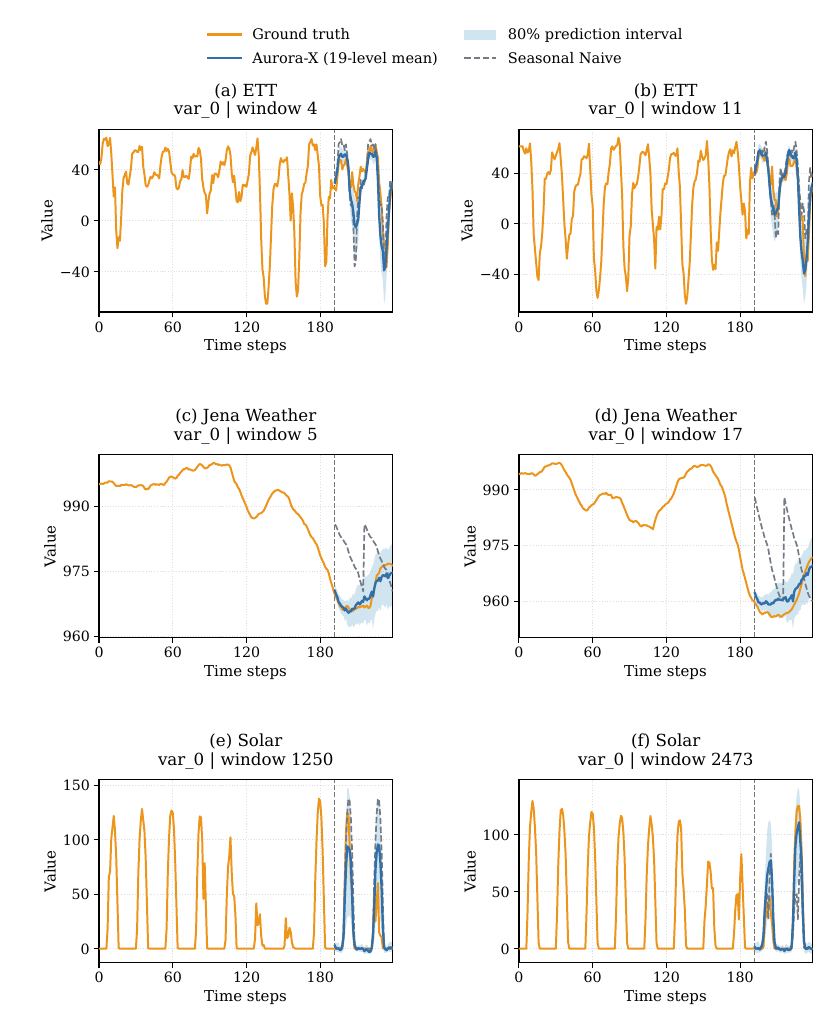}
    \caption{Forecasts on six predefined GIFT-Eval windows from ETT, Jena Weather, and Solar, with horizon 48. Orange shows observed history and future truth; blue is Aurora-X's nineteen-level mean; gray dashed curves show Seasonal Naive. Shading denotes the P10--P90 prediction interval. The vertical dashed line marks the forecast origin. Each panel displays 192 history points from the cached input.}
    \label{fig:general-forecasts-gift}
\end{figure}

\begin{figure}[!htbp]
    \centering
    \includegraphics[width=\linewidth]{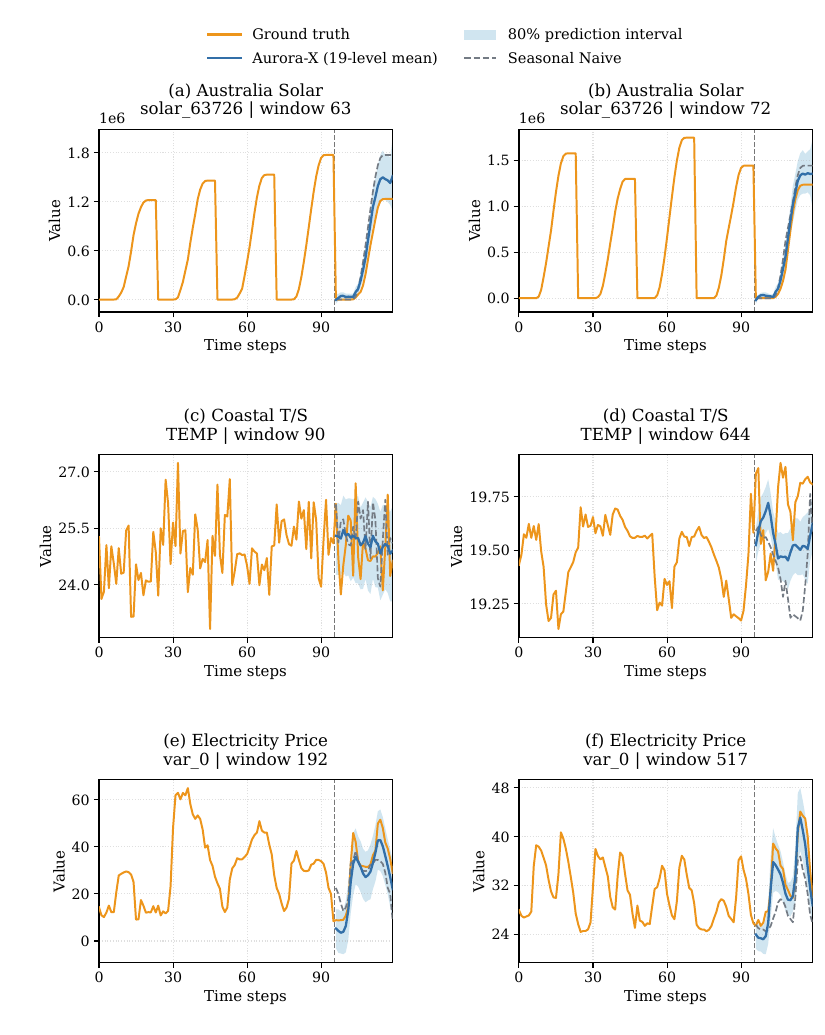}
    \caption{Forecasts on six predefined TIME windows from Australia Solar, Coastal T/S, and electricity prices. Models receive 2,880 history points and predict 24 steps; each panel displays the final 96 history points. Curves and intervals follow Figure~\ref{fig:general-forecasts-gift}. These windows show both close forecasts and departures from the observed future, particularly for irregular fluctuations.}
    \label{fig:general-forecasts-time}
\end{figure}

\begin{figure}[!htbp]
    \centering
    \includegraphics[width=\linewidth]{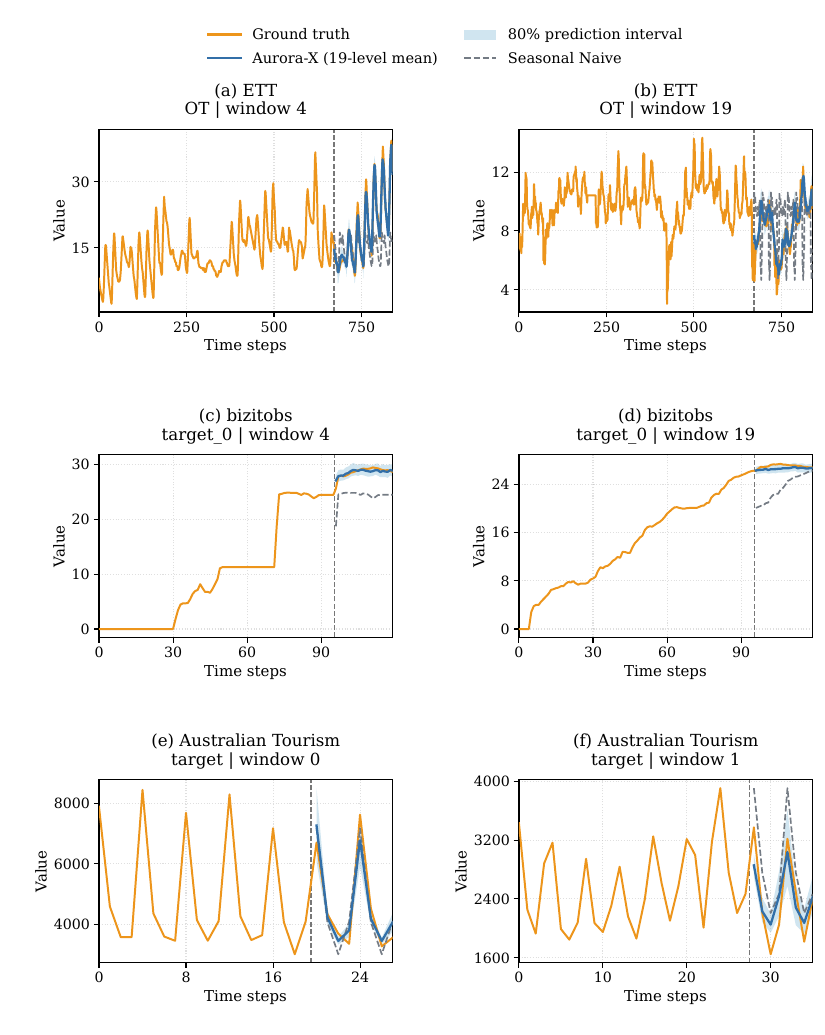}
    \caption{Forecasts on six predefined FEV-Bench windows. Horizons are 168 for ETT, 24 for bizitobs, and 8 for Australian Tourism. The corresponding displays retain 672, 96, and 20 or 28 historical points, respectively. Curves and intervals follow Figure~\ref{fig:general-forecasts-gift}; the Tourism cases illustrate prediction from short observed histories.}
    \label{fig:general-forecasts-fev}
\end{figure}

\begin{figure}[!htbp]
    \centering
    \includegraphics[width=\linewidth]{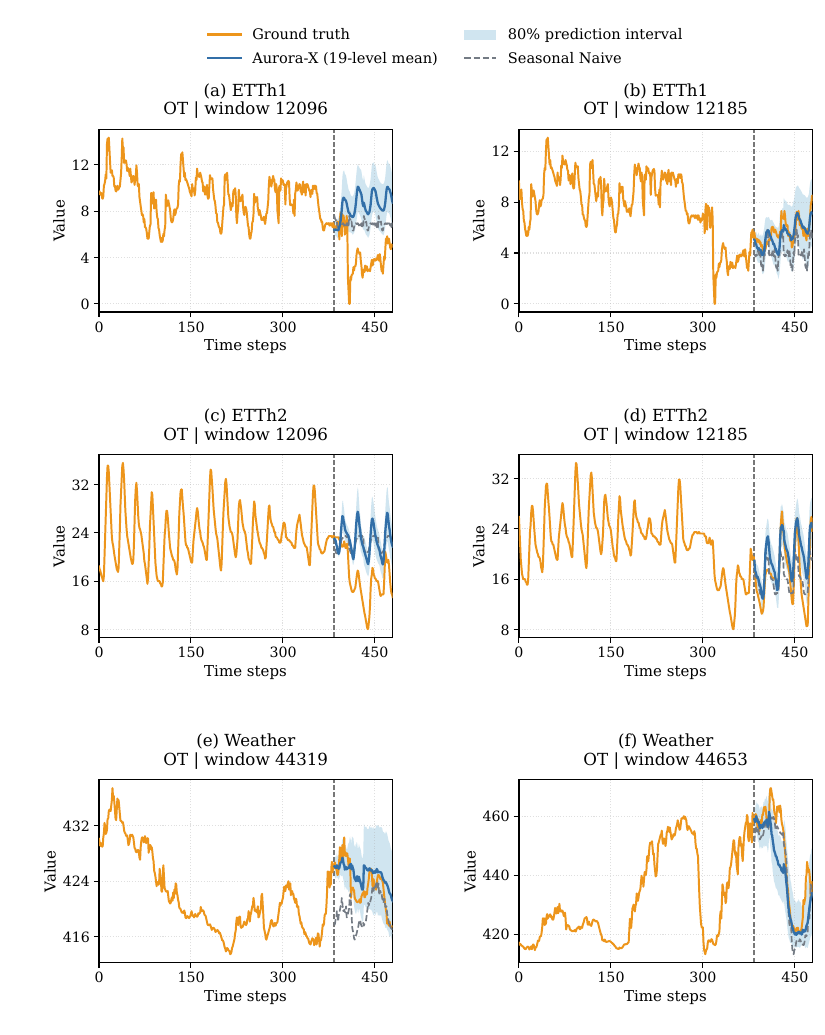}
    \caption{Forecasts on six predefined TFB windows from ETTh1, ETTh2, and Weather, with horizon 96. Each model input contains 528 history points, of which the final 384 are shown. Predictions are restored from training-split standardization to original units. Curves and intervals follow Figure~\ref{fig:general-forecasts-gift}.}
    \label{fig:general-forecasts-tfb}
\end{figure}

Figures~\ref{fig:general-forecasts-gift}--\ref{fig:general-forecasts-tfb} show all 24 predefined test windows from twelve dataset configurations, with two windows per configuration. Windows and display variables were fixed before inference, without filtering by forecast error. Each panel compares Aurora-X with Seasonal Naive, which repeats the latest observed seasonal cycle. The blue forecast averages nineteen queried levels, $0.05,0.10,\ldots,0.95$; the shaded band spans P10--P90 and denotes a nominal 80\% prediction interval. Only these two endpoints are reordered when necessary for plotting. Inputs retain all cached history and jointly modeled targets; displayed values are in original units. The inference token length is 48, except for the two short-history Australian Tourism cases, which use 8. These views complement the arbitrary-quantile examples with recurring cycles, level changes, and irregular fluctuations; empirical interval coverage varies across windows.

\clearpage

\clearpage

\clearpage

\clearpage
\FloatBarrier

\clearpage
\section{Resolution Adaptation and Training Consistency}
\label{app:resolution-equivalence}

Resolution adaptation changes the temporal span represented by each token while retaining a pretrained backbone. This section relates our explicit patch resampling to the projection resizing used in LightGTS~\citep{wang2025lightgts}, then explains how the data-side formulation supports consistent multi-scale training. We compare resolution operators around a shared base model, with matched patch boundaries, normalization, and auxiliary inputs.

\subsection{Equivalent Placement of Resolution Operators}
\label{app:resolution-placement}

Let $P$ be the native patch length and $P_k=kP\in\mathbb N$ the requested temporal span. For a normalized input patch $x_k\in\mathbb R^{P_k}$ and a native-grid output patch $y\in\mathbb R^P$, write linear resampling in matrix form:
\begin{equation}
\begin{aligned}
    D_kx_k&=\mathcal R_{P_k\rightarrow P}(x_k),& D_k&\in\mathbb R^{P\times P_k},\\
    U_ky&=\mathcal R_{P\rightarrow P_k}(y),& U_k&\in\mathbb R^{P_k\times P}.
\end{aligned}
\label{eq:resolution-matrices}
\end{equation}
The matrices use fixed interpolation coordinates and boundary conventions. $D_k$ maps observations to the native grid, and $U_k$ restores predictions to the requested grid; they need not be inverses.

\paragraph{Proposition 1 (data--parameter equivalence).}
For an input projection $W\in\mathbb R^{d\times P}$ with bias $b$, resampling the patch is exactly equivalent to replacing its value projection by $W_k=WD_k$. Likewise, for an output projection $V\in\mathbb R^{P\times d}$ with bias $c$, output resampling is equivalent to $V_k=U_kV$ and $c_k=U_kc$:
\begin{equation}
\begin{aligned}
    W(D_kx_k)+b&=(WD_k)x_k+b,\\
    U_k(Vh+c)&=(U_kV)h+U_kc.
\end{aligned}
\label{eq:resolution-placement}
\end{equation}

\paragraph{Proof.}
Both identities follow from associativity and distributivity of matrix multiplication. In Aurora-X's residual MLP embedding, the same composition is applied to the value columns of both input affine branches; time and mask features are held fixed. Both branches then receive identical pre-activations, so the subsequent nonlinearities, expert routing, and backbone states are identical. On the output side, composing $U_k$ with the quantile head's final affine projection gives the same restored predictions for every quantile query. Thus, the equivalence extends through nonlinear modules when the linear resolution maps are placed at their boundaries.

\subsection{Relationship to LightGTS Flex-Resize}
\label{app:lightgts-flex}

LightGTS derives a particular projection transformation using a Moore--Penrose pseudoinverse. With matched interpolation conventions, converting its row-vector notation to the column-vector convention above gives
\begin{equation}
    R_k^{\mathrm{Flex}}=\delta_k^{-1}U_k^{+},\qquad
    W_k^{\mathrm{Flex}}=WR_k^{\mathrm{Flex}},\qquad
    \delta_k=\sqrt{P/P_k}.
    \label{eq:lightgts-flex}
\end{equation}
Here $\delta_k$ is the scale factor specified by LightGTS. Consequently, Flex-resize itself has an exact data-side realization:
\begin{equation}
    W_k^{\mathrm{Flex}}x_k+b=W(R_k^{\mathrm{Flex}}x_k)+b.
    \label{eq:flex-data-equivalence}
\end{equation}
Equation~(\ref{eq:flex-data-equivalence}) reveals the resampling implicit in parameter resizing. The vector $U_k^{+}x_k$ is the minimum-norm least-squares reconstruction of $P$ native-grid values whose interpolation by $U_k$ best fits the $P_k$ observations. For $P_k>P$, it reduces the sample count; for $P_k<P$, it reconstructs a denser native grid. Thus, Flex-resize implements an interpolation-based upsampling or downsampling operation on the input, followed by the scale factor $\delta_k^{-1}$, while keeping that operation implicit in the projection weights. Its pseudoinverse-derived reconstruction weights need not equal ordinary interpolation weights.

\paragraph{Proposition 2 (conditions for agreement with Flex-resize).}
For a fixed value projection $W$, direct resampling and Flex-resize produce identical outputs of this affine projection for every $x_k$ if and only if $W(D_k-R_k^{\mathrm{Flex}})=0$. Agreement for every value projection requires $D_k=R_k^{\mathrm{Flex}}$. Their projection discrepancy satisfies
\begin{equation}
\begin{aligned}
    z_k^{\mathrm{direct}}-z_k^{\mathrm{Flex}}
        &=W(D_k-R_k^{\mathrm{Flex}})x_k,\\
    \|z_k^{\mathrm{direct}}-z_k^{\mathrm{Flex}}\|_2
        &\leq\|W\|_2\,\|D_k-R_k^{\mathrm{Flex}}\|_2\,\|x_k\|_2.
\end{aligned}
\label{eq:flex-discrepancy}
\end{equation}

\paragraph{Proof.}
Subtracting the two affine projections cancels their common bias and yields the first identity. Equality for all patches is equivalent to the resulting linear map being zero. Requiring equality for all $W$ further forces $D_k=R_k^{\mathrm{Flex}}$. The bound follows from submultiplicativity of the spectral norm.

Aurora-X uses ordinary linear interpolation for $D_k$, whereas LightGTS uses the scaled pseudoinverse $R_k^{\mathrm{Flex}}$. Reverse interpolation is generally different from a pseudoinverse. Accordingly, the exact equivalence concerns placement of a matched operator; approximation to the particular LightGTS operator depends on the discrepancy in Equation~(\ref{eq:flex-discrepancy}).

For a patch generated as $x_k=U_kx_0$, suppressing the common affine bias and explicitly aligning LightGTS's scalar factor gives
\begin{equation}
    z_k^{\mathrm{direct}}-\delta_k z_k^{\mathrm{Flex}}
        =W(D_kU_k-U_k^{+}U_k)x_0.
    \label{eq:resolution-roundtrip}
\end{equation}
If $U_k$ has full column rank, then $U_k^{+}U_k=I_P$: the remaining difference is the round-trip interpolation error measured by $W$. It vanishes on signals recovered by $D_kU_k$ and is small when their recovery error is small. Otherwise, $U_k^{+}U_k$ projects the native signal onto the recoverable subspace. The scalar factor remains part of the implemented Flex-resize; Equation~(\ref{eq:resolution-roundtrip}) explicitly aligns it for this comparison.

\subsection{Consistency of Multi-Scale Training and Inference}

The data-side formulation makes the same resolution transformation available for constructing training examples. Let $\mathcal N_X$ denote RevIN followed by $\operatorname{arcsinh}$, using statistics from the original observed history $X$. For known future covariates $C$ and targets $Y$, construct
\begin{equation}
    X_k=\mathcal A_k(\mathcal N_X(X)),\qquad
    C_k=\mathcal A_k(\mathcal N_X(C)),\qquad
    Y_k=\mathcal A_k(\mathcal N_X(Y)),
    \label{eq:paired-resolution-data}
\end{equation}
where statistics are applied variable-wise and $\mathcal A_k$ applies $D_k$ separately to each patch. Each augmented history--target pair therefore uses the same temporal granularity, with $P_k$ original points represented by $P$ native-grid values.

Let $F_\theta$ denote the native-grid forecaster, including covariate conditioning and a quantile query $\tau$, and let $\mathcal B_k$ apply $U_k$ to each output patch. The corresponding training objective and inference computation are
\begin{equation}
\begin{aligned}
    \mathcal J(\theta)
        &=\mathbb E_{(X,C,Y),k,\tau}
          \left[\ell_\tau\!\left(F_\theta(X_k,C_k;\tau),Y_k\right)\right],\\
    \widehat Y_k(\tau)
        &=\mathcal N_X^{-1}\!\left(\mathcal B_k F_\theta(X_k,C_k;\tau)\right),
\end{aligned}
\label{eq:resolution-train-infer}
\end{equation}
where $\ell_\tau$ averages the masked pinball loss over valid target positions, as in Equation~(\ref{eq:post-training-loss}). Both computations call the same forecaster on the same canonical inputs. Their consistency follows directly from reusing the augmentation operator in inference, with supervision defined on that canonical grid.

This construction fixes the order of preprocessing: history-based normalization and $\operatorname{arcsinh}$ precede resampling, and the augmented arrays are not normalized again. History blocks are aligned to the forecast origin, and future blocks begin at the first forecast step. History statistics and resampling do not depend on future targets; targets are transformed separately using those history statistics. Related variables share scale and block boundaries. Alignment requires identical validity rules and native-grid time features during training and inference. These choices preserve the input interface even when patches contain different amounts of historical information.

\paragraph{Implications for foundation-model training.}
The operator equivalence allows resolution adaptation to be expressed as paired data augmentation while retaining a fixed model interface. Multiple resolutions can be prepared in the data pipeline and batched using the same patch shape, embedding, quantile head, and loss. Scales can be enumerated without requiring a detected period, and the same transformation aligns histories, covariates, and targets. LightGTS also learns across varying periods through its adaptive projections; our design emphasizes explicit paired augmentation and dedicated variable-resolution post-training. This uniform data construction provides direct training--inference alignment and a reusable augmentation pipeline, making it well suited to scaling foundation-model training over heterogeneous sequences.

\clearpage
\section{Theoretical Analysis of Shallow-Pattern Clustering}
\label{app:clustering-theory}

Expert specialization requires a criterion for deciding which patches should share computation. Similarity-based channel clustering~\citep{chen2024similarity} motivates using temporal structure to guide parameter sharing. We develop this connection for Aurora-X's normalized, overlapping expert assignments: the pattern loss measures cluster coherence, its gradient favors sharing among sufficiently similar patches, and the projection regularizer encourages distinct feature directions.

\subsection{Normalized Co-Assignment and Kernel Clustering}
\label{app:cluster-geometry}

Consider one nonempty context with $N$ patches at one MoE layer, and omit batch and layer indices. Let $D\in\{0,1\}^{N\times E}$ contain the hard top-$K$ assignments, with $\sum_eD_{ie}=K$, $1\leq K\leq E$, and $\sigma_{\mathrm{pat}}>0$. The shared expert is excluded. Define
\begin{equation}
    \Omega=\frac{DD^\top}{K},\qquad
    Z_\Omega=\sum_{i,j}\Omega_{ij},\qquad
    \rho=\frac{\sum_{i,j}\Omega_{ij}S_{ij}}{Z_\Omega},\qquad
    \ell_{\mathrm{pat}}=-\rho.
    \label{eq:cluster-normalized-affinity}
\end{equation}
This is the per-context term of Equation~(\ref{eq:pattern-loss}), including self-pairs. Since $\Omega_{ii}=1$, $Z_\Omega\geq N$, so a denominator floor smaller than one is inactive. For the kernel geometry below, assume that projected patch norms exceed the normalization floor, giving $\|u_i\|_2=1$. Their similarity then satisfies
\begin{equation}
    S_{ij}=\exp\!\left(-\frac{\|u_i-u_j\|_2^2}{2\sigma_{\mathrm{pat}}^2}\right)
    =\langle\phi_i,\phi_j\rangle_{\mathcal H},\qquad
    \|\phi_i\|_{\mathcal H}^2=1,
    \label{eq:cluster-kernel-map}
\end{equation}
where $\phi_i$ is a feature representation of the Gaussian kernel in a Hilbert space $\mathcal H$.

\paragraph{Proposition \thesection.1 (within-expert kernel dispersion).}
Let $\mathcal C_e=\{i:D_{ie}=1\}$, $n_e=|\mathcal C_e|$, and, for each nonempty expert assignment, define its feature mean and variance by
\[
    \mu_e=\frac{1}{n_e}\sum_{i\in\mathcal C_e}\phi_i,
    \qquad
    V_e=\frac{1}{n_e}\sum_{i\in\mathcal C_e}\|\phi_i-\mu_e\|_{\mathcal H}^2.
\]
Then the clustering objective admits the exact decomposition
\begin{equation}
    1+\ell_{\mathrm{pat}}
    =\frac{\sum_{e:n_e>0}n_e^2V_e}{\sum_e n_e^2}.
    \label{eq:cluster-dispersion}
\end{equation}

\paragraph{Proof.}
Expanding the variance gives $V_e=1-\|\mu_e\|_{\mathcal H}^2$. Moreover,
\[
    KZ_\Omega=\sum_e n_e^2,\qquad
    K\sum_{i,j}\Omega_{ij}S_{ij}
    =\sum_e\left\|\sum_{i\in\mathcal C_e}\phi_i\right\|_{\mathcal H}^2
    =\sum_{e:n_e>0}n_e^2\|\mu_e\|_{\mathcal H}^2.
\]
Substituting these identities into Equation~(\ref{eq:cluster-normalized-affinity}) proves the result. Each patch may belong to several $\mathcal C_e$, so the identity applies directly to top-$K$ routing.

Thus, the loss measures a weighted average of within-expert pattern dispersion. Experts receive weights proportional to $n_e^2$, reflecting their numbers of co-assigned ordered pairs. It encourages coherent expert assignments, while allowing overlapping pattern groups. For zero or sub-unit normalized features, use the feature map of the positive-semidefinite kernel $k(u,v)=\exp[(u^\top v-1)/\sigma_{\mathrm{pat}}^2]$. The same expansion adds the diagonal-deficit term $\sum_e n_e\sum_{i\in\mathcal C_e}(1-S_{ii})/\sum_e n_e^2$ to Equation~(\ref{eq:cluster-dispersion}). The pure variance interpretation holds under the unit-norm condition.

\paragraph{Corollary (control of dissimilar co-assignments).}
For $0<\delta\leq2$, let $q_\delta$ be the share of total co-assignment weight allocated to pairs with $\|u_i-u_j\|_2\geq\delta$. Under the unit-norm condition,
\begin{equation}
    q_\delta=
    \frac{\sum_{i,j}\Omega_{ij}\mathbf1[\|u_i-u_j\|_2\geq\delta]}{Z_\Omega}
    \leq
    \frac{1+\ell_{\mathrm{pat}}}
    {1-\exp[-\delta^2/(2\sigma_{\mathrm{pat}}^2)]}.
    \label{eq:cluster-distant-pairs}
\end{equation}
Indeed, $1+\ell_{\mathrm{pat}}=\sum_{i,j}\Omega_{ij}(1-S_{ij})/Z_\Omega$, and each such pair contributes at least $1-\exp[-\delta^2/(2\sigma_{\mathrm{pat}}^2)]$. The bound quantifies how a loss close to $-1$ restricts dissimilar pairs that share experts; its weights include self-pairs, as in the implemented objective.

\subsection{Pattern-Dependent Learning Signals for Routing}
\label{app:cluster-gradient}

The normalization by $Z_\Omega$ makes the routing signal depend on relative pattern similarity. To characterize this signal, hold $S$ fixed and consider the continuous extension of Equation~(\ref{eq:cluster-normalized-affinity}).

\paragraph{Proposition \thesection.2 (normalized affinity and straight-through gradients).}
The derivative with respect to an overlap entry is
\begin{equation}
    \frac{\partial\ell_{\mathrm{pat}}}{\partial\Omega_{ij}}
    =-\frac{S_{ij}-\rho}{Z_\Omega}.
    \label{eq:cluster-overlap-gradient}
\end{equation}
For the actual straight-through assignment in Equation~(\ref{eq:routing-overlap}), let $p_i=\operatorname{Softmax}(s_i)$, where $s_i$ is the normalized router-logit vector. Define the affinity of patch $i$ to expert $e$'s assigned patches, centered at the current mean similarity, as
\[
    c_{ie}=\sum_jD_{je}(S_{ij}-\rho).
\]
The backward derivative with respect to $s_{ie}$ is
\begin{equation}
    \frac{\partial\ell_{\mathrm{pat}}}{\partial s_{ie}}
    =-\frac{2p_{ie}}{KZ_\Omega}
    \left(c_{ie}-\sum_f p_{if}c_{if}\right).
    \label{eq:cluster-router-gradient}
\end{equation}

\paragraph{Proof.}
The quotient rule gives Equation~(\ref{eq:cluster-overlap-gradient}). Writing $\Omega=AA^\top/K$, symmetry of $S$ yields
\[
    \left.\frac{\partial\ell_{\mathrm{pat}}}{\partial A}\right|_{A=D}
    =-\frac{2}{KZ_\Omega}
    (S-\rho\mathbf1\mathbf1^\top)D.
\]
The straight-through construction has $\partial A/\partial p=I$ in the backward pass. Applying the softmax Jacobian $\operatorname{diag}(p_i)-p_ip_i^\top$ gives Equation~(\ref{eq:cluster-router-gradient}).

Equation~(\ref{eq:cluster-overlap-gradient}) favors increasing overlap for pairs more similar than the current co-assignment average and decreasing it for less similar pairs. Equation~(\ref{eq:cluster-router-gradient}) describes how this preference reaches the router: the partial derivative with respect to an expert's normalized logit is negative when its pattern-affinity score exceeds the probability-weighted mean across experts. Gradients subsequently pass through logit normalization and the shared routing parameters. They are the implemented straight-through surrogate gradients, rather than derivatives of the discrete top-$K$ map.

The similarity matrix is computed from shallow patches, so its pattern distinctions do not directly depend on distinctions retained in deep hidden states. In addition, for fixed assignments, $\partial\ell_{\mathrm{pat}}/\partial S_{ij}=-\Omega_{ij}/Z_\Omega$: the learned pattern projection receives a signal to increase similarity among co-assigned patches. These two paths couple expert sharing with learning a local pattern metric.

\subsection{When Pattern Grouping Is Preferred to Collapsed Routing}
\label{app:cluster-collapse}

The preceding identities characterize local training signals. The following comparison states when the objective itself favors differentiated assignments over sending every patch to the same experts.

\paragraph{Proposition \thesection.3 (objective gap for pattern-separated routing).}
Partition the patches into $G\geq2$ nonempty pattern groups of sizes $n_g$, and suppose $E\geq GK$. Hold $S$ fixed. Let $\mu_{\mathrm{in}}$ and $\mu_{\mathrm{out}}$ be the mean similarities over within-group and between-group ordered pairs, respectively, including self-pairs within groups. Compare two feasible assignments: each group selects its own disjoint set of $K$ experts, or every patch selects the same $K$ experts. Their pattern losses satisfy
\begin{equation}
    \ell_{\mathrm{collapse}}-\ell_{\mathrm{group}}
    =\left(1-\frac{\sum_gn_g^2}{N^2}\right)
    (\mu_{\mathrm{in}}-\mu_{\mathrm{out}}).
    \label{eq:cluster-collapse-gap}
\end{equation}
Consequently, pattern-separated routing has strictly lower loss when $\mu_{\mathrm{in}}>\mu_{\mathrm{out}}$.

\paragraph{Proof.}
With disjoint expert sets, $\Omega_{ij}=1$ within a group and zero between groups, so $\ell_{\mathrm{group}}=-\mu_{\mathrm{in}}$. Under complete routing collapse, $\Omega_{ij}=1$ for every pair. Its mean similarity is therefore $[\sum_g n_g^2\mu_{\mathrm{in}}+(N^2-\sum_gn_g^2)\mu_{\mathrm{out}}]/N^2$. Subtracting the losses gives Equation~(\ref{eq:cluster-collapse-gap}).

This establishes a preference of the pattern objective under explicit capacity and similarity conditions. It does not assert that discrete-routing optimization always escapes collapse. For example, take two patches with distinct unit pattern features, $K=1$, $E\geq2$, and identical hard assignments. The two rows of $S$ have equal sums, so Equation~(\ref{eq:cluster-router-gradient}) gives zero routing gradient, even though separate assignments have lower pattern loss. Pattern coherence and optimization dynamics are distinct questions.

\subsection{Projection Diversity and the Role of Load Balancing}
\label{app:cluster-projection}

Learning the metric together with routing makes projection diversity relevant: repeated projection directions reduce the information available for comparing patches. The orthogonality term in Equation~(\ref{eq:pattern-loss}) penalizes this degeneracy.

\paragraph{Proposition \thesection.4 (rank constraint induced by row orthogonality).}
Let $m=d_{\mathrm{pat}}$, assume the rows of $\bar W_{\mathrm{pat}}$ have unit norm, and let $r=\operatorname{rank}(\bar W_{\mathrm{pat}})$. Then
\begin{equation}
    \mathcal R_{\mathrm{orth}}
    =\frac{\|\bar W_{\mathrm{pat}}\bar W_{\mathrm{pat}}^\top-I_m\|_F^2}{m^2}
    \geq\frac{1}{r}-\frac{1}{m}.
    \label{eq:cluster-rank-bound}
\end{equation}
In particular, when $m>1$, a penalty below $1/[m(m-1)]$ implies full row rank, and collinear rows incur a penalty of at least $1-1/m$.

\paragraph{Proof.}
Let $G_W=\bar W_{\mathrm{pat}}\bar W_{\mathrm{pat}}^\top$. Its trace is $m$, and it has $r$ positive eigenvalues. Cauchy--Schwarz gives $\|G_W\|_F^2\geq m^2/r$. Thus,
\[
    m^2\mathcal R_{\mathrm{orth}}
    =\|G_W\|_F^2-2\operatorname{tr}(G_W)+m
    =\|G_W\|_F^2-m
    \geq m^2/r-m.
\]
The stated consequences follow by setting $r\leq m-1$ or $r=1$.

This result concerns the diversity of projection directions; distinct inputs can still have the same normalized projection. Load balancing addresses a different aspect of routing. The auxiliary loss depends on marginal expert usage and mean routing probabilities, both of which remain unchanged when token-wise assignments and probabilities are jointly permuted. With the patch similarities held fixed, such a permutation can change the pattern loss. Load balancing therefore acts on marginal utilization, while shallow-pattern clustering guides which patches share experts and orthogonality encourages a richer comparison space.

\clearpage
\section{Additional Related Work}
\label{app:additional-related-work}

\paragraph{Conditioning on structured inputs.}
Aurora-X's progression from individual channels to jointly modeled variables connects to several approaches to dependency learning: DUET organizes multivariate forecasting through dual clustering~\citep{qiu2025duet}, CCD captures cross-correlations with deformable convolutional networks~\citep{cheng2026ccd}, and CoRA introduces a correlation-aware adapter for pretrained time series models~\citep{cheng2026cora}. When external predictors are available, GCGNet~\citep{li2026gcgnet} and KITE~\citep{cheng2026kite} incorporate exogenous variables into graph-consistent generation and knowledge-guided probabilistic forecasting, respectively. The distinction between jointly observed targets and auxiliary predictors is particularly relevant to Aurora-X's staged development of multivariate and covariate capabilities.

Input conditioning has a wider interpretation in visual computing. Pose instructions guide text-to-video generation~\citep{ma2024followpose}, portrait animation admits fine-grained expressive control~\citep{ma2024followyouremoji}, and streaming relighting supports interactive adjustments~\citep{ma2026livelight}; these examples belong to the larger field reviewed in a survey of video controllability~\citep{ma2025controllable}. They illustrate the variety of signals that can specify an output. For Aurora-X, the conditioning interface is defined by target histories, related channels, and available future covariates.

\paragraph{Training across observation regimes.}
Heterogeneity is not limited to the number of channels. ASTGI addresses irregular multivariate observations using graph interactions over space and time~\citep{liu2026astgi}, while a joint temporal--spectral framework~\citep{qiu2026bridging} and a deliberately simple baseline~\citep{liu2026apn} investigate the same forecasting regime from different modeling perspectives. Incomplete inputs introduce a related challenge, examined by GinAR+~\citep{11002729}; Merlin specifically considers missing rates that are not fixed~\citep{yu2025merlin}. Representation and optimization choices supply further axes of variation, including SRSNet's patch-based selection of representation spaces~\citep{wu2025srsnet} and the decomposition-based objective of DBLoss~\citep{qiu2025DBLoss}. Together, these studies contextualize the diversity of data conditions and learning choices that a forecasting curriculum may encounter. Aurora-X develops its capabilities through successive changes to the training configuration, followed by adaptation to multiple temporal resolutions.

\paragraph{Uncertainty estimation and task adaptation.}
Different downstream uses also call for different outputs. $K^2$VAE combines Koopman and Kalman components with variational inference for probabilistic forecasts~\citep{wu2025k2vae}; MetaGNSDformer targets point and interval estimates of lithium-ion battery remaining useful life~\citep{cheng2026metagnsdformer}. Aurora-X exposes uncertainty through a continuously conditioned quantile head. Other work extends the task repertoire of temporal models: STAR adapts a foundation model to anomaly detection~\citep{cheng2026star}, CATCH detects anomalies with frequency patching that accounts for channels~\citep{wu2024catch}, and TAB standardizes benchmarking for detection methods~\citep{qiu2025tab}. TimeART investigates reasoning over time series with agentic tool use~\citep{wu2026timeart}. These directions locate forecasting within a larger collection of predictive, diagnostic, and reasoning capabilities.

\end{document}